\documentclass[conference]{IEEEtran}
\IEEEoverridecommandlockouts
\usepackage{cite}
\usepackage{amsmath,amssymb,amsfonts}
\usepackage{algorithmic}
\usepackage{graphicx}
\usepackage{textcomp}
\usepackage{xcolor}
\usepackage{packages}
\def\BibTeX{{\rm B\kern-.05em{\sc i\kern-.025em b}\kern-.08em
    T\kern-.1667em\lower.7ex\hbox{E}\kern-.125emX}}
\makeatletter
\newcommand{\linebreakand}{%
  \end{@IEEEauthorhalign}
  \hfill\mbox{}\par
  \mbox{}\hfill\begin{@IEEEauthorhalign}
}

\usepackage{amsmath,amsfonts,bm}

\def\eqref#1{equation~\ref{#1}}

\def\1{\bm{1}}

\DeclareMathAlphabet{\mathsfit}{\encodingdefault}{\sfdefault}{m}{sl}
\SetMathAlphabet{\mathsfit}{bold}{\encodingdefault}{\sfdefault}{bx}{n}

\newcommand{\E}{\mathbb{E}}

\newcommand{\R}{\mathbb{R}}

\DeclareMathOperator*{\argmax}{arg\,max}

\newcommand{\ra}[1]{\renewcommand{\arraystretch}{#1}}
\newcommand{\auc}{\text{AUROC$\uparrow$\%}}
\newcommand{\fpr}{\text{FPR$\downarrow_{95\%}$}\%}

\newcommand{\x}{\mathbf{x}}
\newcommand{\Q}{\mathcal{Q}}
\newcommand{\logit}{\mathbf{u}}
\newcommand\Tstrut{\rule{0pt}{2.6ex}}         
\newcommand\Bstrut{\rule[-0.9ex]{0pt}{0pt}}   

\newcommand{\classifier}{h_{\theta}}

\def\myeq{\mathrel{\ensurestackMath{\stackon[1pt]{=}{\scriptscriptstyle\Delta}}}}
\useunder{\uline}{\ul}{}

\definecolor{Gray}{gray}{0.85}
\newcolumntype{a}{>{\columncolor{Gray}}c}
\definecolor{lightcyan}{rgb}{0.88, 1.0, 1.0}
\definecolor{lightyellow}{rgb}{1.0, 1.0, 0.88}
\newcolumntype{b}{>{\columncolor{lightcyan}}r}
\newcolumntype{y}{>{\columncolor{lightyellow}}r}
\definecolor{lightmauve}{rgb}{0.86, 0.82, 1.0}
\definecolor{almond}{rgb}{0.94, 0.87, 0.8}
\definecolor{lightapricot}{rgb}{0.99, 0.84, 0.69}
\definecolor{snow}{rgb}{1.0, 0.98, 0.98}
\newcolumntype{d}{>{\columncolor{snow}}c}

\newtheorem{definition}{Definition}
\crefname{definition}{Definition}{Definitions}
\Crefname{definition}{Definition}{Definitions}

\newcommand{\EE}{\mathbb{E}}

\newcommand{\rset}{\mathbb{R}}

\newcommand{\calX}{\mathcal{X}}

\newcommand{\calY}{\mathcal{Y}}

\newcommand{\calU}{\mathcal{U}}
\newcommand{\calK}{\mathcal{K}}
\newcommand{\calZ}{\mathcal{Z}}

\newcommand{\Ell}{\mathcal{L}}

\definecolor{darkgreen}{rgb}{0., .5, .1}
\definecolor{bostonuniversityred}{rgb}{0.8, 0.0, 0.0}

\definechangesauthor[color=darkgreen]{MR}
\definechangesauthor[color=bostonuniversityred]{FG}
\definechangesauthor[color=blue]{MP}
\newcommand{\classSoftProb}[1]{p_{\widehat{Y}|X}(#1|\x; \theta)}
\newcommand{\classSoftProbAdvIn}[1]{p_{\widehat{Y}|X}(#1|\x_\ell^{\prime}; \theta)}
\newcommand{\classSoftProbAdvInGini}[1]{p^2_{\widehat{Y}|X}(#1|\x_\ell^{\prime}; \theta)}

\newcommand{\mead}{\textsc{Mead}}
\newcounter{todocounter}

\begin{document}
\title{A Minimax Approach Against Multi-Armed Adversarial Attacks Detection
\thanks{$^\star$ equal contribution.\\
The work of Federica Granese was supported by the European Research Council (ERC) project HYPATIA under the European Union’s
Horizon 2020 research and innovation program. Grant agreement N. 835294.}
}

\author{\IEEEauthorblockN{Federica Granese$^\star$}
\IEEEauthorblockA{
\textit{Lix, Inria-Saclay}\\
\textit{Institute Polytechnique de Paris} \\
\textit{Sapienza University of Rome}\\
Palaiseau, France \\
federica.granese@inria.fr}
\and
\IEEEauthorblockN{Marco Romanelli$^\star$}
\IEEEauthorblockA{\textit{New York University} \\
New York, NY, USA \\
mr6852@nyu.edu}
\and
\IEEEauthorblockN{Siddharth Garg}
\IEEEauthorblockA{\textit{New York University} \\
New York, NY, USA \\
sg175@nyu.edu}
\linebreakand
\IEEEauthorblockN{Pablo Piantanida}
\IEEEauthorblockA{
\textit{International Laboratory on Learning Systems (ILLS)} \\
\textit{CNRS - CentraleSupélec}\\
Montréal, Canada \\
pablo.piantanida@centralesupelec.fr}
}
\maketitle

\begin{abstract}
Multi-armed adversarial attacks, in which multiple algorithms and objective loss functions are simultaneously used at evaluation time, have been shown to be highly successful in fooling state-of-the-art adversarial examples detectors while requiring no specific side information about the detection mechanism. By formalizing the problem at hand, we can propose a solution that aggregates the soft-probability outputs of multiple pre-trained detectors according to a minimax approach. The proposed framework is mathematically sound, easy to implement, and modular, allowing for integrating existing or future detectors. Through extensive evaluation on popular datasets (e.g., CIFAR10 and SVHN), we show that our aggregation consistently outperforms individual state-of-the-art detectors against multi-armed adversarial attacks, making it an effective solution to improve the resilience of  available  methods.
\end{abstract}

\section{Introduction}
\label{sec:intro}
In recent years, the need for deep learning models that are both reliable and accurate has sparked significant interest in the field of trustworthy AI across multiple research domains. Efforts that aim to provide a deeper understanding of the limitations and capabilities of deep learning models and to develop methods that can improve their reliability and robustness in real-world applications have been focused on several key areas. Detection of misclassified samples~\cite{GraneseRGPP2021NeurIPS,GeifmanE19,GangradeKS21}, identification of out-of-distribution patterns~\cite{Eduardo,vyas2018outofdistribution,gram_matrice,ovadia2019trust,liu2020energybased,baseline,ZhangGR2021ICML,LinRLCVPR21}, enhancement of model robustness against adversarial attacks~\cite{MadryMSTV2018ICLR,zhang2019theoretically,alayrac2019labels,PicotMBLBAP2022TPAMI,robey2021adversarial,EngstromTTSM19}, and detection of adversarial attacks~\cite{aldahdooh2022adversarial,NSS,LID,Mahalanobis,MagNet,FS,KDBU} are the most relevant research directions in the field.

In particular, we consider the problem of adversarial examples. These examples are crafted patterns specifically designed starting from `natural' or `clean' samples to fool a model into making incorrect predictions. To combat this issue, there are two main strategies: robust training and adversarial detection. Robust training (e.g.,~\cite{MadryMSTV2018ICLR,PicotMBLBAP2022TPAMI,Zimmermann2022,ZhouW00WZL22,Rebuffi}) aims to make a model more resistant to adversarial examples, while adversarial detection (e.g.,~\cite{aldahdooh2022adversarial,PangZHD000L22,RaghuramCJB21}) attempts to identify and reject such examples. Our focus is on the second defense strategy, adversarial detection. Recent findings have shown that attackers with little or no information about the specific defense can still cause significant damage, highlighting the importance of ongoing research to develop robust and effective adversarial detection methods.

Traditional detection methods are often evaluated using a single attack strategy, which does not accurately reflect real-world threats. More recent papers such as~\cite{TramerCBM20} highlight the importance of testing proposed defenses against adaptive attacks. These are attacks that are specifically designed to target a specific defense method and take advantage of a large amount of prior knowledge about the defense mechanism, such as the loss function optimized by the defense. This worst-case scenario evaluation is crucial when proposing new detection methods to assess their robustness. However, it appears that no defense is completely invulnerable when so much side information is provided. Crucially,~\cite{GranesePRMP2022ECMLPKDD} has shown that even attackers with less side information can easily fool a detector using a combination of well-known attacks, without any prior knowledge of the detector itself. In a multi-armed attack scenario, a given pattern is perturbed using multiple strategies and loss functions simultaneously\footnote{Henceforth, the terms ``multi-armed'' and ``simultaneous'' will be used interchangeably.} and detection is considered successful only if all attacks are correctly identified. 

Although~\cite{GranesePRMP2022ECMLPKDD} highlights the problem of simultaneous adversarial attack detection, it does not provide a solution. In this paper, we aim to address this issue by proposing a simple yet effective method for aggregating multiple detection methods to create a ``team of experts'' using a minimax approach. Our proposed framework is highly flexible, allowing for the combination of any existing or future supervised or unsupervised method as long as its output can be interpreted as a probability distribution over two categories. Additionally, our modular aggregator allows pre-trained detectors to be reused without additional training or data, and can be easily extended to new detection methods.


\subsection{Summary of contributions}
Our contributions are threefold:
\begin{itemize}
\item To the best of our knowledge, our proposed aggregation framework is the first to combine the expertise of different adversarial examples detectors and address the problem of simultaneous attack detection as highlighted in~\cite{GranesePRMP2022ECMLPKDD}. This method can aggregate pre-trained detectors without the need for additional training.
\item From a theoretical perspective, we revisit the simultaneous attack detection problem as formulated in~\cite{GranesePRMP2022ECMLPKDD} and formalize it as a minimax cross-entropy risk. Based on this formulation, we derive a surrogate loss function and use it to characterize our optimal soft-detector in~\cref{eq:minimaxProb4}, leading to our proposed solution.
\item Empirical evaluations of our proposed solution on popular datasets, such as CIFAR10 and SVHN, show that it leads to higher and more consistent performance compared to the state-of-the-art (SOTA) in the simultaneous attack setup, even when using simple detectors that individually perform worse than SOTA detectors, as demonstrated in~\cref{sec:experiments}.
\end{itemize}


\subsection{Related works}
\paragraph{Detection mechanisms:}
\label{sec:detectors}
Methods to defend deep models against adversarial attacks can be grouped into two main families: methods that are designed to increase the targeted model's robustness by re-training it~\cite{FGSM,MadryMSTV2018ICLR,PicotMBLBAP2022TPAMI,Cihang2006-14536,TramerKPGBM18}, and methods engineered to detect adversarial examples at evaluation time~\cite{NSS,LID,KDBU,FS,MagNet,Mahalanobis}. The work in~\cite{aldahdooh2022adversarial} provides a recent and thorough survey about the state-of-the-art detection methods, which fall under two main categories:  \textit{supervised} and \textit{unsupervised}. Detectors within the former category extract features either directly from the targeted network's layer~\cite{NSS,KDBU} or by using statistical tools~\cite{LID,Mahalanobis}. To do so, both natural and adversarial examples are necessary. 
Generally, the adversarial samples are created according to a single fixed algorithm and a given loss function, which are then also used to create the examples at evaluation time. Methods falling under the unsupervised category only rely on the features of natural samples that can be extracted using different techniques (e.g., \textit{feature squeezing}~\cite{FS}) or can be based on autoencoders training procedures with the scope of minimizing the reconstruction error~\cite{MagNet}.

\paragraph{Attack algorithms:}
\label{sec:attacks}
Since~\cite{SzegedyZSBEGF13} first shed light on the problem, several machine learning models, including state-of-the-art neural networks, have been found to be vulnerable to adversarial examples. Over the years, a plethora of algorithms to generate adversarial samples has been proposed and overall, we can group them into two main categories: \textit{white-box} and \textit{black-box} attacks. We talk about \textit{white-box} attacks when the adversary knows everything about the target model (its architecture and weights). \textit{Gradient-based} attacks belong to this category. They rely on finding the perturbation direction, i.e., the sign of gradient at each pixel of the input, that maximizes the attacker's objective value. 

Examples of gradient-based attacks are the \textit{Fast Gradient Sign Method} (FGSM)~\cite{FGSM}, the \textit{Basic Iterative Method} (BIM)~\cite{BIM} and the \textit{Projected Gradient Descent} method (PGD)~\cite{MadryMSTV2018ICLR}. BIM and PGD can be seen as iterative versions of FGSM (one-step perturbation). Unlike BIM, PGD attacks start from a random perturbation in L$_p$-ball around the input sample. Another powerful attack is the \textit{Carlini-Wagner} attack (CW)~\cite{CarliniW2017SP}, which directly minimizes the additive noise constrained by a function which assure the misclassification of the perturbed sample. 
We conclude the list of white-box attacks by mentioning the \textit{DeepFool} attack (DF)~\cite{DF}, which is an iterative method based on a local linearization of the targeted classifier, and the resolution of the resulting simplified adversarial problem. In the case of \textit{black-box} attacks, the adversary has no access to the internals of the target model, hence it creates attacks by querying the model and monitoring 
outputs of the 
model to attack. Examples of black-box attacks are the \textit{Square Attack} (SA)~\cite{SA}, which iteratively searches
for a random perturbation, and checks if it increases the attacker's objective at each step; the \textit{Hop Skip Jump} attack (HOP)~\cite{HOP} which estimates the gradient direction to perturb, and the \textit{Spatial Transformation Attack} (STA)~\cite{EngstromTTSM19} which transforms the original samples by applying small translations and rotations to them.
It is worth to mention that there also exists \textit{gray-box} attacks, i.e. when the adversary knows the training data but not the internals of the model. These attacks rely on the transferability property of the adversarial examples: to create attacks these methods build a substitute model that performs the same task as the target model. 
A special class of attacks are the so-called \textit{adaptive attacks}~\cite{athalye2018obfuscated,TramerCBM20,CarliniW2017SP,yao2021automated} where attacks are specifically designed to target a given defence. In this scenario, the attacker is supposed to have full knowledge of both the targeted classifier and the underlying defence.

We refer to the survey in~\cite{aldahdooh2022adversarial} and references therein for a comprehensive discussion of these topics.  
\section{Main Definitions and Preliminaries}
Adversarial examples are carefully crafted input patterns designed to deceive a target classifier into making an incorrect decision, while remaining as similar as possible to the original sample. This section will provide a brief overview of the key concepts related to this topic.
\subsection{Target classifier}
Let $\calX\subseteq\rset^d$ be the input space and let $\calY=\{1,\dots, C\}$ be the label space related to a classification task. We denote by $P_{XY}$ the unknown data distribution over $\calX\times\calY$. 
Throughout the paper, we refer to the \textit{classifier} with
$\classSoftProb{y}$, i.e. the parametric soft-probability model, where $y\in\calY$, $\widehat{Y}$ is random variable representing the classifier's inference, and $\mathbf{\theta}\in\Theta$ represents the learned parameters. The function $\classifier:\calX\rightarrow\R^{|\calY|}$ outputs the logits vector of the classifier given an input sample.
The induced hard decision of the classifier is defined as $g_{\theta}:\mathcal{X}\rightarrow\mathcal{Y}$
s.t.
${g_{\mathbf{\theta}}(\x)\myeq \arg\max_{y\in\calY} \classSoftProb{y}}$.

\subsection{Adversarial problem}
\label{sec:adv_problem}
Let us consider a natural sample, denoted by $\x\in\calX$, along with its true label, $y\in\calY$. An attacker aims to deceive the model $g_{\theta}$ by crafting an adversarial example, $\x_\ell^{\prime}\in\mathcal{I}\subseteq\R^d$, where $\mathcal{I}$ is a held-out set of images that is distributed according to $P_{XY}$ but that was not used during training. The symbol $\ell$ denotes the objective loss function $\ell(\x, \x_\ell^{\prime};\mathbf{\theta})$ optimized by the attacker; $\varepsilon$ is perturbation magnitude, and L$_p$, $p\in\{1, 2, \infty\}$ is the norm constraint. The goal of the attack is to obtain an $\x_\ell^{\prime}$ such that $g_{\theta}(\x_\ell^\prime)\neq g_{\theta}(\x)$, in order to force the target model to make a prediction error.  As thoroughly investigated in~\cite{SzegedyZSBEGF13}, the adversarial generation problem is difficult to tackle and it is commonly relaxed as follows
\begin{equation}
    \mathbf{x_{\ell}}^{\prime} \equiv \mathbf{x_{\ell}}^{\prime} (\x)  =\underset{\mathbf{x_\ell}^{\prime}\in \R^{d} \,:\,\lVert\mathbf{x_{\ell}}^{\prime} - \x\rVert_{p}<\varepsilon
  }{\argmax}\ell(\x,\mathbf{x_{\ell}}^{\prime}; \mathbf{\theta)},
  \label{eq:relaxed_adv_problem}
\end{equation}
where $\x_\ell^{\prime}$ is updated iteration by iteration starting from an initial given value.
The objective function $\ell$ traditionally used is the Adversarial Cross-Entropy (ACE)~\cite{SzegedyZSBEGF13,MadryMSTV2018ICLR}:
\begin{equation}
\label{eq:ACE}
    \ell_{\text{ACE}}(\mathbf{x},\mathbf{x_\ell}^{\prime}; \theta) =  \E_{Y|\x }\big[ - \log \classSoftProbAdvIn{Y} \big],
\end{equation}
where the expectation is understood to be over the ground true conditional distribution of $Y$ given $\x$.
Recent developments in the fields of robustness and misclassification detection ~\cite{GraneseRGPP2021NeurIPS,PicotMBLBAP2022TPAMI,zhang2019theoretically} have inspired the work on multi-armed attacks in~\cite{GranesePRMP2022ECMLPKDD}, which incorporates novel objective functions for generating diverse adversarial examples. These functions are briefly summarized below.
\begin{itemize}
\item The Kullback-Leibler divergence (KL):
\begin{align}
\label{eq:KL}
    \ell_{\text{KL}}\left(\x, \mathbf{x_{\ell}}^{\prime}; \theta \right)=\E_{{\widehat{Y}|\x}}\left[\log\left(\frac{\classSoftProb{\widehat{Y}}}{\classSoftProbAdvIn{\widehat{Y}}}\right)  \right].
\end{align}
\item The Fisher-Rao objective (FR)~\cite{PicotMBLBAP2022TPAMI}:
\begin{align}  
\label{eq:FR}
\ell_{\text{FR}}(\x, \mathbf{x_{\ell}}^{\prime}; \theta) &=2 \arccos \left(\mathcal{E}\right),
\end{align}
where $\mathcal{E}=\sum_{y \in \mathcal{Y}} \sqrt{\classSoftProb{y}
 \classSoftProbAdvIn{y}}$.
\item The Gini Impurity score (Gini)~\cite{GraneseRGPP2021NeurIPS}:
\begin{align}
    \label{eq:Gini}
    \ell_{\text{Gini}}(\cdot, \mathbf{x_{\ell}}^{\prime}; \theta)
    &=1 - \sqrt{\sum_{y\in\calY}\classSoftProbAdvInGini{y}}.
\end{align}
\end{itemize}
\section{Multi-armed adversarial attack detection and \textsc{Mead}}
\label{sec:mead_position}
Finding a framework to assess the robustness of adversarial attack detection is crucial in establishing trust in this defense. Except for defenses that are formally certified to be robust within a certain radius~\cite{CohenRK2019ICML} and whose practical usability is still under investigation and appears to be effective mainly against black-box attacks~\cite{MahoFLM2022ICASSP}, the majority of defenses presented in the literature require extensive empirical evaluation. The authors of~\cite{TramerCBM20} suggest that for each defense, adaptive attacks should be handcrafted by providing side information to the attacker on the internal mechanism of the defense mechanism. For instance, revealing the loss function optimized by the defense is often enough to craft powerful attacks by reversing the gradient descent on natural samples. 

While adaptive attacks require disclosing much information about the defense mechanism, even more alarmingly,~\cite{GranesePRMP2022ECMLPKDD} has exposed that much less information is required to mount multi-arm attacks that drastically affect the performance of SOTA adversarial detection mechanisms.
In particular, according to the latter framework, the target classifier is attacked simultaneously with multiple attack strategies without extra information on the specific detector. To create a set of simultaneous attacks, multiple perturbed versions of the same natural input sample are created according to the set of attack strategies, discarding those that are unable to fool the target classifier,  perturbation magnitude, $\varepsilon$, and the norm,
$\Ell_p$. The detector is then evaluated on all the crafted adversarial examples, and only if all the attacks are correctly detected is the detection successful.
Interestingly enough,~\cite{GranesePRMP2022ECMLPKDD} provides empirical evidence for the ``no-free-lunch-theorem'' in~\cite{TramerCBM20}, which states that for each possible attack, a defense can be deceived that provides no guarantees of robustness against any other attack. The multi-armed attack scenario, in particular, suggests that there may exist attacks that are just as damaging as adaptive attacks but require much less information about the specific detector being used, making this a more realistic and likely scenario to occur.
 
In this paper, we aim to investigate possible defenses against the multi-armed attack scenario. To do this, we formalize the problem and propose a solution incorporating an information-theoretic minimax approach. An analysis of the adaptive attack within our proposed framework can be found in~\cref{subsec:adaptive_attacks}.

Finally, it is worth noting that recent work has started to look for a connection between adversarial training and adversarial examples detection~\cite{Tramer2021}

\section{Formalization of the Problem of Detecting multi-armed Adversarial Attacks}\label{sec:math_framework}
In this section, we begin by formalizing the problem of multi-armed attacks as proposed in~\cite{GranesePRMP2022ECMLPKDD}. We then delve deeper into the topic of optimal detectors, and demonstrate how to apply our proposed solution to practical use-cases.

\subsection{Statistical model}


Let $\mathcal{K}$ be the countable set of indexes corresponding to each  possible attack, e.g., based on various attack algorithms and loss functions, as described in~\cref{sec:adv_problem}. Let  $\mathcal{M}=\big\{P^{(k)}_{XZ}\,  : \, k\in\mathcal{K} \big\}$ be the set of  joint probability distributions on  $\calX\times\calZ$ which are indexed with ${k,~\forall k\in\calK}$, where $\calX$ is the input (feature) space and $\calZ=\{0, 1\}$ indicates a binary space label for the adversarial example detection task. At the evaluation time,  the attacker selects an arbitrary strategy $k\in \mathcal{K}$ and then samples an input  according to $p^{(k)}_{X|Z}(\x|z=1)$ which corresponds to the probability density function induced by the chosen attack $k$ where $p^{(k)}_{X|Z}(\x|z=0)= p_X(\x) $ \emph{almost surely} corresponds to the probability distribution of the natural samples. The learner is given a set of \textit{soft-detectors} models:
$$
\Q = \left\{q_{\widehat{Z}|\logit}^{(k)} \, :\, \calU \mapsto [0,1]^2  \right\}_{k\in \calK},
$$
which have possibly been trained to detect attacks according to each strategy $k\in\mathcal{K}$, e.g.,  ${q_{\widehat{Z}|\logit}^{(k)} \equiv  p_{\widehat{Z}|U}(z|\logit; \psi_k)}$ with parameters $\psi_k$ and $\logit\in\calU = \{\classifier(\x)~|~\x\in \R^d\}$ denotes the  space of logits. 
The set of possible detectors $\Q$ is available to the defender. However, the specific attack chosen by the attacker at the test time is unknown. In the remainder of this section, we formally devise an optimal detector that exploits full knowledge of the set $\Q$. 

\subsection{A novel objective for detection under simultaneous attacks}
Consider a fixed input sample $\mathbf{x_0}$ and let $\mathbf{u_0}=\classifier(\mathbf{x_0})$. Clearly, the problem at hand consists in finding an optimal soft-detector  ${q}^{\star}_{\widehat{Z}|\mathbf{u_0}}$ that performs well simultaneously over all possible attacks in  $\calK$. This can be formalized as the solution to the following minimax problem: 
\begin{equation}
\mathcal{L}(\Q, \mathbf{x_0}) = \min_{{q}_{\widehat{Z}|\mathbf{u_0}}} \max_{k\in\calK}\, \EE_{q^{(k)}_{\widehat{Z}| \mathbf{u_0}}}\left[ -\log {q}_{\widehat{Z}|\mathbf{u_0}}  \right], \label{optimal-loss}
\end{equation}
which requires to solve \eqref{optimal-loss} for $\Q$ and for each given input sample $\mathbf{x_0}$. 
It is important to note that the minimization is performed over all (detectors)  distributions ${q}_{\widehat{Z}|\mathbf{u_0}}$, including elements that are not part of the set $\mathcal{Q}$. 

That being said, the objective in~\cref{optimal-loss} is not tractable computationally. To overcome this issue, we derive a surrogate (an upper bound) that can be computationally optimized. For any   arbitrary choice of  ${q}_{\widehat{Z}|\mathbf{u_0}}$, we have 
\begin{align}
  \max_{k\in\calK}\,  \EE_{q^{(k)}_{\widehat{Z}| \mathbf{u_0}}} &\left[ -\log {q}_{\widehat{Z}|\mathbf{u_0}}  \right] \leq \nonumber\\ 
  &\underbrace{\max_{k\in\calK}\, \EE_{q^{(k)}_{\widehat{Z}| \mathbf{u_0}}}\left[ -\log q^{(k)}_{\widehat{Z}| \mathbf{u_0}} \right]}_{=\textrm{constant term}} + \nonumber\\ &+
  \max_{k\in\calK}\, \EE_{q^{(k)}_{\widehat{Z}| \mathbf{u_0}}}\left[\log\left(\frac{q^{(k)}_{\widehat{Z}|\mathbf{u_0}}}{{q}_{\widehat{Z}|\mathbf{u_0}}}\right)\right].  
  \label{eq-missing}
\end{align}
\begin{proof}[Proof of~\cref{eq-missing}]
\begin{align*}
&\max_{k\in\calK}\, \EE_{q^{(k)}_{\widehat{Z}| \mathbf{u_0}}}\left[ -\log {q}_{\widehat{Z}|\mathbf{u_0}}  \right] =\\
&=
\max_{k\in\calK}\, \left[\EE_{q^{(k)}_{\widehat{Z}| \mathbf{u_0}}}\left[ -\log q^{(k)}_{\widehat{Z}| \mathbf{u_0}} \right] + \EE_{q^{(k)}_{\widehat{Z}| \mathbf{u_0}}}\left[\log\left(\frac{q^{(k)}_{\widehat{Z}|\mathbf{u_0}}}{{q}_{\widehat{Z}|\mathbf{u_0}}}\right)\right]\right]\\
&\leq \max_{k\in\calK}\, \EE_{q^{(k)}_{\widehat{Z}| \mathbf{u_0}}}\left[-\log q^{(k)}_{\widehat{Z}| \mathbf{u_0}} \right] +\\&~~~~~~~~~~~~~~~~~~~~~~~~~~~+ \max_{k\in\calK}\, \EE_{q^{(k)}_{\widehat{Z}| \mathbf{u_0}}}\left[\log\left(\frac{q^{(k)}_{\widehat{Z}|\mathbf{u_0}}}{{q}_{\widehat{Z}|\mathbf{u_0}}}\right)\right].     
\end{align*}
\end{proof}
Observe that the first term in \eqref{eq-missing} of the upper bound  is constant w.r.t. the choice of  ${q}_{\widehat{Z}|\mathbf{u_0}}$ and the second term is well-known as being equivalent to the \emph{average worst-case  regret}~\cite{BarronRY1998TInfT}. This upper bound provides a surrogate to our intractable objective in \eqref{optimal-loss} that can be minimized over all ${q}_{\widehat{Z}|\mathbf{u_0}}$.  We can formally state our problem as follows: 
\begin{align}
    \label{eq:minimaxProb1}
  \tilde{\mathcal{L}}(\Q,\mathbf{x_0}) = &    \min_{{q}_{\widehat{Z}|\mathbf{u_0}}}\max_{k\in\calK}\, \EE_{q^{(k)}_{\widehat{Z}| \mathbf{u_0}}}\left[\log\left(\frac{q^{(k)}_{\widehat{Z}|\mathbf{u_0}}}{{q}_{\widehat{Z}|\mathbf{u_0}}}\right)\right]=\nonumber\\=&\min_{{q}_{\widehat{Z}|\mathbf{u_0}}}\max_{\textcolor{black}{P_{\Omega}}}\, \EE_{\Omega}\left[D_{\textrm{KL}}\left(q^{(\Omega)}_{\widehat{Z}|\mathbf{u_0}}\big \| {q}_{\widehat{Z}|\mathbf{u_0}}\right)\right],
\end{align}
\textcolor{black}{where the $\min$ is taken over all the possible distributions ${q}_{\widehat{Z}|\mathbf{u_0}}$; 
and $\Omega$ is a discrete random variable with $P_{\Omega}$ denoting a generic  probability  distribution whose probabilities are  $(\omega_1,\dots,\omega_{|\calK|})$, i.e., $P_{\Omega}(k)=\omega_k$;}
and $D_{\textrm{KL}}(\cdot\|\cdot)$ is the Kullback–Leibler divergence, representing the expected value of regret of ${q}_{\widehat{Z}|U}$ w.r.t. the worst-case distribution in   $\Q$. 
\begin{proof}[Proof of~\cref{eq:minimaxProb1}]
    The equality hold by noticing that 
\begin{align*}
    \max_{P_{\Omega}}&\, \EE_{\Omega}\left[D_{\textrm{KL}}\left(q^{(\Omega)}_{\widehat{Z}|\mathbf{u_0}}\big \| {q}_{\widehat{Z}|\mathbf{u_0}}\right)\right] \nonumber\\
    &\leq
    \max_{k\in\calK}\, \EE_{q^{(k)}_{\widehat{Z}| \mathbf{u_0}}}\left[\log\left(\frac{q^{(k)}_{\widehat{Z}|\mathbf{u_0}}}{{q}_{\widehat{Z}|\mathbf{u_0}}}\right)\right],
\end{align*}
and moreover, 
\begin{align*}
    \max_{k\in\calK}\, \EE_{q^{(k)}_{\widehat{Z}| \mathbf{u_0}}}\left[\log\left(\frac{q^{(k)}_{\widehat{Z}|\mathbf{u_0}}}{{q}_{\widehat{Z}|\mathbf{u_0}}}\right)\right]& = \EE_{\bar{\Omega}}\left[D_{\textrm{KL}}\left(q^{(\bar{\Omega})}_{\widehat{Z}|\mathbf{u_0}}\big \| {q}_{\widehat{Z}|\mathbf{u_0}}\right)\right],
\end{align*}
by choosing the random variable $\bar{\Omega}$ with uniform probability over the set of maximizers  ${\overline{\mathcal{K}}=\argmax_{k\in\calK}\, \EE_{q^{(k)}_{\widehat{Z}| \mathbf{u_0}}}\left[\log\left(\frac{q^{(k)}_{\widehat{Z}|\mathbf{u_0}}}{{q}_{\widehat{Z}|\mathbf{u_0}}}\right)\right]}$, zero otherwise. 
\end{proof}
The convexity of the KL-divergence allows us to rewrite~\cref{eq:minimaxProb1} as follows:
\begin{align}
    \label{eq:minimaxProb3}
    &\min_{{q}_{\widehat{Z}|\mathbf{u_0}}}\max_{\textcolor{black}{P_\Omega}}\EE_{\Omega}\left[D_{\textrm{KL}}\left(q^{(\Omega)}_{\widehat{Z}|\mathbf{u_0}}\big \| {q}_{\widehat{Z}|\mathbf{u_0}}\right)\right]=\nonumber\\&=  \max_{\textcolor{black}{P_\Omega}}\min_{\widehat{q}_{\widehat{Z}|\mathbf{u_0}}}\EE_{\Omega}\left[D_{\textrm{KL}}\left(q^{(\Omega)}_{\widehat{Z}|\mathbf{u_0}}\big \| {q}_{\widehat{Z}|\mathbf{u_0}}\right)\right]. \end{align}
\begin{proof}[Proof of~\cref{eq:minimaxProb3}]
We consider a zero-sum game with a concave-convex mapping defined on a product of convex sets. The sets of all probability distributions ${q}_{\widehat{Z}|\mathbf{u_0}}$ and $P_\Omega$ are two nonempty convex sets, bounded and finite dimensional. On the other hand, $\big(P_\Omega,{q}_{\widehat{Z}|\mathbf{u_0}} \big)\rightarrow  \EE_{\Omega}\left[D_{\textrm{KL}}\left(q^{(\Omega)}_{\widehat{Z}|\mathbf{u_0}}\big \| {q}_{\widehat{Z}|\mathbf{u_0}}\right)\right]$ is a concave-convex mapping, i.e., $P_\Omega\rightarrow  \EE_{\Omega}\left[D_{\textrm{KL}}\left(q^{(\Omega)}_{\widehat{Z}|\mathbf{u_0}}\big \| {q}_{\widehat{Z}|\mathbf{u_0}}\right)\right]$  is concave and ${q}_{\widehat{Z}|\mathbf{u_0}} \rightarrow  \EE_{\Omega}\left[D_{\textrm{KL}}\left(q^{(\Omega)}_{\widehat{Z}|\mathbf{u_0}}\big \| {q}_{\widehat{Z}|\mathbf{u_0}}\right)\right]$ is convex for every $\big(P_\Omega,{q}_{\widehat{Z}|\mathbf{u_0}} \big)$.  Then, by classical min-max theorem ~\cite{vonNeumann1928-VONZTD-2} we have that~\cref{eq:minimaxProb3} holds. 
\end{proof}
The solution to~\cref{eq:minimaxProb3} provides the optimal distribution \textcolor{black}{$P_{\Omega}^\star$}, i.e. the collection of weights $\{w_k^{\star}\}$, which leads to our soft-detector~\cite{BarronRY1998TInfT}: 
\begin{align}
    \label{eq:minimaxProb4}
 \widehat{q}^{~\star}_{\widehat{Z}|\mathbf{u_0}} = \sum_{k\in\calK}w^{\star}_k \cdot q^{(k)}_{\widehat{Z}|\mathbf{u_0}}  , \ \ \text{ with } \ \ 
 \textcolor{black}{ P_{\Omega}^\star=\argmax_{\{\omega_k\}}I_{\mathbf{u_0}}(\Omega;\widehat{Z}), }
\end{align}
where $I_{\mathbf{u_0}}(\cdot;\cdot)$ denotes the Shannon mutual information between the random variable \textcolor{black}{$\Omega$}, distributed according to \textcolor{black}{$\{\omega_k\}$}, and the binary soft-prediction variable $\widehat{Z}$, distributed according to $q^{(k)}_{\widehat{Z}|\mathbf{u_0}} $ and  conditioned on the particular test example $\mathbf{u_0}$. 
\begin{proof}[Proof of~\cref{eq:minimaxProb4}]
It is enough to show that 
\begin{align}
\min_{\widehat{q}_{\widehat{Z}|\mathbf{u_0}}}\EE_{\Omega}\left[D_{\textrm{KL}}\left(q^{(\Omega)}_{\widehat{Z}|\mathbf{u_0}}\big \| {q}_{\widehat{Z}|\mathbf{u_0}}\right)\right] = I_{\mathbf{u_0}}(\Omega;\widehat{Z}), \label{eq-missing-appex}
    \end{align}
for every random variable $\Omega$ distributed according to an arbitrary probability distribution $P_{\Omega}$ and each distribution $q^{(\Omega)}_{\widehat{Z}|\mathbf{u_0}}$. We begin by showing that 
\begin{align*}
\EE_{\Omega}\left[D_{\textrm{KL}}\left(q^{(\Omega)}_{\widehat{Z}|\mathbf{u_0}}\big \| {q}_{\widehat{Z}|\mathbf{u_0}}\right)\right] & \geq  I_{\mathbf{u_0}}(\Omega;\widehat{Z}), 
    \end{align*}
for any arbitrary distributions $P_{\Omega}$ and   $q^{(\Omega)}_{\widehat{Z}|\mathbf{u_0}}$. To this end, we use the following identities: 
\begin{align}
&\EE_{\Omega}\left[D_{\textrm{KL}}\left(q^{(\Omega)}_{\widehat{Z}|\mathbf{u_0}}\big \| {q}_{\widehat{Z}|\mathbf{u_0}}\right)\right]
=\EE_{\Omega}\EE_{ q^{(\Omega)}_{\widehat{Z}|\mathbf{u_0}}}\left(\log\frac{q^{(\Omega)}_{\widehat{Z}|\mathbf{u_0}}}{ {q}_{\widehat{Z}|\mathbf{u_0}}}\right)=\nonumber\\
&= \EE_{\Omega}\EE_{ q^{(\Omega)}_{\widehat{Z}|\mathbf{u_0}}}\left(\log\frac{q^{(\Omega)}_{\widehat{Z}|\mathbf{u_0}}}{P_{\widehat{Z}} }\right) + \nonumber D_{\textrm{KL}}\left( P_{\widehat{Z}}  \| {q}_{\widehat{Z}|\mathbf{u_0}} \right)=\nonumber\\
&= I_{\mathbf{u_0}}(\Omega;\widehat{Z}) + D_{\textrm{KL}}\left( P_{\widehat{Z}}  \| {q}_{\widehat{Z}|\mathbf{u_0}} \right)\geq  I_{\mathbf{u_0}}(\Omega;\widehat{Z}), \label{eq-missing-appex-B}
\end{align}
where $P_{\widehat{Z}}$ denotes the marginal distribution of $q^{(\Omega)}_{\widehat{Z}|\mathbf{u_0}}$ w.r.t. $P_{\Omega}$ and the last inequality follows since the KL divergence is positive. Finally, it is easy to check that by selecting $ {q}_{\widehat{Z}|\mathbf{u_0}} = P_{\widehat{Z}} $ the lower bound in \eqref{eq-missing-appex-B} is achieved which proves the identity in expression \eqref{eq-missing-appex}. By taking the maximum overall probability distributions  $P_{\Omega}$ at both sides of expression \eqref{eq-missing-appex} the claim follows. 
\end{proof}

\noindent\textbf{From theory to our practical detector.} According to our derivation in~\cref{eq:minimaxProb4}, the optimal detector turns out to be given by a  mixture of the $|\calK|$ detectors belonging to the class $\Q$, with  weights carefully optimized to maximize the mutual information between $\Omega$ and the predicted variable $\widehat{Z}$ for each detector in the class $\Q$. Using this key ingredient, it is straightforward to devise our optimal  detector. 

\begin{definition}
\label{def:salad}
For any $0\leq\gamma\leq1$ and a given $\mathbf{x_0} \in\calX $, let us define the following detector $\textsc{D}:\mathbb{R}^d\rightarrow\{0, 1\}$:  
\begin{align}
\textsc{d}(\x_0)\myeq \mathds{1} \left[ {q}^{~\star}_{\widehat{Z}|\mathbf{u_0}}(\hat{z}=1|\classifier(\mathbf{x_0}))>\gamma \right],
\label{eq:salad}
\end{align}
where $\mathds{1} \left[\cdot\right]$ is the indicator function.
\end{definition}
\section{Experimental Results}
\label{sec:experiments}
\begin{figure}
    \centering
    \includegraphics[width=\columnwidth]{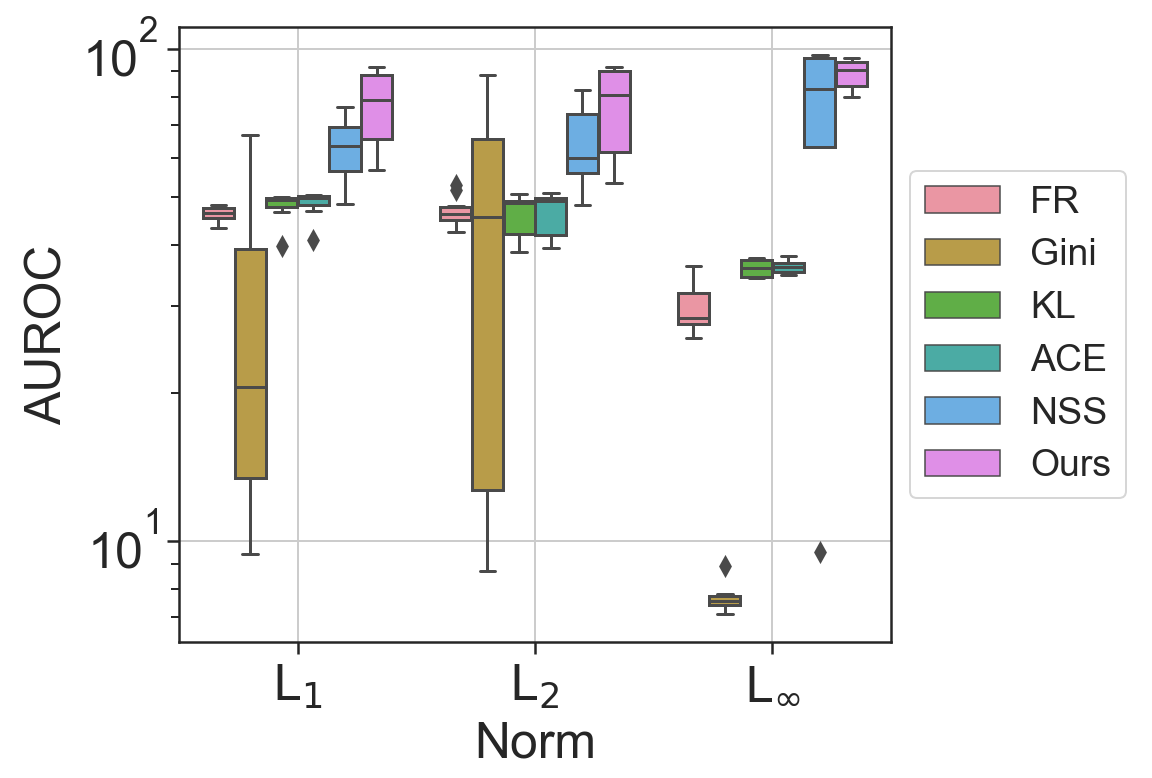}
    \caption{The \textit{shallow} detectors are named after the loss function used to craft the attacks they are trained to detect. Overall, the SOTA method NSS clearly outperforms all the individual shallow detectors. The aggregation we propose allows to use the shallow models to attain a detector whose performance are consistently comparable and in many cases better than SOTA.}
    \label{fig:box_plot}
\end{figure}
\begin{figure*}[t]
	\centering
	\begin{subfigure}[b]{1\columnwidth}
	    \centering
	    \includegraphics[width=\columnwidth]{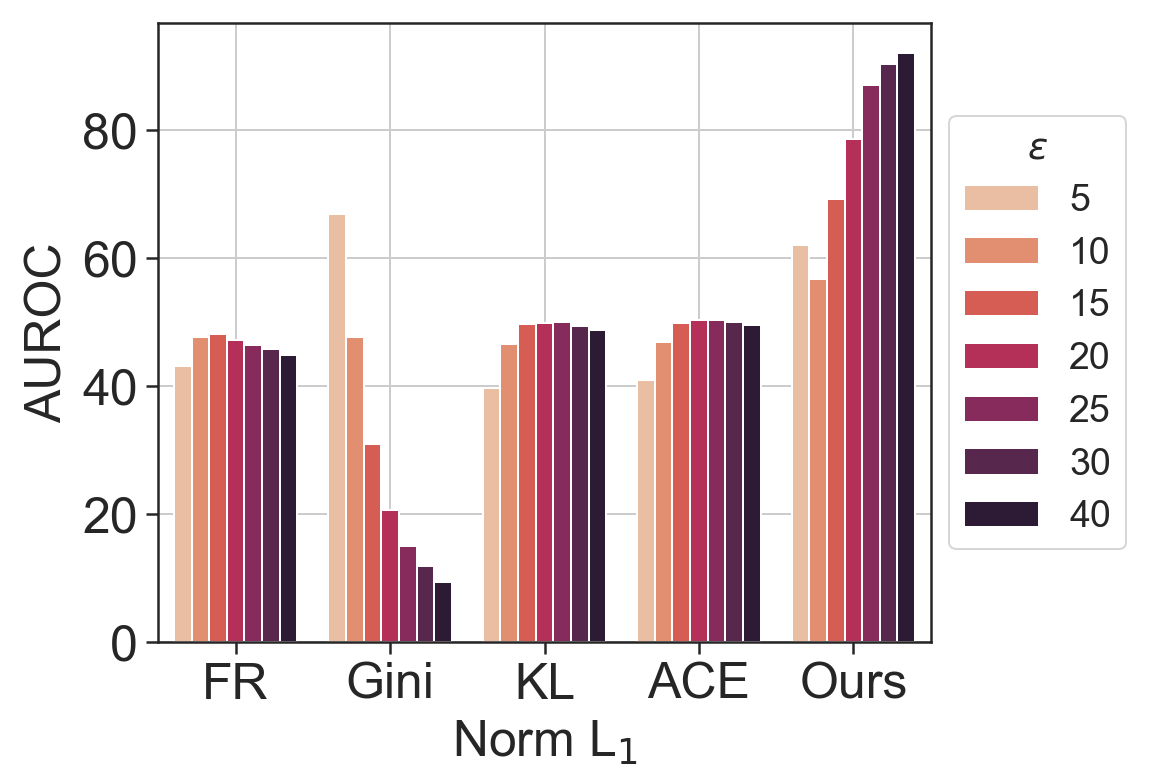}
	    \caption{}
	    \label{fig:eps_l1}
	\end{subfigure}
        \begin{subfigure}[b]{1\columnwidth}
            \centering
            \includegraphics[width=\columnwidth]{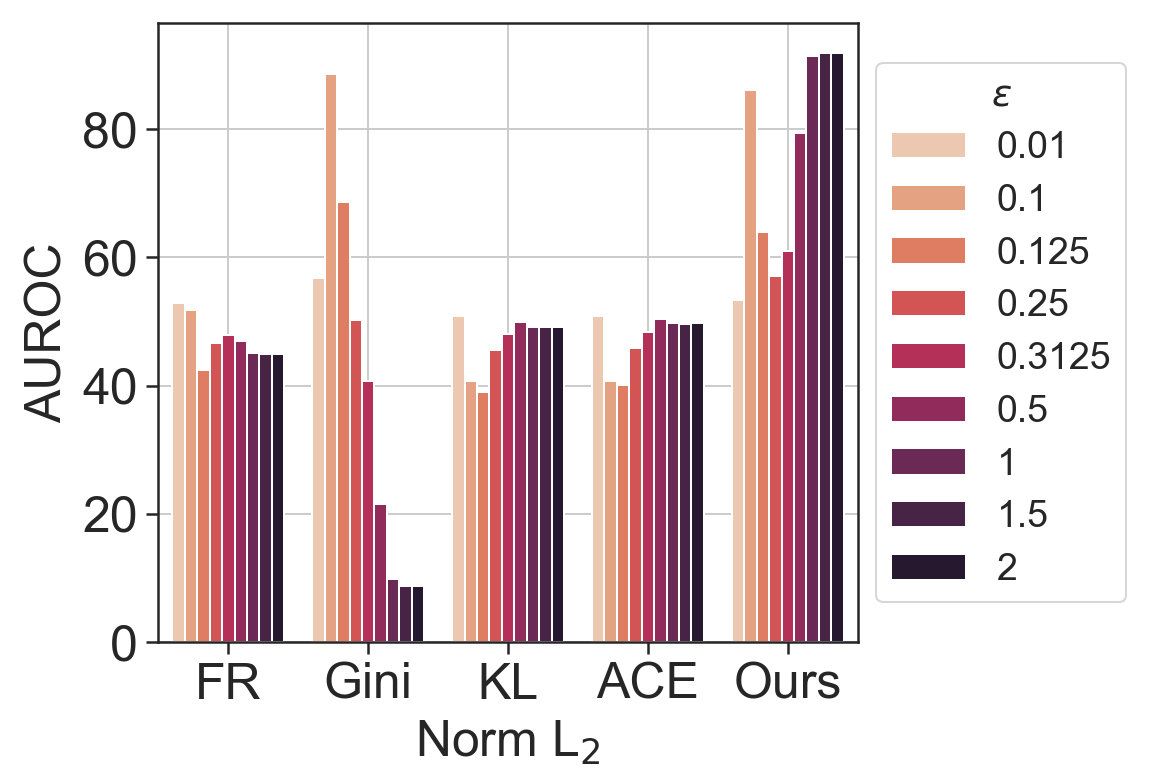}
            \caption{}
        \label{fig:eps_l2}
	\end{subfigure}
        \begin{subfigure}[b]{1\columnwidth}
            \centering
            \includegraphics[width=\columnwidth]{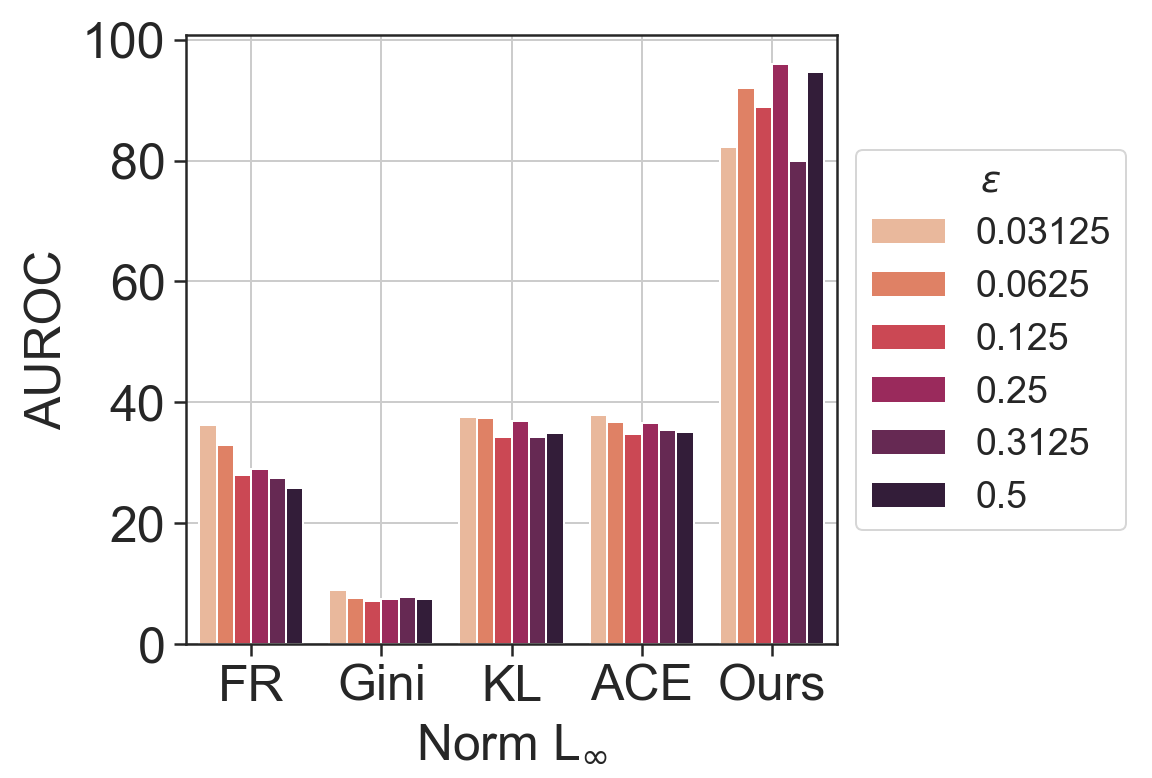}
            \caption{}
        \label{fig:eps_linf}
        \end{subfigure}
	\caption{Performance of the various detectors grouped by L$_p$-norm and perturbation magnitude $\varepsilon$ on CIFAR10. Each \textit{shallow} detector is named after the loss function used to craft the attacks they it is trained to detect. The plot shows how our method consistently attains better performance the the single one on all the different adversarial attacks, supporting the claim of optimality in~\cref{sec:math_framework}.}
	\label{fig:eps_norm}
\end{figure*}
\begin{table}[t]
\centering
\caption{{\mead}. Each cell corresponds to attacks simultaneously executed on the targeted classifier. 
Attacks created using all the losses in~\cref{sec:adv_problem} are marked with $^\star$.
Attacks such as SA and DF are not dependent on the choice for the loss, but are equally considered as part of the multi-armed framework.
Empty cells correspond to combinations of perturbation magnitude and norm constraint that are usually not considered in the literature.}
\ra{1.3}
\resizebox{\columnwidth}{!}{%
\begin{tabular}{@{}rd|d|d|d@{}}
\multicolumn{1}{c}{} & \multicolumn{1}{c}{L$_1$} & \multicolumn{1}{c}{L$_2$} & \multicolumn{1}{c}{L$_\infty$} & \multicolumn{1}{c}{No norm} \\ \cmidrule{2-5}
$\varepsilon=0.01$ & 
- & CW2 & - & -\\ \cmidrule{2-5}
$\varepsilon=0.03125$ & 
- & - & PGDi$^{\star}$,FGSM$^{\star}$,BIM$^{\star}$ & -\\ \cmidrule{2-5}
$\varepsilon=0.0625$ & 
- & - & PGDi$^{\star}$,FGSM$^{\star}$,BIM$^{\star}$ & -\\\cmidrule{2-5}
$\varepsilon=0.1$ & 
- & HOP & - & -\\ \cmidrule{2-5}
$\varepsilon=0.125$ & 
- & PGD2$^{\star}$& PGDi$^{\star}$,FGSM$^{\star}$,BIM$^{\star}$,SA & -\\ \cmidrule{2-5}
$\varepsilon=0.25$ &
- & PGD2$^{\star}$ & PGDi$^{\star}$,FGSM$^{\star}$,BIM$^{\star}$ & -\\\cmidrule{2-5}
$\varepsilon=0.3125$ & 
- & PGD2$^{\star}$& PGDi$^{\star}$,FGSM$^{\star}$,BIM$^{\star}$,CWi & -\\ \cmidrule{2-5}
$\varepsilon=0.5$ & 
- & PGD2$^{\star}$& PGDi$^{\star}$,FGSM$^{\star}$,BIM$^{\star}$ & -\\ \cmidrule{2-5}
$\varepsilon=1$ & 
- & PGD2$^{\star}$& - & -\\ \cmidrule{2-5}
$\varepsilon=1.5$ & 
- & PGD2$^{\star}$& - & -\\ \cmidrule{2-5}
$\varepsilon=2$ & 
- & PGD2$^{\star}$& - & -\\ \cmidrule{2-5}
$\varepsilon=5$ & 
PGD1$^{\star}$& - & - & -\\ \cmidrule{2-5}
$\varepsilon=10$ &
PGD1$^{\star}$& - & - & -\\ \cmidrule{2-5}
$\varepsilon=15$ & 
PGD1$^{\star}$& - & - & -\\ \cmidrule{2-5}
$\varepsilon=20$ & 
PGD1$^{\star}$& - & - & -\\ \cmidrule{2-5}
$\varepsilon=25$ & 
PGD1$^{\star}$& - & - & -\\ \cmidrule{2-5}
$\varepsilon=30$ &
PGD1$^{\star}$& - & - & -\\ \cmidrule{2-5}
$\varepsilon=40$ & 
PGD1$^{\star}$& - & - & -\\ 
\midrule
No $\varepsilon$ & 
- & DF & - & -\\ \cmidrule{2-5}



\begin{tabular}{r}
     max. rotation $=30$\\
     max. translation $=8$\\
\end{tabular} &
- & - & - & STA\\
\end{tabular}
}
\label{tab:attacks}
\end{table}

We test our proposed solution by deploying it against the multi-armed adversarial attacks framework introduced in~\cite{GranesePRMP2022ECMLPKDD}, and by evaluating its detection performance. The source code to reproduce our results can be found in the Supplementary Material.

In our empirical evaluation, we assume that a third party provides us with four simple supervised detectors. Each of them is trained to detect a single specific kind of attack. 
This is a reasonable assumption, as many methods in the literature are able to successfully detect at least one type of attack and fail at detecting others. 
In addition, to emphasize the role played by the proposed method, these detectors are merely shallow networks (3 fully-connected layers with 256 nodes each), which are only allowed to observe the logits of the target classifier to distinguish between natural and adversarial samples. 
Due to their specifics, these individual shallow detectors are bound to perform very poorly, i.e. much worse than SOTA detectors, against attacks they have not been trained on, as shown in~\cref{fig:box_plot}. This aspect enhances the value of our solution, which attains favorable performance by aggregating detectors that individually exhibit subpar performance w.r.t. SOTA adversarial examples detection methods.  

\subsection{Evaluation framework}
\label{sec:eval_framework}
\noindent\textbf{Evaluation setup: {\mead}}.
We consider all the attack algorithms mentioned in {\mead}~\cite{GranesePRMP2022ECMLPKDD}, 
and we group them by the corresponding norm and the perturbation magnitude. For each natural sample and each gradient-based attack algorithm (i.e., FGSM, PGD or BIM), we create four adversarial examples, each corresponding to one of the loss functions described in~\cref{sec:adv_problem}. 
\Cref{tab:attacks} reports all the attacks in the multi-armed setting. Each cell corresponds to a group of attacks crafted according to the algorithm (reported in the cell), the associated norm (indicated by the column label) and perturbation magnitude (indicated by the row label) and one of the considered four loss functions.
Thus, for example, when we consider L$_\infty$ norm and $\varepsilon=0.125$, the detector is evaluated on $4 + 4 + 4 + 1 = 13$ simultaneous adversarial attacks. Note that we discard the perturbed examples that do not fool the classifier as, by definition, they are neither natural nor adversarial.\\\\
\noindent\textbf{Evaluation metrics}.
Following the evaluation setup described above, for each sample and for each group of attacks corresponding to each cell in 
~\cref{tab:attacks} we consider a detection successful, i.e. a true positive, if and only if all the adversarial attacks are detected. 
Otherwise, we report a false negative. 
We use the classical definitions of \textit{true negative} and \textit{false positive} for the natural samples detection. This means that a true negative is a natural sample detected as natural, and a false positive is a natural sample detected as adversarial.
We measure the performance of the detectors in terms of $i)$ \underline{\auc}~\cite{davis2006relationship} (the \textit{Area Under the Receiver Operating Characteristic curve}) which represents the ability of the detector to discriminate between adversarial and natural examples (higher is better); $ii)$ \underline{FPR at 95 \% TPR (\fpr)}, i.e., the percentage of natural examples detected as adversarial when 95 \% of the adversarial examples are detected (lower is better).\\\\
\noindent\textbf{Datasets and pre-trained classifiers}.
\label{par:dataset_classifier}
We run our experiments on CIFAR10~\cite{Cifar} and SVHN~\cite{SVHN} image datasets. For both, the pre-trained target classifier is a ResNet-18 models that has been trained for $100$ epochs, using SGD optimizer with a learning rate equal to $0.1$, weight decay equal to $10^{-5}$, and momentum equal to $0.9$. The accuracy achieved by the classifiers on the original clean data is 99\% for CIFAR10 and 100\% for SVHN over the train split; 93.3\% for CIFAR10 and 95.5\% for SVHN over the test split.\\\\
\noindent\textbf{Detectors}.
\label{par:detectors}
The proposed method aggregates four simple pre-trained detectors.
The detectors are four fully-connected neural networks, composed of 3 layers of 256 nodes each. All the detectors are trained for 100 epochs, using SGD optimizer with learning rate of 0.01 and weight decay 0.0005. They are trained to distinguish between natural and adversarial examples created according to the PGD algorithm, under L$_\infty$ norm constraint and perturbation magnitude $\varepsilon=0.125$ for CIFAR10 and $\varepsilon=0.25$ for SVHN. Each detector is trained on natural and adversarial examples generated using one of the loss
functions mentioned in~\cref{sec:adv_problem} (i.e., ACE~\cref{eq:ACE}, KL~\cref{eq:KL}, FR~\cref{eq:FR}, or Gini~\cref{eq:Gini}) to craft its adversarial training samples. We want to point out that the purpose of this paper is not creating a new supervised detector, but rather to show a method to aggregate a set of pre-trained detectors. Moreover, it is important to notice that either supervised and unsupervised methods can be added to or pool of experts (cf.~\cref{app:sota}), provided that they output a confidence on the input sample being or not an adversarial example. We further expand on the selection of the $\varepsilon$ parameter of the adversarial examples used at training time in~\cref{app:varius_eps} (cf.~\cref{tab:cifar10_salad,tab:svhn_salad}).\\\\
\noindent\textbf{NSS~\cite{NSS}}. 
\label{sec:nss}
We compare the proposed method with NSS, which is the best among the supervised SOTA methods against multi-armed adversarial attacks (cf.~\cite{GranesePRMP2022ECMLPKDD}).
NSS characterizes the adversarial perturbations through the use of \textit{natural scene statistics}, i.e., statistical properties that can be altered by the presence of adversarial perturbations. 
NSS is trained by using PGD algorithm, L$_\infty$ norm constraint and perturbation magnitude $\varepsilon=0.03125$ for CIFAR10 and $\varepsilon=0.0625$ for SVHN. We further expand on the selection of the $\varepsilon$ parameter of the adversarial examples used at training time in~\cref{tab:cifar10_nss,tab:svhn_nss,app:varius_eps}. \\\\ 
\noindent\textbf{On the optimization of~\cref{eq:minimaxProb4}}. For the optimization of~\cref{eq:minimaxProb4}, we rely on the \texttt{SciPy}~\cite{2020SciPy-NMeth} library, the \texttt{optimize} package, and the \texttt{minimize} function which uses the \textit{Sequential Least Squares Programming} (SLSQP) algorithm to find the optimum. Further details can be found in~\cref{app:optimization}.\\

\subsection{Discussion}
\begin{figure*}[t]
	\centering
	\begin{subfigure}[b]{ .8\columnwidth}
	    \centering
	    \includegraphics[width=\columnwidth]{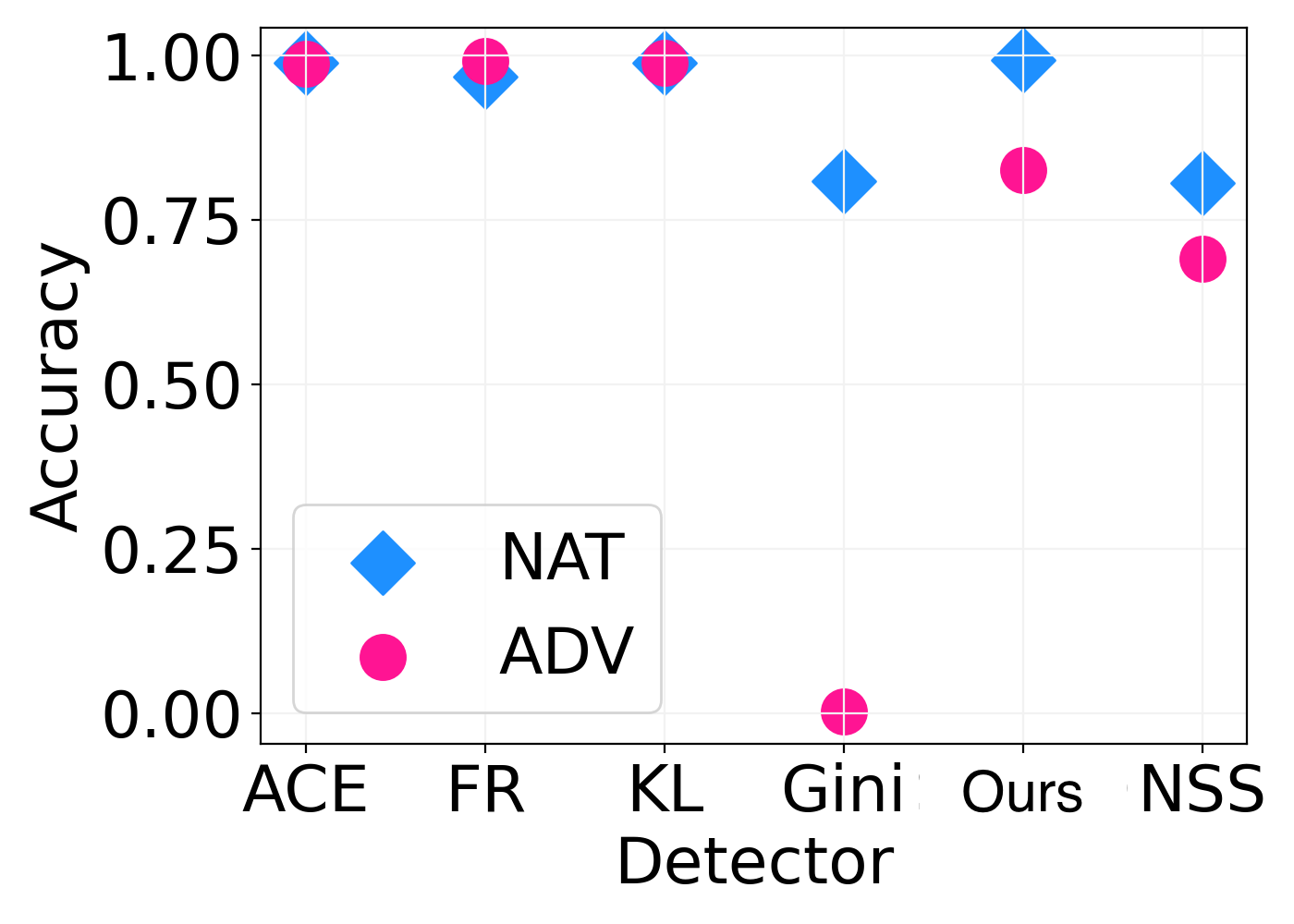}
	    \caption{Attacks crafted with PGD algorithm, the FR loss, $\varepsilon=40$, and norm constraint L$_1$}
	    \label{fig:acc_pgd1}
	\end{subfigure}
\hspace{50pt}
		\begin{subfigure}[b]{ .8\columnwidth}
		\centering
		\includegraphics[width=\columnwidth]{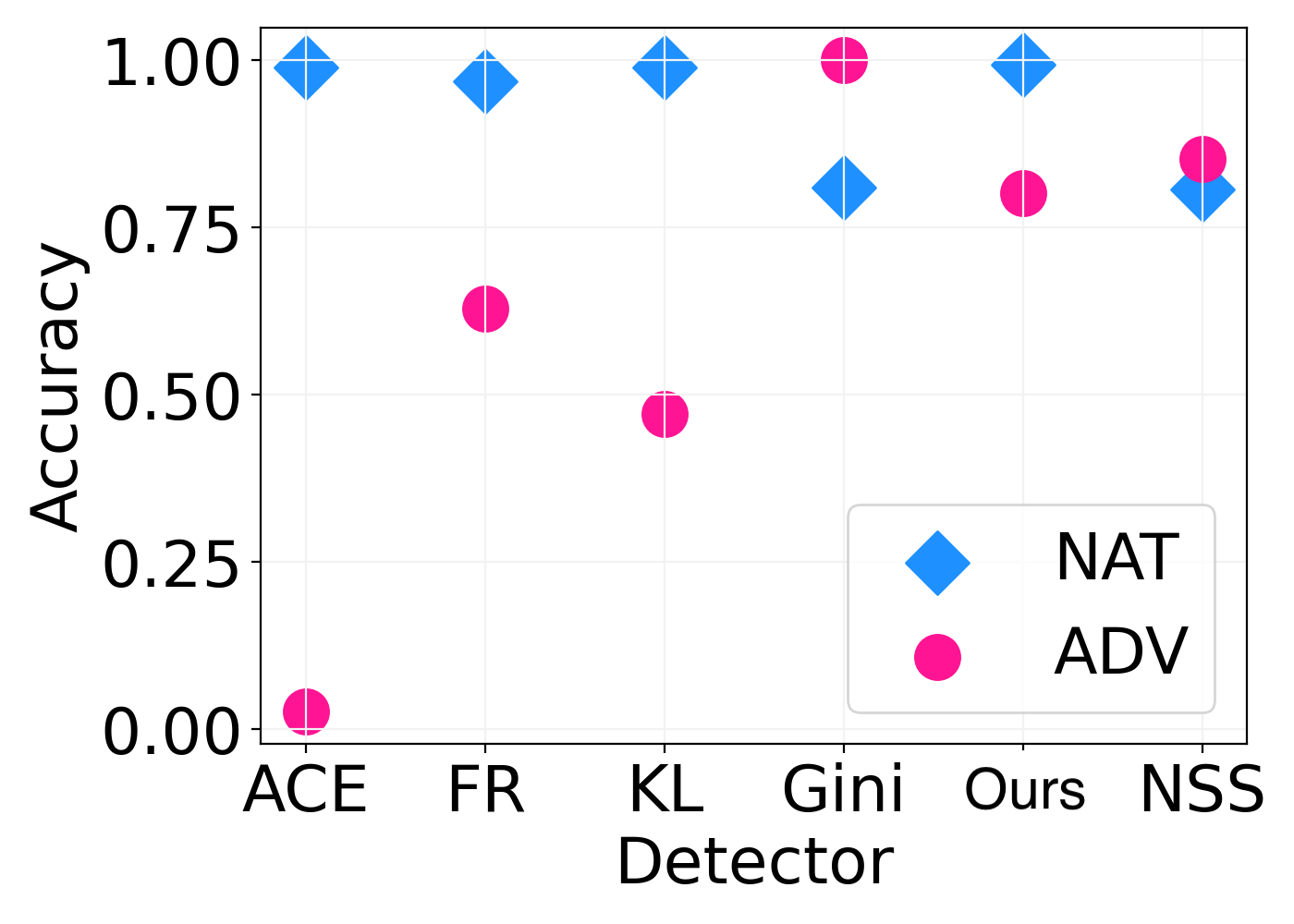}
		\caption{Attacks crafted with FGSM algorithm, the FR loss, $\varepsilon=40$, and norm constraint L$_\infty$}
		\label{fig:acc_fgsm}
	\end{subfigure}

	\begin{subfigure}[b]{ .8\columnwidth}
		\centering
		\includegraphics[width=\columnwidth]{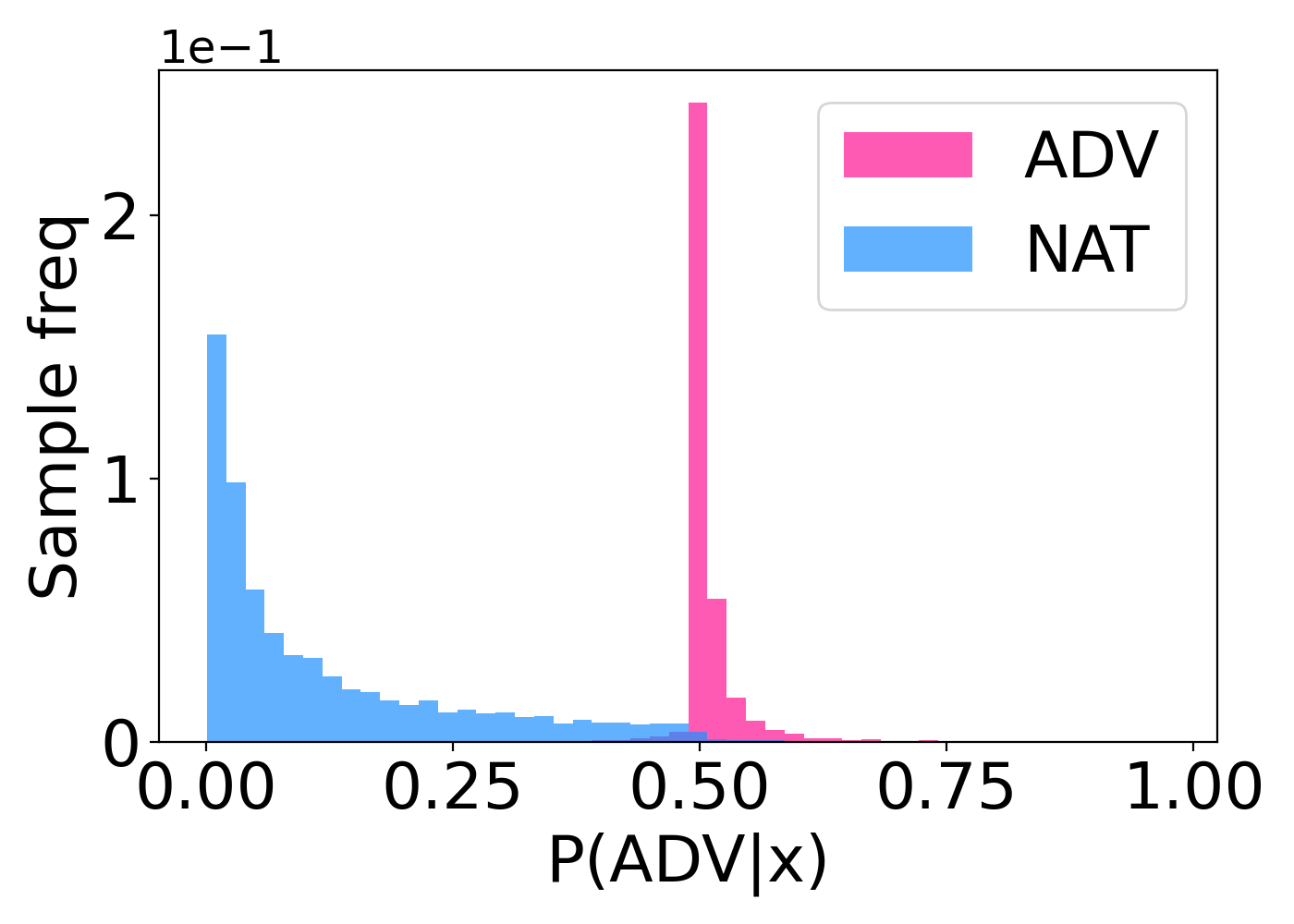}
		\caption{Ours against attacks crafted with PGD algorithm, the FR loss, $\varepsilon=40$, and norm constraint L$_1$}
		\label{fig:hist_salad}
	\end{subfigure}
\hspace{50pt}
	\begin{subfigure}[b]{ .8\columnwidth}
		\centering
		\includegraphics[width=\columnwidth]{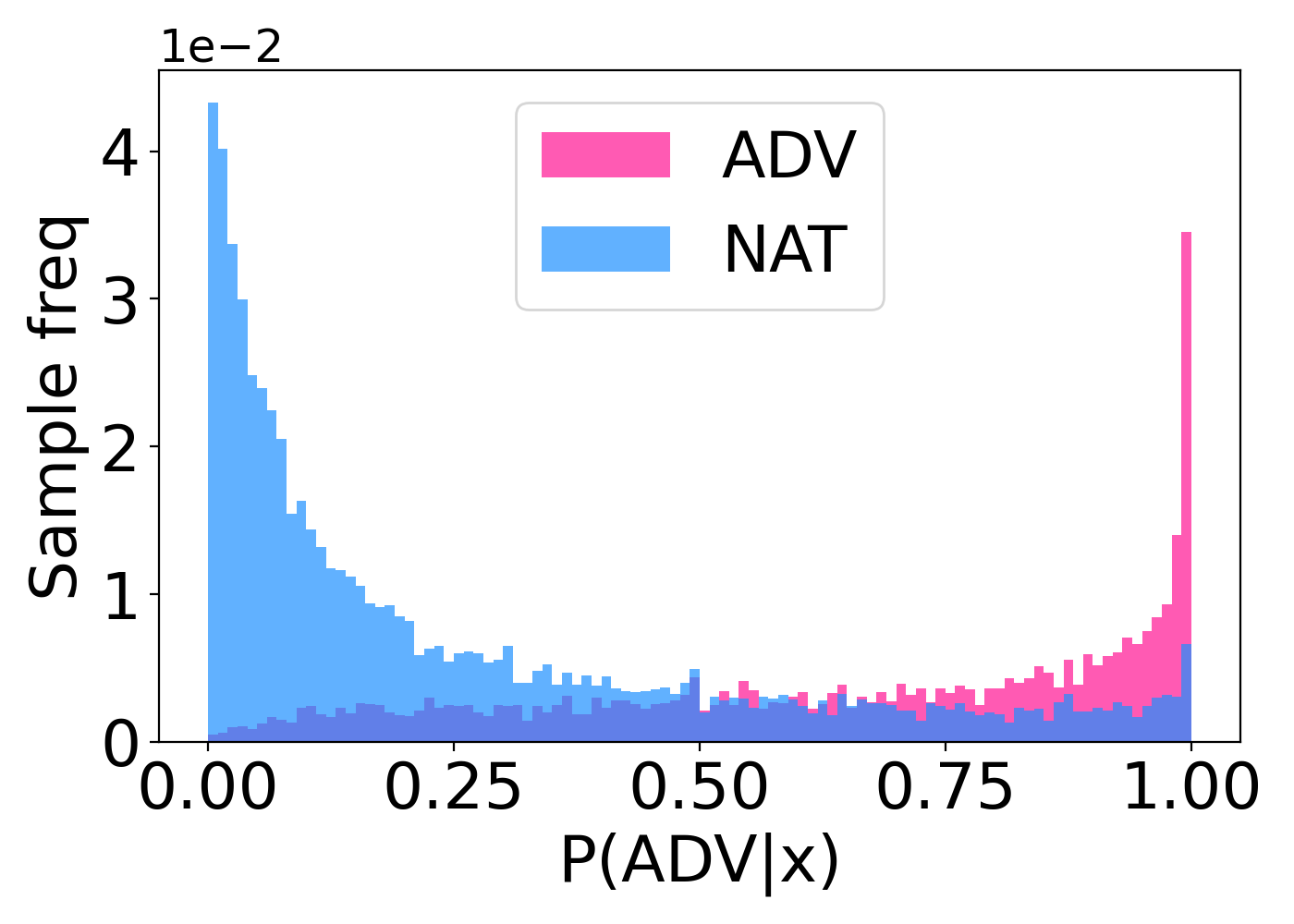}
		\caption{NSS against attacks crafted with PGD algorithm, the FR loss, $\varepsilon=40$, and norm constraint L$_1$}
		\label{fig:hist_nss}
	\end{subfigure}
	\caption{Discrimination performances. 
	In~\cref{fig:acc_pgd1} and~\cref{fig:acc_fgsm}, the accuracies of the detectors on natural and adversarial examples; in~\cref{fig:hist_salad} and~\cref{fig:hist_nss} we show how the proposed method and NSS split the data samples. We report the results for the detection of adversarial examples in pink, and the results for the detection of natural examples in blue.}
	\label{fig:evaluation}
\end{figure*}
\begin{table*}[!htbp]
\centering
\caption{Comparison between the proposed method and NSS on CIFAR10 and SVHN. The $^\star$ symbol means the perturbation mechanism is executed in parallel four times starting from the same original clean sample, each time using one of the objective losses between ACE~\cref{eq:ACE}, KL~\cref{eq:KL}, FR~\cref{eq:FR}, Gini~\cref{eq:Gini}.}
\ra{1.3}
\resizebox{2\columnwidth}{!}{%
\begin{tabular}{@{}r|bbcbbcyycyy@{}}\toprule
& \multicolumn{5}{c}{CIFAR10} & & \multicolumn{5}{c}{SVHN} \\
\cmidrule{2-6} \cmidrule{8-12}
& \multicolumn{2}{c}{NSS} & \phantom{abc}& \multicolumn{2}{c}{Ours} &
  \phantom{abc} & \multicolumn{2}{c}{NSS} & \phantom{abc}& \multicolumn{2}{c}{Ours} \\ \cmidrule{2-3}
\cmidrule{5-6} \cmidrule{8-9} \cmidrule{11-12}
  & \auc & \fpr  && \auc & \fpr && \auc & \fpr && \auc & \fpr\\ 
 \midrule

\textbf{Norm L$_1$}\\

\underline{PGD1$^\star$}\\
$\varepsilon=5$ & 
48.5 & 94.2 && 
\textbf{62.1} & \textbf{87.1} &&
40.2 & 91.3 && 
\textbf{76.9} & \textbf{79.0}\\
$\varepsilon=10$ &
54.0 & \textbf{90.3} &&
\textbf{56.8} & 90.6 &&
36.9 & 91.3 &&
\textbf{73.0} & \textbf{82.5} \\
$\varepsilon=15$ & 
58.8 & 86.8 &&
\textbf{69.3} & \textbf{84.4} &&
35.6 & 91.3&&
\textbf{78.9} & \textbf{72.5} \\
$\varepsilon=20$ & 
63.5 & 82.3 && 
\textbf{78.7} & \textbf{73.1} &&
36.1 & 91.3 &&
\textbf{83.6} & \textbf{60.7}\\
$\varepsilon=25$ & 
67.7 & 77.2 && 
\textbf{87.1} & \textbf{50.8} &&
37.8 & 91.3 &&
\textbf{87.0} & \textbf{48.6}\\
$\varepsilon=30$ &
71.4 & 73.4 &&
\textbf{90.3} & \textbf{35.4} &&
39.8 & 91.3 &&
\textbf{89.3} & \textbf{37.2}\\
$\varepsilon=40$ & 
76.1 & 67.3 && 
\textbf{92.1} & \textbf{26.4} &&
43.1 & 91.3 &&
\textbf{92.6} & \textbf{20.0}\\
\midrule

\textbf{Norm L$_2$}\\

\underline{PGD2$^\star$}\\
$\varepsilon=0.125$ & 
48.3 & 94.3 &&
\textbf{63.9} & \textbf{85.4} &&
40.8 & 91.3 &&
\textbf{80.2} & \textbf{74.5}\\
$\varepsilon=0.25$ & 
53.2 & 91.2 && 
\textbf{57.1} & \textbf{90.5} &&
37.2 & 91.3 &&
\textbf{74.0} & \textbf{81.7}\\
$\varepsilon=0.3125$ & 
55.8 & 89.2 && 
\textbf{61.0} & \textbf{88.9} &&
36.1 & 91.3 &&
\textbf{75.2} & \textbf{79.4}\\
$\varepsilon=0.5$ & 
63.3 & 82.6 && 
\textbf{79.4} & \textbf{73.2}&&
35.9 & 91.3 &&
\textbf{82.5} &\textbf{64.4}\\
$\varepsilon=1$ & 
76.4 & 67.5 && 
\textbf{91.4} & \textbf{26.4}&&
42.5 & 91.3 &&
\textbf{92.3} & \textbf{24.7}\\
$\varepsilon=1.5$ & 
81.0 & 63.0 && 
\textbf{91.9} & \textbf{24.2} &&
46.3 & 91.3 &&
\textbf{94.1} & \textbf{7.5}\\
$\varepsilon=2$ & 
82.6 & 62.3 &&
\textbf{91.9} & \textbf{24.1} &&
49.8 & 91.3 &&
\textbf{94.9} & \textbf{5.3}\\

\underline{DeepFool}\\
No $\varepsilon$ & 
57.0 & 91.7 &&
\textbf{81.9} & \textbf{54.8}&&
41.3 & 91.3 &&
\textbf{94.9} & \textbf{12.0}\\

\underline{CW2}\\
$\varepsilon=0.01$ & 
\textbf{56.4} & \textbf{90.8} &&
53.4 & 92.2&&
41.0 & \textbf{91.3} &&
\textbf{54.2} & 92.0\\

\underline{HOP}\\
$\varepsilon=0.1$ & 
66.1 & 87.0 &&
\textbf{86.1} & \textbf{49.1}&&
67.6 & 84.2 &&
\textbf{96.0} & \textbf{10.2}\\
\midrule

\textbf{Norm L$_\infty$}\\
\underline{PGDi$^\star$, FGSM$^\star$, BIM$^\star$}\\\
$\varepsilon=0.03125$ & 
\textbf{83.0} & \textbf{55.3} &&
82.3 & 59.7&&
\textbf{86.3} & \textbf{46.9} &&
81.4 & 64.9\\
$\varepsilon=0.0625$ & 
\textbf{96.0} & \textbf{17.2} &&
92.0 & 29.6 &&
88.9 & \textbf{0.7} &&
\textbf{89.1} & 33.3\\
$\varepsilon=0.25$ &
\textbf{97.3} & \textbf{0.6} &&
95.9 & 8.8 &&
51.6 & 88.9 &&
\textbf{92.3} & \textbf{16.4}\\
$\varepsilon=0.5$ & 
82.5 & 100.0 &&
\textbf{94.6} & \textbf{9.7} &&
46.7 & 86.7 &&
\textbf{92.9} & \textbf{14.4}\\

\underline{PGDi$^\star$, FGSM$^\star$, BIM$^\star$, SA}\\
$\varepsilon=0.125$ & 
9.4 & 99.9 &&
\textbf{88.9} & \textbf{40.8} &&
32.9 & 91.3 && 
\textbf{89.2} & \textbf{29.1} \\

\underline{PGDi$^\star$, FGSM$^\star$, BIM$^\star$, CWi}\\
$\varepsilon=0.3125$ &
63.2 & 99.1 &&
\textbf{80.0} & \textbf{61.1} &&
41.3 & 91.3 &&
\textbf{88.2} & \textbf{33.1}\\
\midrule
\textbf{No norm}\\
\underline{STA}\\
No $\varepsilon$ & 
\textbf{88.5} & \textbf{38.8} && 
82.7 & 52.4 &&
\textbf{91.2} & \textbf{0.2} &&
90.2 & 23.2\\
\bottomrule
\end{tabular}
}
\label{tab:final_table}
\end{table*}

\begin{table}[!htbp]
\centering
\caption{Comparison between Ours and Ours+NSS on CIFAR10. The $^\star$ symbol means the perturbation mechanism is executed in parallel four times starting from the same original clean sample, each time using one of the objective losses between ACE~\cref{eq:ACE}, KL~\cref{eq:KL}, FR~\cref{eq:FR}, Gini~\cref{eq:Gini}. We focus only in the cases in which the proposed method is outperformed  from the corresponding competitors.}
    \resizebox{\columnwidth}{!}{%
    \begin{tabular}{r|bbcbb}
    \toprule
    & \multicolumn{5}{c}{CIFAR10}\\
    \cmidrule{2-6} 
    & \multicolumn{2}{c}{Ours} & \phantom{abc}& \multicolumn{2}{c}{Ours+NSS} \\ \cmidrule{2-3}
    \cmidrule{5-6} 
      & \auc & \fpr  && \auc & \fpr \\ 
    \midrule
    
    \textbf{Norm L$_2$}\\
    
    \underline{CW2}\\
    $\varepsilon=0.01$ & 
    53.4 & 92.2&&
    \textbf{54.1} & \textbf{91.3}\\
    \midrule
    
    \textbf{Norm L$_\infty$}\\
    \underline{PGDi$^\star$, FGSM$^\star$, BIM$^\star$}\\\
    $\varepsilon=0.03125$ & 
    82.3 & 59.7&&
    \textbf{89.9} & \textbf{34.4}\\
    $\varepsilon=0.0625$ & 
    92.0 & 29.6 &&
    \textbf{96.4} & \textbf{9.0}\\
    $\varepsilon=0.25$ &
    95.9 & 8.8 &&
    \textbf{96.7} & \textbf{3.5}\\
    \midrule
    \textbf{No norm}\\
    \underline{STA}\\
    No $\varepsilon$ & 
    82.7 & 52.4 &&
    \textbf{87.3} & \textbf{35.4}\\
    \bottomrule
    \end{tabular}
    }
    \label{tab:salad_and_nss}
    
    
    
    
\end{table}

We now present the main experimental results to show the effectiveness of the proposed aggregation method for adversarial attack detection. Further discussion on these results, as well as additional experiments can be found in~\cref{app:experiments}.\\
\subsubsection{The \textit{shallow} detectors}
\label{sec:discussion}
\cref{fig:evaluation,fig:box_plot,fig:eps_norm} provides a graphical interpretation of the detection performance when ResNet18, trained on CIFAR10, is the target classifier. The single detectors are named after the loss function used to craft the adversarial examples on which each detector is trained along with the natural samples. The main takeaway from~\cref{fig:box_plot} is the observation that, when considered individually, the shallow detectors are clearly subpar w.r.t. state of the art adversarial attacks detection mechanism. On the contrary, the aggregation provided by our method results in detection performance that are comparable to SOTA performance and, in some cases, outperform well established detection mechanisms. 

\Cref{fig:eps_norm} sheds light on the fact that the mixture of experts attained by our proposed method can consistently improve the detection of adversarial examples over several multi-armed attacks mounted using different norms and perturbation magnitudes.

\textbf{One main takeaway of this paper is that, if we are provided with generally non-robust detectors whose performance is good only against a limited amount of attacks (as it is confirmed by~\cref{fig:box_plot,fig:eps_norm}), we can successfully aggregate them through the proposed method to obtain a consistently better detection.}\\

In~\cref{fig:evaluation} we consider attacks crafted according to the PGD algorithm, the FR loss, $\varepsilon=40$, and norm constraint L$_1$ (cf.~\cref{fig:acc_pgd1,fig:hist_salad,fig:hist_nss}), and attacks crafted according to the FGSM algorithm, FR loss, $\varepsilon=0.5$, and L$_\infty$ norm in~\cref{fig:acc_fgsm}.
We also report the performance of the considered detectors in terms of detection accuracy over the natural examples in blue and the adversarial examples in pink. 
As we can observe, the individual detectors, which are named after the loss functions ACE, FR, KL, and Gini, exhibit different behaviors for the specific attack. In~\cref {fig:acc_pgd1}, the Gini detector drastically fails at detecting the attack as its accuracy plummets to 0\% on the adversarial examples. 
In the same way, the FR and KL detectors but mostly the ACE detector, perform poorly against FGSM (cf.~\cref{fig:acc_fgsm}). On the contrary,
our method, benefiting from the aggregation, obtains favorable results in both cases, confirming what we had previously observed. \\

The histograms in~\cref{fig:hist_salad,fig:hist_nss} show how the method we propose and NSS separate natural (blue) and adversarial examples (pink), respectively. The values along the horizontal axis represent the probability of being classified as adversarial, and the vertical axis represents the frequency of the samples within the bins.
The detection error is proportional to the area of overlap between the blue and the pink histograms.~\cref{fig:hist_salad} and~\cref{fig:hist_nss} suggest that the proposed method achieves lower detection error on the considered attack, as it is confirmed in~\cref{tab:final_table} where our proposed method attains 92.1 {\auc}, while NSS only achieves 76.1 {\auc} and. Additional plots are provided in~\cref{app:additional_plot}.

In particular, the performance attained by the proposed method is consistent across the larges part of the considered multi-armed adversarial attacks, as confirmed in ~\cref{tab:final_table,fig:box_plot}.\\

\subsubsection{Evaluation of the proposed aggregator in {\mead}}
On \underline{CIFAR10}, our aggregator achieves maximum AUROC improvement w.r.t. NSS is 79.5 percentage points and happens for attacks under $L_{\infty}$-norm constraint, $\varepsilon = 0.125$ and PGD$^\star$, FGSM$^\star$, BIM$^\star$, SA, i.e. when as many as 13 different simultaneous adversarial attacks are mounted. 
Similarly, for our proposed method the maximum attained FPR at 95\% TPR improvement w.r.t. NSS is 90.3 percentage points and happens for attacks under $L_{\infty}$-norm constraint, $\varepsilon = 0.5$ and PGD$^\star$, FGSM$^\star$, BIM$^\star$, i.e., when as many as 12 different simultaneous adversarial attacks are mounted.  
Our aggregator outperforms NSS in the case of the attacks with L$_1$ and L$_2$ norm, regardless of the algorithm or the perturbation magnitude, and in the case of L$_{\infty}$ norm with large perturbations.
However, for the attacks with L$_\infty$ norm and small $\varepsilon$, although the proposed method's performance is comparable to that of NSS, we notice a slight degradation. To shed light on this, we remind that individual detectors aggregated are based on the classifier's logits; NSS, on the other hand, extracts natural scene statistics from the inputs. This more sophisticated technique makes NSS perform well when tested on attacks with similar $\varepsilon$ and the same norm as the ones seen at training time. Similar conclusions can be drawn for the results on \underline{SVHN} (cf.~\cref{tab:final_table}).

\cref{tab:salad_and_nss} shows the modularity of the proposed method when SOTA detection methods, NSS (a) and FS (b), are plugged in as a fifth detector. We test Ours+NSS on the attacks on which our aggregator was outperformed by the competitors.
In all the cases, Ours+NSS outperforms ``Ours'' either in terms of AUROC and FPR. In most 
of the cases, Ours+NSS is also better than the individual competitor.
In~\cref{app:sota} we provide further insights on this by showing that the same behavior is observed when we plug a SOTA unsupervised method as fifth detector in our pool. 
\subsubsection{Evaluation of the proposed aggregator in the non-simultaneous setting}
\begin{table*}[!htbp]
\centering
\caption{The proposed method and NSS in the non-simultaneous setting. The column names ACE, KL, FR, and Gini denote the loss function used to craft the attacks. HOP, DeepFool, CW2, and STA attacks have already been considered individually in ~\cref{tab:final_table}.}
\ra{1.3}
\resizebox{2\columnwidth}{!}{%
\begin{tabular}{@{}r|bcbcbcb@{}}
\toprule
& \multicolumn{7}{c}{CIFAR10}\\
\cline{2-8}
& \multicolumn{7}{c}{Ours {\auc}~({\fpr})~~--~~NSS {\auc}~({\fpr})}\\
\cline{2-8}
& \multicolumn{1}{c}{ACE} & \phantom{abc} & \multicolumn{1}{c}{KL} & \phantom{abc} & \multicolumn{1}{c}{FR} & \phantom{abc} & \multicolumn{1}{c}{Gini}\\
\cline{2-2}\cline{4-4}\cline{6-6}\cline{8-8}
\underline{PGD1}\\
 $\varepsilon =$ 5 & 
\textbf{ 66.2 (83.6)}  -- 49.9 (93.5)  &&
\textbf{ 64.2 (85.7)} -- 49.6 (93.0)  &&
\textbf{ 63.0 (87.1)} -- 49.9 (93.3)  &&
\textbf{ 80.7 (58.4)} -- 50.3 (93.2) \\
 $\varepsilon =$ 10 &
\textbf{ 62.6 (87.5) }-- 56.9 (88.4) && 
\textbf{ 62.3 (88.2) }-- 56.6 (88.3) &&
\textbf{ 63.1 (86.5) }-- 57.0 (88.1) &&
\textbf{ 86.9 (46.0)} -- 57.1 (88.8) \\
 $\varepsilon =$ 15 &
 \textbf{74.2 (81.4)} -- 63.1 (83.0) &&
\textbf{ 75.2 (80.6)} -- 62.8 (83.1) &&
 \textbf{75.3 (79.4)} -- 63.2 (82.5) &&
\textbf{ 90.0 (31.1)}  -- 63.5 (84.0) \\
 $\varepsilon =$ 20 & 
 \textbf{86.8 (65.3)} -- 68.5 (77.1)  &&
 \textbf{87.5 (63.1)} -- 68.1 (77.3)  &&
 \textbf{86.9 (63.3)} -- 68.7 (76.4)  &&
 \textbf{91.7 (31.2)}  -- 69.9 (77.6) \\
 $\varepsilon =$ 25 &
 \textbf{93.9 (38.4) }-- 73.1 (71.1)  &&
 \textbf{94.3 (36.2)} -- 72.7 (71.8)  &&
 \textbf{93.7 (41.1)} -- 73.4 (70.9)  &&
 \textbf{92.3 (28.9)} -- 75.0 (71.4)  \\
 $\varepsilon =$ 30 & 
 \textbf{97.1 (12.3)} -- 77.1 (64.5)  &&
 \textbf{97.2 (12.6)} -- 76.8 (65.1)  &&
 \textbf{96.8 (15.9)} -- 77.4 (65.2)  &&
 \textbf{92.6 (27.9)} -- 78.6 (67.3)  \\
 $\varepsilon =$ 40 &
 \textbf{98.9 (1.0)} -- 83.5 (52.7) && 
 \textbf{99.0 (1.0)} -- 83.3 (53.5) && 
 \textbf{98.8 (1.0)} -- 83.6 (52.7) && 
 \textbf{92.7 (27.4)} -- 80.1 (64.9) \\
 \midrule
 \underline{PGD2}\\                                                 
 $\varepsilon =$ .125 & 
 \textbf{67.9 (81.1)} -- 49.5 (93.8) &&
 \textbf{65.4 (84.3}) -- 49.1 (93.5) &&
\textbf{ 63.9 (86.6)} -- 49.6 (93.5) &&
 \textbf{80.6 (58.4)} -- 49.5 (94.3) \\
 $\varepsilon =$ .25 &
 \textbf{62.3 (87.5)} -- 55.9 (89.1)  &&
 \textbf{62.1 (88.0)} -- 55.6 (89.2)  &&
 \textbf{62.6 (87.6)} -- 55.8 (89.4)  &&
 \textbf{86.7 (46.5)} -- 55.9 (89.8) \\
 $\varepsilon =$ .3125  &
 \textbf{66.5 (86.1) }-- 59.4 (86.5) &&
 \textbf{67.0 (85.9)} -- 59.0 (86.6) &&
 \textbf{67.8 (84.8)} -- 59.3 (86.6) &&
 \textbf{88.4 (42.2)} -- 59.3 (87.7) \\
 $\varepsilon =$ .5 &
 \textbf{86.4 (67.1)} -- 68.3 (77.4) &&
 \textbf{87.2 (64.5)} -- 68.0 (77.4) &&
 \textbf{86.7 (64.0)} -- 68.4 (77.2) && 
 \textbf{91.4 (31.4)} -- 69.0 (78.7) \\
 $\varepsilon =$ 1 & 
 \textbf{98.9 (0.9)} -- 84.4 (50.6) &&
 \textbf{98.9 (0.9)} -- 84.3 (50.5) && 
 \textbf{98.8 (0.9) }-- 84.7 (50.7) && 
 \textbf{92.5 (27.2)} -- 79.3 (66.8)\\
 $\varepsilon =$ 1.5 & 
 \textbf{99.2 (0.9)} -- 92.8 (28.7) &&
 \textbf{99.3 (0.9)} -- 92.7 (28.9) && 
 \textbf{99.3 (0.7)} -- 93.0 (27.3) && 
 \textbf{92.5 (27.2)} -- 79.5 (66.5) \\
 $\varepsilon =$ 2 &
 \textbf{99.3 (0.8)} -- 96.8 (13.9) &&
 \textbf{99.3 (0.8)} -- 96.9 (13.1) &&
 \textbf{99.3 (0.9)} -- 95.9 (17.2) &&
 \textbf{92.5 (27.2)} -- 79.5 (66.5) \\
 \midrule
 \underline{PGDi}\\               
 $\varepsilon =$ .03125 &
 \textbf{99.1 (0.9)} -- 92.3 (31.0) &&
 \textbf{99.1 (0.9)}  -- 92.1 (31.9) &&
 \textbf{99.0 (0.9)} -- 92.2 (30.7)  &&
 \textbf{94.8 (21.5)} -- 89.0 (44.0) \\
 $\varepsilon =$ .0625 &
 \textbf{99.3 (0.8)} -- 99.1 (3.3) &&
 \textbf{99.3 (0.8)} -- 99.1 (3.3) &&
 \textbf{99.3 (0.8)} -- 99.1 (3.6) &&
 97.4 \textbf{(8.0) }-- \textbf{98.1} (8.1)\\
 $\varepsilon =$ .125 &
 99.3 (0.7) -- \textbf{99.7 (0.6)} &&
 99.3 (0.9) -- \textbf{99.7 (0.6)} &&
 99.3 (0.8)  -- \textbf{99.6 (0.6)} && 
 97.3 (7.3) -- \textbf{99.6 (0.6)}\\
 $\varepsilon =$ .25 &
 99.3 (0.7) -- \textbf{99.7 (0.6)} && 
 99.3 (0.9) -- \textbf{99.7 (0.6)} && 
 99.3 (0.8)  -- \textbf{99.7 (0.6)} && 
 97.1 (7.3) -- \textbf{99.6 (0.6)}\\
 $\varepsilon =$ .3125 &
 99.3 (0.9) -- \textbf{99.7 (0.6)} &&
 99.3 (0.8) -- \textbf{99.7 (0.6)} &&
 99.3 (0.8) -- \textbf{99.7 (0.6)} &&
 97.1 (7.4) -- \textbf{99.7 (0.6)}\\
 $\varepsilon =$ .5 &
 99.3 (0.8) -- \textbf{99.7 (0.6)} &&
 99.3 (0.8) -- \textbf{99.7 (0.6)} &&
 99.3 (0.8)  -- \textbf{99.7 (0.6)} &&
 97.1 (7.3) -- \textbf{99.6 (0.6)}\\
 \midrule
 \underline{FGSM}\\
 $\varepsilon =$ .03125 &
 89.2 (47.5) -- \textbf{94.1 (26.7)} &&
 91.3 (40.6) -- \textbf{94.0 (27.0)} &&
 92.6 (34.1) -- \textbf{96.8 (15.0)} &&
 90.7 (42.7) -- \textbf{96.6 (15.3) }      \\
 $\varepsilon =$ .0625 &
 96.4 (18.5) -- \textbf{99.4 (1.3)} &&
 96.2 (18.7) -- \textbf{99.4 (1.4)} && 
 97.6 (10.3) -- \textbf{99.6 (0.6)} &&
 97.4 (11.9) -- \textbf{99.6 (0.6)} \\
 $\varepsilon =$ .125 &
 99.3 (3.4) -- \textbf{99.7 (0.6)} &&
 99.1 (4.3) -- \textbf{99.7 (0.6)} &&
 99.3 (2.5) -- \textbf{99.5 (0.6)} &&
 99.3 (2.4) -- \textbf{99.5 (0.6)}\\
 $\varepsilon =$ .25 &
 \textbf{99.8 (0.6)} -- 99.7 \textbf{(0.6)} &&
 \textbf{99.7 }(0.8) -- \textbf{99.7 (0.6)} &&
 \textbf{99.6} (1.1) -- 97.9 \textbf{(0.6)} &&
\textbf{ 99.6 }(1.1)  -- 97.7 \textbf{(0.6)} \\
 $\varepsilon =$ .3125 &
 \textbf{99.7 }(0.9) -- \textbf{99.7 (0.6)} &&
 \textbf{99.7 }(0.9) -- \textbf{99.7 (0.6)} &&
 \textbf{99.5} (1.5) -- 95.8 \textbf{(0.6) }&& 
 \textbf{99.5 }(1.5) -- 95.6 \textbf{(0.6)}\\
 $\varepsilon =$ .5 & 
 99.0 (4.9) -- \textbf{99.7 (0.6)} && 
 99.2 (2.7) -- \textbf{99.7 (0.6)} &&
 \textbf{99.2 (2.4)} -- 84.9 (100.0) &&
 \textbf{99.2 (2.4) }-- 84.8 (100.0) \\
 \midrule
 \underline{BIM}\\
 $\varepsilon =$ .03125 & 
 \textbf{98.3 (4.6)} -- 90.3 (37.7) &&
 \textbf{98.3 (4.4)} -- 90.2 (38.1) && 
 \textbf{97.8 (7.2)} -- 90.5 (37.0) &&
 \textbf{92.2 (32.6)} -- 88.2 (45.1) \\
 $\varepsilon =$ .0625 &
\textbf{ 99.4 (0.8)} -- 98.2 (7.5) &&
 \textbf{99.4 (0.9)} -- 98.2 (7.5) &&
 \textbf{99.4 (0.8) }-- 98.3 (7.3) &&
 96.6 (13.1) --\textbf{ 97.3 (12.9)}       \\
 $\varepsilon =$ .125 &
 99.3 (0.9) -- \textbf{99.6 (0.7)} &&
 99.3 (0.9) -- \textbf{99.7 (0.7)} && 
 99.3 (0.8) -- \textbf{99.6 (0.7)} && 
 97.8 (6.9) -- \textbf{99.3 (1.9)}\\
 $\varepsilon =$ .25 &
 99.3 (0.8) -- \textbf{99.7 (0.6)} &&
 99.3 (0.9) -- \textbf{99.7 (0.6)} &&
 99.3 (0.8) -- \textbf{99.7 (0.6)} &&
 97.4 (7.2) -- \textbf{99.6 (0.6)}\\
 $\varepsilon =$ .3125 &
 99.3 (0.9) -- \textbf{99.7 (0.6) }&&
 99.3 (0.8) -- \textbf{99.7 (0.6)} &&
 99.3 (0.9) -- \textbf{99.7 (0.6)} &&
 97.1 (7.4) -- \textbf{99.7 (0.6)}\\
 $\varepsilon =$ .5 &
 99.3 (0.8) -- \textbf{99.7 (0.6)} &&
 99.3 (0.8) -- \textbf{99.7 (0.6)} &&
 99.3 (0.8) -- \textbf{99.7 (0.6)} &&
 96.3 (7.3) -- \textbf{99.7 (0.6)}         \\
 \midrule
 \underline{SA} \\
 $\varepsilon =$ .125 &
 \textbf{91.2 (39.6)} -- 9.4 (99.9) &&
\textbf{ 91.2 (39.6)} -- 9.4 (99.9)  &&
\textbf{ 91.2 (39.6)} -- 9.4 (99.9) &&
\textbf{ 91.2 (39.6)} -- 9.4 (99.9)        \\
 \midrule
 \underline{CWi}\\
 $\varepsilon =$ .3125 & 
\textbf{ 80.7 (60.8) }-- 64.6 (89.8) &&
\textbf{ 80.7 (60.8)} -- 64.6 (89.8) &&
\textbf{ 80.7 (60.8)} -- 64.6 (89.8) &&
\textbf{ 80.7 (60.8)} -- 64.6 (89.8) \\
 \bottomrule
\end{tabular}
}
\label{tab:single_setting}
\end{table*}

In these experiments, we move from the simultaneous adversarial attack scenario to one where the different detectors are aggregated to detect one single attack at a time, as usually done in the literature. We report the complete results~\cref{tab:single_setting}. Crucially, these experiments show that ensemble detectors can also improve the performance for specific attacks. In particular, we would like to draw attention to the fact that we outperform NSS in the vast majority of the cases. Moreover, we achieve a maximum gain of 82.8 percentage points in terms of {\auc} (cf. SA attack) and 97.6 percentage points in terms of {\fpr} (cf. FGSM with $\varepsilon =0.5$ attack). On the other side, the competitor outperforms our proposed method only in a few cases, achieving a maximum gain of 5.9 percentage points in terms of {\auc} and 27.4 percentage points in terms of {\fpr} (cf. FGSM with $\varepsilon$=0.03125 attack in both the cases), and these gains are much lower than those obtained by the proposed method.\\

\section{Final remarks}
\label{sec:conclusion}
We introduced a new method to tackle the multi-armed adversarial attacks introduced in~\textsc{Mead}~\cite{GranesePRMP2022ECMLPKDD}.
We formalized the multi-armed attack detection problem as a minimax cross-entropy risk and derived a surrogate loss function. Based on this, we characterized our optimal soft-detector which results in a mixture of experts as the solution to a minimax problem. Our empirical results show that aggregating simple detectors using our method results in consistently improved detection performance. The achieved performance is comparable and in large set of cases better than the best state-of-the-art (SOTA) method in the multi-armed attack scenarios. Our method has two key benefits: it is modular, allowing existing and future methods to be integrated, and it is general, able to recognize adversarial examples from various attack algorithms and loss functions. Additionally, our aggregator can potentially be extended to aggregate both supervised and unsupervised SOTA adversarial detection methods.
 
As future work, it would be interesting to apply our detector aggregator to topics beyond simultaneous adversarial attack detection. As long as the detector outputs can be interpreted as a probability distribution across two categories, any existing or future supervised or unsupervised method can be combined using our proposed approach, making the aggregator a new ensemble technique. An example of this extension is intrusion detection, where an improved detection framework is highly desired, particularly with the use of ensemble learners~\cite{TL21}.

Limitations of the proposed method come from the fact it relies on a collection of detectors whose expertise is combined to obtain a more robust adversarial detection. Such models could be potentially poisoned by a malicious actor, drastically reducing the aggregator's reliability. We think this could have a potentially severe societal impact if the proposed method happened to be deployed with no additional checks on the quality of the available detectors.
\clearpage
\bibliography{biblio}
\bibliographystyle{ieeetr}
\clearpage
\appendix
 \subsection{On the optimization of~\cref{eq:minimaxProb4}}
\label{app:optimization}
The maximization problem in~\cref{eq:minimaxProb4} is well-posed  given that the mutual information is a concave function of $\omega\in\Omega$.
Although from the theoretical point of view,~\cref{eq:minimaxProb4} guarantees the optimal solution for the average regret minimization problem, in practice, we have to deal with some technical limitations.
For the optimization of ~\cref{eq:minimaxProb4}, we rely on the \texttt{SciPy}~\cite{2020SciPy-NMeth} library, package \texttt{optimize}, function \texttt{minimize}\footnote{Therefore we invert the sign of the objective function.} which uses the \textit{Sequential Least Squares Programming} (SLSQP) algorithm to find the optimum. This algorithm relies on local optimization  and is particularly straightforward when dealing with non-linear equations and equality and inequality constraints, as in our case. Overall, we obtained the satisfactory results provided in the paper by assigning default values to all the parameters and by setting a uniform distribution $[\omega_1, \omega_2, \omega_3, \omega_4]=[.25, .25, .25, .25]$ as the initial point in the solutions space.

Although these results are satisfactory and confirm the value of the sound theoretical framework we propose in~\cref{sec:math_framework}. 
We are well aware that, in some cases, as in~\cref{fig:acc_pgd1}, the proposed aggregation slightly underperforms in terms of accuracy w.r.t. the best detector in the set of allowed detectors. In this regard, we would like to raise a couple of points that are interesting for practitioners and possible future research:
\begin{enumerate}
    \item For each input sample, we solve one different optimization problem: although the algorithm above always reaches the end with a success state, given the finite amount of iterations and the tolerance which decides the stopping criterion, further sample-by-sample parameter optimization may be required. At this time, we have not delved into the problem, and we leave this for future research.
    \item The hard decisions made by the single detectors only depend on the $\argmax$ of their soft-probabilities. On the contrary,
    the optimization in~\cref{eq:minimaxProb4} considers the complete soft-probability distributions output by every single detector. Indeed, although the hard decision on two randomly considered samples can be right for both, often, the confidence in these decisions can be very different (i.e., two correctly classified samples may have utterly different associated soft probabilities).
    Further research on how differently accurate detectors influence the optimization in~\cref{eq:minimaxProb4} is left  for future work.
\end{enumerate}

\subsection{Supplementary Results of~\Cref{sec:experiments}}
\label{app:experiments}
In the following, we provide further discussions on the experiments in~\cref{sec:experiments} that have not been included in the main paper. \\
\subsubsection{Experimental environment}
\label{app:enviroment}
We run each experiment on a machine equipped with an Intel(R) Xeon(R) Gold 6226 CPU, 2.70GHz clock frequency, and a Tesla V100-SXM2-32GB GPU.
\subsubsection{Time measurements}
\begin{table}[!htb]
    \centering
\resizebox{\columnwidth}{!}{%
\begin{tabular}{r|c}
\toprule
     Training 1 single detector in our method &  \texttt{1h45m10s}\\
     Evaluating the optimization in our method & \texttt{1m35s} (for one attack)\\
     \midrule
     Training NSS & \texttt{3m30s}\\
     Evaluating NSS & \texttt{20s} (for one attack)\\
     \midrule
     \midrule

     On the largest set of simultaneous attacks (13
     attacks):\\
     
     Ours & \texttt{1m35s * 13 $\sim$ 21m}\\
     NSS & \texttt{20s * 13 $\sim$ 4m}\\
     \bottomrule
\end{tabular}
}
\label{tab:table_measurements}
\end{table}
\subsection{On the {\mead} framework}
\subsubsection{State-of-the-art (SOTA) detectors}
\label{app:sota}
\cite{GranesePRMP2022ECMLPKDD} suggests NSS~\cite{NSS} and FS~\cite{FS} as the most robust methods in the simultaneous attacks detection scheme (i.e., {\mead}). We remind that NSS is a supervised method that extracts the \textit{natural scene statistics} of the natural and adversarial examples to train a SVM. On the contrary,  FS is an unsupervised method that uses \textit{feature squeezing} (i.e.,
reducing the color depth of images and using smoothing to
reduce the variation among the pixels) to compare the model’s predictions.

\begin{figure}
    \centering
    \includegraphics[width=\columnwidth]{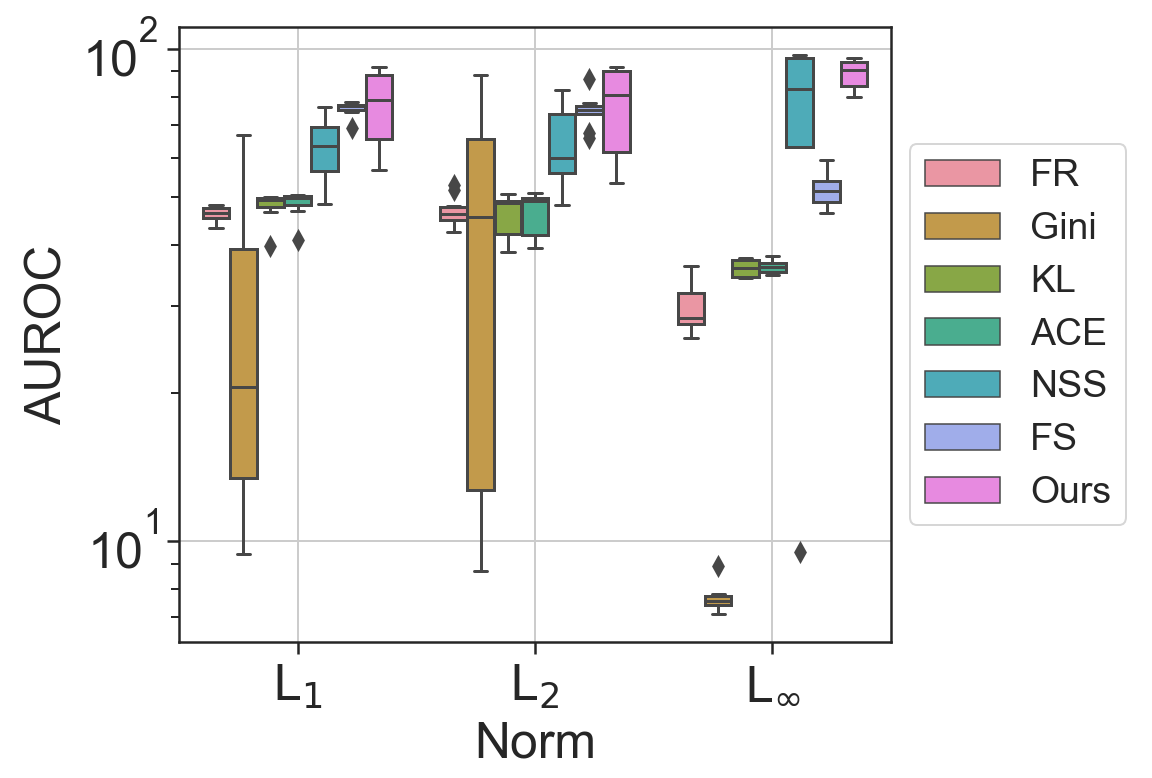}
    \caption{The \textit{shallow} detectors are named after the loss function used to craft the attacks they are trained to detect. Overall, the SOTA methods NSS and FS outperform all the individual shallow detectors. The aggregation we propose allows using the shallow models to attain a detector whose performance is consistently comparable and, in many cases, better than SOTA.}
    \label{fig:box_plot_fs}
\end{figure}

In particular, we choose NSS as a method to compare for multiple reasons:
\begin{enumerate}
    \item NSS achieves the best overall score in terms of {\auc} and {\fpr} among the SOTA against simultaneous attacks (cf. Tab. 3~\cite{GranesePRMP2022ECMLPKDD}).
    \item NSS achieves the best score in terms of {\auc} and {\fpr} under the L$_\infty$ norm where the biggest group of simultaneous attacks are evaluated (see~\cref{tab:attacks}). This is stressed in the plots in~\cref{fig:box_plot_fs}. Moreover, FS reaches better performance w.r.t. the proposed method only with PGD1 and PGD2 when the perturbation magnitude is small and in CW2. 
    \item The case study for our aggregator in the experimental section is based on supervised detectors as a consequence the comparison with a supervised detector was a natural choice.
\end{enumerate}
For the sake of completeness, the performances of NSS and FS under {\mead} are given in~\cref{fig:box_plot_fs}.
As shown before for Ours+NSS, in~\cref{tab:salad_and_fs} we propose an analysis of the performance of our method before and after adding the FS unsupervised detection mechanism to the pull of available detectors, showing a stark improvement in the latter case.
\subsubsection{Attacks}
We want to emphasize that, differently from the literature, we are the first to consider a defense mechanism against the simultaneous attack setting in which we detect attacks based on four different losses. More specifically, for each 'clean dataset' (in our case CIFAR10 and SVHN):
\begin{itemize}
    \item No. of adversarial examples generated with:
    \begin{itemize}
        \item L$_1$ norm: 7 (no. of $\varepsilon$) * 1 (PGD algorithm) * 4 (no. of losses) = 28 ('adversarial datasets')
        \item L$_2$ norm: 7 (no. of $\varepsilon$) * 1 (PGD algorithm) * 4 (no. of losses) + 3 (CW2, HOP, DeepFool) = 31 ('adversarial datasets')
        \item L$_\infty$ norm: 6 (no. of $\varepsilon$) * 3 (PGD, FGSM, BIM algorithms) * 4 (no. of losses) + 2 = 74 ('adversarial datasets')
        \item No norm: 1 ('adversarial dataset')
    \end{itemize}
    \item[$=>$] For a total of \texttt{28 + 31 + 74 + 1 = \textbf{134}} 'adversarial datasets' for each 'clean dataset'.
\end{itemize}
Moreover, it is interesting to notice that the experiments on CIFAR10 and SVHN represent a satisfying choice to show that state-of-the-art detection mechanisms struggle to maintain good performance when they are faced with the framework of simultaneous attacks. That said, we leave the evaluation of larger datasets as future work.\\
\subsubsection{Simulations adversarial attack according to different $\varepsilon$}
\label{app:varius_eps}
As discussed in~\cref{sec:experiments}, both NSS and the \textit{shallow} detectors aggregated via the proposed method are trained on natural and adversarial examples created with PGD algorithm and L$_\infty$ norm constraint. We show in~\cref{tab:cifar10_nss,tab:cifar10_salad,tab:svhn_nss,tab:svhn_salad} the results of the two methods according to $\varepsilon\in\left\{.03125, .0625,.125, .25, .3125, .5\right\}$. 
\begin{table}[]
    \centering
    \caption{Comparison between Ours and Ours+FS on CIFAR10. The $^\star$ symbol means the perturbation mechanism is executed in parallel four times starting from the same original clean sample, each time using one of the objective losses between ACE~\cref{eq:ACE}, KL~\cref{eq:KL}, FR~\cref{eq:FR}, Gini~\cref{eq:Gini}. We focus only in the cases in which the proposed method is outperformed  from the corresponding competitors.}
    \resizebox{\columnwidth}{!}{%
    \begin{tabular}{{@{}r|bbcbb@{}}}
    \toprule
    & \multicolumn{5}{c}{CIFAR10}\\
    \cmidrule{2-6} 
    & \multicolumn{2}{c}{Ours} & \phantom{abc}& \multicolumn{2}{c}{Ours+FS} \\ \cmidrule{2-3}
    \cmidrule{5-6} 
      & \auc & \fpr  && \auc & \fpr \\ 
     \midrule
    
    \textbf{Norm L$_1$}\\
    
    \underline{PGD1$^\star$}\\
    $\varepsilon=5$ & 
    62.1 & 87.1 &&
    \textbf{69.4} & \textbf{74.5}\\
    $\varepsilon=10$ &
    56.8 & 90.6 && 
    \textbf{76.8} & \textbf{64.5}\\
    $\varepsilon=15$ & 
    69.3 & 84.4 && 
    \textbf{77.6} & \textbf{60.3}\\
    \midrule
    
    \textbf{Norm L$_2$}\\
    
    \underline{PGD2$^\star$}\\
    $\varepsilon=0.125$ & 
    63.9 & 85.4 &&
    \textbf{67.9} & \textbf{76.4}\\
    $\varepsilon=0.25$ & 
    57.1 & 90.5 && 
    \textbf{76.0} & \textbf{64.7}\\
    $\varepsilon=0.3125$ & 
    61.0 & 88.9 && 
    \textbf{77.2} & \textbf{62.9}\\
    \underline{CW2}\\
    $\varepsilon=0.01$ & 
    53.4 & 92.2 && 
    \textbf{86.4} & \textbf{46.8}\\
    \bottomrule
    \end{tabular}
    }
    \label{tab:salad_and_fs}
\end{table}
\begin{table*}[!htbp]
\centering
\caption{Simultaneous attacks detection: NSS on CIFAR10. We train NSS on natural and adversarial examples created with PGD algorithm and L$_\infty$ norm constraint. The perturbation magnitude $\varepsilon$ is shown in the columns. We indicate in \textbf{bold} the best result.}
\resizebox{2\columnwidth}{!}{%
\begin{tabular}{r|cc|cc|cc|cc|cc|cc}
\toprule
 &
  \multicolumn{12}{c}{\textbf{NSS}} \Bstrut \\ \cline{2-13} &
   
  \multicolumn{2}{c|}{0.03125} & \multicolumn{2}{c|}{0.0625}& \multicolumn{2}{c|}{0.125}& \multicolumn{2}{c|}{0.25} &
  \multicolumn{2}{c|}{0.3125} &   \multicolumn{2}{c}{0.5}
  \Tstrut\Bstrut \\ \cmidrule{2-13} 
  &
  \multicolumn{1}{c}{\auc} &
  \fpr &
  \multicolumn{1}{c}{\auc} &
  \fpr &
  \multicolumn{1}{c}{\auc} &
  \fpr &
  \multicolumn{1}{c}{\auc} &
  \fpr &
  \multicolumn{1}{c}{\auc} &
  \fpr &
  \multicolumn{1}{c}{\auc} &
  \fpr \\
  \cmidrule{2-13}
  \textbf{Norm L$_1$}\\\textcolor{gray}{PGD1}\\
  $\varepsilon$ = 5 &
\multicolumn{1}{c}{\textbf{48.5}} & \textbf{94.2} &
\multicolumn{1}{c}{47.7} & 94.7 &
\multicolumn{1}{c}{46.6} & 95.6 &
\multicolumn{1}{c}{46.8} & 95.5 &
\multicolumn{1}{c}{47.0} & 95.4 &
\multicolumn{1}{c}{46.5} & 95.6
 \\
$\varepsilon$ = 10 &
\multicolumn{1}{c}{\textbf{54.0}} & \textbf{90.3} &
\multicolumn{1}{c}{53.4} & 90.8 &
\multicolumn{1}{c}{51.6} & 94.3 &
\multicolumn{1}{c}{50.4} & 94.9 &
\multicolumn{1}{c}{50.4} & 94.9 &
\multicolumn{1}{c}{50.9} & 94.7
 \\
$\varepsilon$ = 15 &
\multicolumn{1}{c}{\textbf{58.8}} & \textbf{86.8} &
\multicolumn{1}{c}{58.1} & 87.4 &
\multicolumn{1}{c}{55.8} & 92.8 &
\multicolumn{1}{c}{53.8} & 94.2 &
\multicolumn{1}{c}{53.2} & 94.4 &
\multicolumn{1}{c}{54.5} & 93.7
 \\
$\varepsilon$ = 20 &
\multicolumn{1}{c}{\textbf{63.5}} & \textbf{82.3} &
\multicolumn{1}{c}{62.7} & 82.7 &
\multicolumn{1}{c}{60.1} & 90.7 &
\multicolumn{1}{c}{57.4} & 93.2 &
\multicolumn{1}{c}{56.7} & 93.6 &
\multicolumn{1}{c}{58.2} & 92.3
 \\
$\varepsilon$ = 25 &
\multicolumn{1}{c}{\textbf{67.7}} & \textbf{77.2} &
\multicolumn{1}{c}{66.8} & 78.4 &
\multicolumn{1}{c}{64.0} & 87.8 &
\multicolumn{1}{c}{61.0} & 92.0 &
\multicolumn{1}{c}{60.1} & 92.6 &
\multicolumn{1}{c}{61.9} & 90.6
 \\
$\varepsilon$ = 30 &
\multicolumn{1}{c}{\textbf{71.4}} & \textbf{73.4} &
\multicolumn{1}{c}{70.5} & 73.5 &
\multicolumn{1}{c}{67.6} & 83.7 &
\multicolumn{1}{c}{64.4} & 90.4 &
\multicolumn{1}{c}{63.4} & 91.4 &
\multicolumn{1}{c}{65.4} & 88.2
 \\
$\varepsilon$ = 40 &
\multicolumn{1}{c}{\textbf{76.1}} & \textbf{67.3} &
\multicolumn{1}{c}{75.3} & 68.0 &
\multicolumn{1}{c}{72.6} & 75.4 &
\multicolumn{1}{c}{69.4} & 87.2 &
\multicolumn{1}{c}{68.5} & 88.9 &
\multicolumn{1}{c}{70.4} & 83.4
 \\
  
  \midrule
  
  
 \textbf{Norm L$_2$} \\ \textcolor{gray}{PGD2}\\

  $\varepsilon$ = 0.125 &
\multicolumn{1}{c}{\textbf{48.3}} & \textbf{94.3} &
\multicolumn{1}{c}{47.5} & 94.8 &
\multicolumn{1}{c}{46.6} & 95.6 &
\multicolumn{1}{c}{46.7} & 95.5 &
\multicolumn{1}{c}{47.1} & 95.4 &
\multicolumn{1}{c}{46.5} & 95.6
 \\
$\varepsilon$ = 0.25 &
\multicolumn{1}{c}{\textbf{53.2}} & \textbf{91.2} &
\multicolumn{1}{c}{52.6} & 91.6 &
\multicolumn{1}{c}{50.9} & 94.6 &
\multicolumn{1}{c}{50.0} & 95.0 &
\multicolumn{1}{c}{50.0} & 95.0 &
\multicolumn{1}{c}{50.3} & 94.8
 \\
$\varepsilon$ = 0.3125 &
\multicolumn{1}{c}{\textbf{55.8}} & \textbf{89.2} &
\multicolumn{1}{c}{55.2} & 89.9 &
\multicolumn{1}{c}{53.3} & 93.7 &
\multicolumn{1}{c}{51.7} & 94.6 &
\multicolumn{1}{c}{51.5} & 94.7 &
\multicolumn{1}{c}{52.3} & 94.3
 \\
$\varepsilon$ = 0.5 &
\multicolumn{1}{c}{\textbf{63.3}} & \textbf{82.6} &
\multicolumn{1}{c}{62.6} & 83.0 &
\multicolumn{1}{c}{60.0} & 90.7 &
\multicolumn{1}{c}{57.4} & 93.2 &
\multicolumn{1}{c}{56.7} & 93.5 &
\multicolumn{1}{c}{58.2} & 92.4
 \\
$\varepsilon$ = 1 &
\multicolumn{1}{c}{\textbf{76.4}} & \textbf{67.5} &
\multicolumn{1}{c}{75.7} & 67.8 &
\multicolumn{1}{c}{73.1} & 75.0 &
\multicolumn{1}{c}{70.1} & 86.7 &
\multicolumn{1}{c}{69.2} & 88.5 &
\multicolumn{1}{c}{71.0} & 83.0
 \\
$\varepsilon$ = 1.5 &
\multicolumn{1}{c}{\textbf{81.0}} & 63.0 &
\multicolumn{1}{c}{80.5} & \textbf{62.7} &
\multicolumn{1}{c}{78.5} & 63.5 &
\multicolumn{1}{c}{76.2} & 80.7 &
\multicolumn{1}{c}{75.6} & 83.2 &
\multicolumn{1}{c}{76.9} & 74.4
 \\
$\varepsilon$ = 2 &
\multicolumn{1}{c}{\textbf{82.6}} & 62.3 &
\multicolumn{1}{c}{82.1} & \textbf{61.6} &
\multicolumn{1}{c}{80.6} & 62.5 &
\multicolumn{1}{c}{78.6} & 78.5 &
\multicolumn{1}{c}{78.1} & 81.2 &
\multicolumn{1}{c}{79.1} & 72.1
 \\

  \textcolor{gray}{DeepFool} \\ 
No $\varepsilon$ & 
\multicolumn{1}{c}{\textbf{57.0}} & \textbf{91.7} &
\multicolumn{1}{c}{56.7} & \textbf{91.7} &
\multicolumn{1}{c}{55.6} & 93.6 &
\multicolumn{1}{c}{54.6} & 94.1 &
\multicolumn{1}{c}{54.2} & 94.3 &
\multicolumn{1}{c}{54.7} & 94.0
\Bstrut \\ 

  \textcolor{gray}{CW2} \\ 
$\varepsilon$ = 0.01 &
\multicolumn{1}{c}{\textbf{56.4}} & \textbf{90.8} &
\multicolumn{1}{c}{55.9} & 90.9 &
\multicolumn{1}{c}{54.5} & 93.7 &
\multicolumn{1}{c}{53.4} & 94.3 &
\multicolumn{1}{c}{53.0} & 94.5 &
\multicolumn{1}{c}{53.6} & 94.1
\Bstrut \\ 

  \textcolor{gray}{HOP}   \\ 
$\varepsilon$ = 0.1 &
\multicolumn{1}{c}{\textbf{66.1}} & \textbf{87.0} &
\multicolumn{1}{c}{65.1} & 88.2 &
\multicolumn{1}{c}{63.0} & 91.3 &
\multicolumn{1}{c}{61.2} & 92.6 &
\multicolumn{1}{c}{60.8} & 92.9 &
\multicolumn{1}{c}{61.6} & 92.1
\Bstrut \\ \midrule
  

 \textbf{Norm L$_\infty$} \\ \textcolor{gray}{PGDi, FGSM, BIM}\\

  $\varepsilon$ = 0.03125 &
\multicolumn{1}{c}{\textbf{83.0}} & 55.3 &
\multicolumn{1}{c}{82.1} & \textbf{55.2} &
\multicolumn{1}{c}{80.3} & 57.8 &
\multicolumn{1}{c}{77.4} & 77.0 &
\multicolumn{1}{c}{76.8} & 81.3 &
\multicolumn{1}{c}{78.3} & 65.4
 \\
$\varepsilon$ = 0.0625 &
\multicolumn{1}{c}{\textbf{96.0}} & \textbf{17.2} &
\multicolumn{1}{c}{94.6} & 17.4 &
\multicolumn{1}{c}{94.9} & 19.2 &
\multicolumn{1}{c}{94.3} & 21.6 &
\multicolumn{1}{c}{94.4} & 21.1 &
\multicolumn{1}{c}{94.4} & 21.1
 \\
$\varepsilon$ = 0.25 &
\multicolumn{1}{c}{\textbf{97.3}} & \textbf{0.6} &
\multicolumn{1}{c}{94.7} & 5.9 &
\multicolumn{1}{c}{96.5} & 2.5 &
\multicolumn{1}{c}{96.9} & 1.7 &
\multicolumn{1}{c}{97.2} & 1.1 &
\multicolumn{1}{c}{96.7} & 2.1
 \\
$\varepsilon$ = 0.5 &
\multicolumn{1}{c}{\textbf{82.5}} & \textbf{100.0} &
\multicolumn{1}{c}{80.4} & \textbf{100.0} &
\multicolumn{1}{c}{81.9} & \textbf{100.0} &
\multicolumn{1}{c}{82.2} & \textbf{100.0} &
\multicolumn{1}{c}{82.4} & \textbf{100.0} &
\multicolumn{1}{c}{82.0} & \textbf{100.0}
 \\


  \textcolor{gray}{PGDi, FGSM, BIM, SA} \\ 
$\varepsilon$ = 0.125 &
\multicolumn{1}{c}{9.4} & \textbf{99.9} &
\multicolumn{1}{c}{10.4} & 100.0 &
\multicolumn{1}{c}{26.2} & \textbf{99.9} &
\multicolumn{1}{c}{30.9} & 100.0 &
\multicolumn{1}{c}{\textbf{33.8}} & 100.0 &
\multicolumn{1}{c}{27.3} & 100.0
 \\
  
 \textcolor{gray}{PGDi, FGSM, BIM, CWi} \\ 
$\varepsilon$ = 0.3125 &
\multicolumn{1}{c}{\textbf{63.2}} & 99.1 &
\multicolumn{1}{c}{62.7} & \textbf{99.0} &
\multicolumn{1}{c}{61.9} & 99.3 &
\multicolumn{1}{c}{60.9} & 99.5 &
\multicolumn{1}{c}{60.5} & 99.5 &
\multicolumn{1}{c}{61.2} & 99.4
\Bstrut \\ \midrule
  
  \textbf{No norm} \\ \textcolor{gray}{STA}\\

No $\varepsilon$ & 
\multicolumn{1}{c}{88.5} & 38.8 &
\multicolumn{1}{c}{92.0} & 25.1 &
\multicolumn{1}{c}{92.1} & 22.4 &
\multicolumn{1}{c}{\textbf{93.3}} & \textbf{18.3} &
\multicolumn{1}{c}{92.7} & 19.6 &
\multicolumn{1}{c}{92.7} & 19.7
\Bstrut \\ 
  \bottomrule
\end{tabular}
}
\label{tab:cifar10_nss}
\end{table*}

\begin{table*}[!htbp]
\centering
\caption{Simultaneous attacks detection: the proposed method on CIFAR10. We train NSS on natural and adversarial examples created with PGD algorithm and L$_\infty$ norm constraint. The perturbation magnitude $\varepsilon$ is shown in the columns. We indicate in \textbf{bold} the best result.}
\resizebox{2\columnwidth}{!}{%
\begin{tabular}{r|cc|cc|cc|cc|cc|cc}
\toprule
 &
  \multicolumn{12}{c}{\textbf{Ours }} \Bstrut \\ \cline{2-13}  &
   
  \multicolumn{2}{c|}{0.03125} & \multicolumn{2}{c|}{0.0625}& \multicolumn{2}{c|}{0.125}& \multicolumn{2}{c|}{0.25} &
  \multicolumn{2}{c|}{0.3125} &   \multicolumn{2}{c}{0.5}
  \Tstrut\Bstrut \\ \cmidrule{2-13} 
  &
  \multicolumn{1}{c}{\auc} &
  \fpr &
  \multicolumn{1}{c}{\auc} &
  \fpr &
  \multicolumn{1}{c}{\auc} &
  \fpr &
  \multicolumn{1}{c}{\auc} &
  \fpr &
  \multicolumn{1}{c}{\auc} &
  \fpr &
  \multicolumn{1}{c}{\auc} &
  \fpr \\
  \cmidrule{2-13}
\textbf{Norm L$_1$}\\\textcolor{gray}{PGD1}\\
 $\varepsilon$ = 5 &
\multicolumn{1}{c}{\textbf{69.7}} & 82.5 &
\multicolumn{1}{c}{65.5} & \textbf{81.5} &
\multicolumn{1}{c}{62.1} & 87.1 &
\multicolumn{1}{c}{56.3} & 93.8 &
\multicolumn{1}{c}{53.2} & 94.8 &
\multicolumn{1}{c}{48.5} & 95.5\\
$\varepsilon$ = 10 &
\multicolumn{1}{c}{62.3} & \textbf{83.3} &
\multicolumn{1}{c}{\textbf{62.7}} & 86.3 &
\multicolumn{1}{c}{56.8} & 90.6 &
\multicolumn{1}{c}{52.1} & 94.7 &
\multicolumn{1}{c}{52.9} & 94.6 &
\multicolumn{1}{c}{50.9} & 95.0\\
$\varepsilon$ = 15 &
\multicolumn{1}{c}{66.6} & \textbf{72.7} &
\multicolumn{1}{c}{\textbf{73.9}} & 77.9 &
\multicolumn{1}{c}{69.3} & 84.4 &
\multicolumn{1}{c}{65.5} & 89.0 &
\multicolumn{1}{c}{64.3} & 91.0 &
\multicolumn{1}{c}{60.4} & 93.1\\
$\varepsilon$ = 20 &
\multicolumn{1}{c}{72.8} & \textbf{58.0} &
\multicolumn{1}{c}{\textbf{83.7}} & 59.3 &
\multicolumn{1}{c}{78.7} & 73.1 &
\multicolumn{1}{c}{73.8} & 82.5 &
\multicolumn{1}{c}{73.5} & 85.4 &
\multicolumn{1}{c}{69.2} & 90.3\\
$\varepsilon$ = 25 &
\multicolumn{1}{c}{76.8} & 42.4 &
\multicolumn{1}{c}{\textbf{89.4}} & \textbf{35.9} &
\multicolumn{1}{c}{87.1} & 50.8 &
\multicolumn{1}{c}{81.3} & 68.6 &
\multicolumn{1}{c}{79.3} & 78.0 &
\multicolumn{1}{c}{74.8} & 87.2\\
$\varepsilon$ = 30 &
\multicolumn{1}{c}{79.1} & 31.1 &
\multicolumn{1}{c}{\textbf{91.7}} & \textbf{21.4} &
\multicolumn{1}{c}{90.3} & 35.4 &
\multicolumn{1}{c}{84.3} & 61.2 &
\multicolumn{1}{c}{81.9} & 73.5 &
\multicolumn{1}{c}{77.5} & 85.3\\
$\varepsilon$ = 40 &
\multicolumn{1}{c}{80.8} & 22.2 &
\multicolumn{1}{c}{\textbf{93.0}} & \textbf{15.0} &
\multicolumn{1}{c}{92.1} & 26.4 &
\multicolumn{1}{c}{85.9} & 56.8 &
\multicolumn{1}{c}{83.1} & 71.4 &
\multicolumn{1}{c}{78.8} & 84.5\\
  
  \midrule
  
  
 \textbf{Norm L$_2$} \\ \textcolor{gray}{PGD2}\\

  $\varepsilon$ = 0.125 &
\multicolumn{1}{c}{\textbf{71.3}} & 80.8 &
\multicolumn{1}{c}{67.0} & \textbf{80.2} &
\multicolumn{1}{c}{63.9} & 85.4 &
\multicolumn{1}{c}{56.2} & 93.8 &
\multicolumn{1}{c}{53.8} & 94.7 &
\multicolumn{1}{c}{48.6} & 95.5\\
$\varepsilon$ = 0.25 &
\multicolumn{1}{c}{\textbf{63.1}} & \textbf{83.4} &
\multicolumn{1}{c}{62.8} & 86.7 &
\multicolumn{1}{c}{57.1} & 90.5 &
\multicolumn{1}{c}{52.3} & 94.6 &
\multicolumn{1}{c}{52.6} & 94.7 &
\multicolumn{1}{c}{49.9} & 95.2\\
$\varepsilon$ = 0.3125 &
\multicolumn{1}{c}{64.1} & \textbf{79.3} &
\multicolumn{1}{c}{\textbf{67.3}} & 83.1 &
\multicolumn{1}{c}{61.0} & 88.9 &
\multicolumn{1}{c}{58.0} & 92.8 &
\multicolumn{1}{c}{57.7} & 93.3 &
\multicolumn{1}{c}{54.5} & 94.4\\
$\varepsilon$ = 0.5 &
\multicolumn{1}{c}{72.9} & \textbf{58.9} &
\multicolumn{1}{c}{\textbf{83.7}} & 60.7 &
\multicolumn{1}{c}{79.4} & 73.2 &
\multicolumn{1}{c}{74.6} & 81.4 &
\multicolumn{1}{c}{73.4} & 85.4 &
\multicolumn{1}{c}{68.8} & 90.5\\
$\varepsilon$ = 1 &
\multicolumn{1}{c}{81.0} & 21.7 &
\multicolumn{1}{c}{\textbf{92.9}} & \textbf{15.5} &
\multicolumn{1}{c}{91.4} & 26.4 &
\multicolumn{1}{c}{85.5} & 57.2 &
\multicolumn{1}{c}{82.9} & 72.2 &
\multicolumn{1}{c}{78.7} & 84.7\\
$\varepsilon$ = 1.5 &
\multicolumn{1}{c}{81.5} & 19.2 &
\multicolumn{1}{c}{\textbf{93.2}} & \textbf{14.2} &
\multicolumn{1}{c}{91.9} & 24.2 &
\multicolumn{1}{c}{85.9} & 56.3 &
\multicolumn{1}{c}{83.2} & 71.9 &
\multicolumn{1}{c}{79.2} & 84.4\\
$\varepsilon$ = 2 &
\multicolumn{1}{c}{81.6} & 19.0 &
\multicolumn{1}{c}{\textbf{93.2}} & \textbf{14.1} &
\multicolumn{1}{c}{91.9} & 24.1 &
\multicolumn{1}{c}{85.9} & 56.3 &
\multicolumn{1}{c}{83.3} & 71.8 &
\multicolumn{1}{c}{79.2} & 84.4
 \\

  \textcolor{gray}{DeepFool} \\ 
No $\varepsilon$ & 
\multicolumn{1}{c}{\textbf{91.1}} & \textbf{22.0} &
\multicolumn{1}{c}{87.4} & 33.9 &
\multicolumn{1}{c}{81.9} & 54.8 &
\multicolumn{1}{c}{70.0} & 84.4 &
\multicolumn{1}{c}{64.2} & 91.5 &
\multicolumn{1}{c}{56.3} & 94.4
\Bstrut \\ 

  \textcolor{gray}{CW2} \\ 
$\varepsilon$ = 0.01 &
\multicolumn{1}{c}{52.9} & \textbf{90.5} &
\multicolumn{1}{c}{50.7} & 90.6 &
\multicolumn{1}{c}{\textbf{53.4}} & 92.2 &
\multicolumn{1}{c}{53.1} & 94.4 &
\multicolumn{1}{c}{52.0} & 94.8 &
\multicolumn{1}{c}{50.9} & 95.0
\Bstrut \\ 

  \textcolor{gray}{HOP}   \\ 
  $\varepsilon$ = 0.1 &
\multicolumn{1}{c}{\textbf{91.3}} & \textbf{20.9} &
\multicolumn{1}{c}{89.0} & 31.0 &
\multicolumn{1}{c}{86.1} & 49.1 &
\multicolumn{1}{c}{77.0} & 80.7 &
\multicolumn{1}{c}{72.4} & 88.1 &
\multicolumn{1}{c}{64.3} & 92.8
\Bstrut \\ \midrule
  

 \textbf{Norm L$_\infty$} \\ \textcolor{gray}{PGDi, FGSM, BIM}\\
  
  $\varepsilon$ = 0.03125 &
\multicolumn{1}{c}{67.2} & 77.3 &
\multicolumn{1}{c}{77.8} & 65.2 &
\multicolumn{1}{c}{\textbf{82.3}} & \textbf{59.7} &
\multicolumn{1}{c}{78.0} & 72.1 &
\multicolumn{1}{c}{73.7} & 83.8 &
\multicolumn{1}{c}{64.1} & 92.2\\
$\varepsilon$ = 0.0625 &
\multicolumn{1}{c}{69.0} & 83.6 &
\multicolumn{1}{c}{85.3} & 47.4 &
\multicolumn{1}{c}{\textbf{92.0}} & \textbf{29.6} &
\multicolumn{1}{c}{90.7} & 35.7 &
\multicolumn{1}{c}{88.0} & 45.6 &
\multicolumn{1}{c}{81.3} & 78.3\\
$\varepsilon$ = 0.25 &
\multicolumn{1}{c}{72.0} & 67.4 &
\multicolumn{1}{c}{91.8} & 23.2 &
\multicolumn{1}{c}{\textbf{95.9}} & \textbf{8.8} &
\multicolumn{1}{c}{94.1} & 15.4 &
\multicolumn{1}{c}{92.6} & 19.5 &
\multicolumn{1}{c}{91.6} & 26.5\\
$\varepsilon$ = 0.5 &
\multicolumn{1}{c}{58.3} & 84.8 &
\multicolumn{1}{c}{84.2} & 44.1 &
\multicolumn{1}{c}{\textbf{94.6}} & \textbf{9.7} &
\multicolumn{1}{c}{91.2} & 16.5 &
\multicolumn{1}{c}{90.5} & 18.8 &
\multicolumn{1}{c}{91.3} & 22.3
 \\


  \textcolor{gray}{PGDi, FGSM, BIM, SA} \\ 
$\varepsilon$ = 0.125 &
\multicolumn{1}{c}{69.0} & 79.1 &
\multicolumn{1}{c}{84.1} & 41.9 &
\multicolumn{1}{c}{\textbf{88.9}} & \textbf{40.8} &
\multicolumn{1}{c}{86.6} & 52.3 &
\multicolumn{1}{c}{85.4} & 60.4 &
\multicolumn{1}{c}{80.7} & 79.0
 \\
  
 \textcolor{gray}{PGDi, FGSM, BIM, CWi} \\ 
$\varepsilon$ = 0.3125 &
\multicolumn{1}{c}{66.6} & 75.0 &
\multicolumn{1}{c}{\textbf{80.6}} & \textbf{51.5} &
\multicolumn{1}{c}{80.0} & 61.1 &
\multicolumn{1}{c}{72.0} & 84.0 &
\multicolumn{1}{c}{67.2} & 90.0 &
\multicolumn{1}{c}{60.0} & 93.6
\Bstrut \\ \midrule
  
  \textbf{No norm} \\ \textcolor{gray}{STA}\\

No $\varepsilon$ & 
\multicolumn{1}{c}{84.8} & \textbf{33.8} &
\multicolumn{1}{c}{\textbf{85.0}} & 41.5 &
\multicolumn{1}{c}{82.7} & 52.4 &
\multicolumn{1}{c}{72.9} & 77.7 &
\multicolumn{1}{c}{70.2} & 81.7 &
\multicolumn{1}{c}{63.1} & 92.1
\Bstrut \\ 
  \bottomrule
\end{tabular}
}
\label{tab:cifar10_salad}
\end{table*}

\begin{table*}[!htbp]
\centering
\caption{Simultaneous attacks detection: NSS on SVHN. We train NSS on natural and adversarial examples created with PGD algorithm and L$_\infty$ norm constraint. The perturbation magnitude $\varepsilon$ is shown in the columns. We indicate in \textbf{bold} the best result.}
\resizebox{2\columnwidth}{!}{%
\begin{tabular}{r|cc|cc|cc|cc|cc|cc}
\toprule
 &
  \multicolumn{12}{c}{\textbf{NSS}} \Bstrut \\ \cline{2-13} &
   
  \multicolumn{2}{c|}{0.03125} & \multicolumn{2}{c|}{0.0625}& \multicolumn{2}{c|}{0.125}& \multicolumn{2}{c|}{0.25} &
  \multicolumn{2}{c|}{0.3125} &   \multicolumn{2}{c}{0.5}
  \Tstrut\Bstrut \\ \cmidrule{2-13} 
  &
  \multicolumn{1}{c}{\auc} &
\fpr &
  \multicolumn{1}{c}{\auc} &
\fpr &
  \multicolumn{1}{c}{\auc} &
\fpr &
  \multicolumn{1}{c}{\auc} &
\fpr &
  \multicolumn{1}{c}{\auc} &
\fpr &
  \multicolumn{1}{c}{\auc} &
\fpr \\\cmidrule{2-13}
\textbf{Norm L$_1$} \\ \textcolor{gray}{PGD1} \\
 $\varepsilon$ = 5 &
\multicolumn{1}{c}{37.9} & 89.3 &
\multicolumn{1}{c}{\textbf{40.2}} & 91.3 &
\multicolumn{1}{c}{37.2} & 89.2 &
\multicolumn{1}{c}{4.9} & 35.5 &
\multicolumn{1}{c}{0.3} & 8.5 &
\multicolumn{1}{c}{0.0} & \textbf{3.1}\\
$\varepsilon$ = 10 &
\multicolumn{1}{c}{33.7} & 89.3 &
\multicolumn{1}{c}{\textbf{36.9}} & 91.3 &
\multicolumn{1}{c}{34.6} & 89.2 &
\multicolumn{1}{c}{6.0} & 35.5 &
\multicolumn{1}{c}{0.4} & 8.5 &
\multicolumn{1}{c}{0.0} & \textbf{3.1}\\
$\varepsilon$ = 15 &
\multicolumn{1}{c}{31.9} & 89.3 &
\multicolumn{1}{c}{\textbf{35.6}} & 91.3 &
\multicolumn{1}{c}{34.4} & 89.2 &
\multicolumn{1}{c}{7.6} & 35.5 &
\multicolumn{1}{c}{0.5} & 8.5 &
\multicolumn{1}{c}{0.1} & \textbf{3.1}\\
$\varepsilon$ = 20 &
\multicolumn{1}{c}{31.5} & 89.3 &
\multicolumn{1}{c}{\textbf{36.1}} & 91.3 &
\multicolumn{1}{c}{35.7} & 89.2 &
\multicolumn{1}{c}{9.5} & 35.5 &
\multicolumn{1}{c}{0.6} & 8.5 &
\multicolumn{1}{c}{0.1} & \textbf{3.1}\\
$\varepsilon$ = 25 &
\multicolumn{1}{c}{32.8} & 89.3 &
\multicolumn{1}{c}{37.8} & 91.3 &
\multicolumn{1}{c}{\textbf{38.2}} & 89.2 &
\multicolumn{1}{c}{11.7} & 35.5 &
\multicolumn{1}{c}{0.9} & 8.5 &
\multicolumn{1}{c}{0.1} & \textbf{3.1}\\
$\varepsilon$ = 30 &
\multicolumn{1}{c}{34.5} & 89.3 &
\multicolumn{1}{c}{39.8} & 91.3 &
\multicolumn{1}{c}{\textbf{40.6}} & 89.2 &
\multicolumn{1}{c}{14.1} & 35.5 &
\multicolumn{1}{c}{1.2} & 8.5 &
\multicolumn{1}{c}{0.1} & \textbf{3.1}\\
$\varepsilon$ = 40 &
\multicolumn{1}{c}{37.9} & 89.3 &
\multicolumn{1}{c}{43.1} & 91.3 &
\multicolumn{1}{c}{\textbf{43.4}} & 89.0 &
\multicolumn{1}{c}{16.4} & 35.5 &
\multicolumn{1}{c}{2.2} & 8.5 &
\multicolumn{1}{c}{0.3} & \textbf{3.1}\\
  \midrule

  
\textbf{Norm L$_2$} \\ \textcolor{gray}{PGD2} \\
  $\varepsilon$ = 0.125 &
\multicolumn{1}{c}{38.7} & 89.3 &
\multicolumn{1}{c}{\textbf{40.8}} & 91.3 &
\multicolumn{1}{c}{37.6} & 89.2 &
\multicolumn{1}{c}{4.7} & 35.5 &
\multicolumn{1}{c}{0.3} & 8.5 &
\multicolumn{1}{c}{0.0} & \textbf{3.1}\\
$\varepsilon$ = 0.25 &
\multicolumn{1}{c}{34.0} & 89.3 &
\multicolumn{1}{c}{\textbf{37.2}} & 91.3 &
\multicolumn{1}{c}{34.6} & 89.2 &
\multicolumn{1}{c}{5.4} & 35.5 &
\multicolumn{1}{c}{0.3} & 8.5 &
\multicolumn{1}{c}{0.0} & \textbf{3.1}\\
$\varepsilon$ = 0.3125 &
\multicolumn{1}{c}{32.6} & 89.3 &
\multicolumn{1}{c}{\textbf{36.1}} & 91.3 &
\multicolumn{1}{c}{34.1} & 89.2 &
\multicolumn{1}{c}{6.1} & 35.5 &
\multicolumn{1}{c}{0.4} & 8.5 &
\multicolumn{1}{c}{0.0} & \textbf{3.1}\\
$\varepsilon$ = 0.5 &
\multicolumn{1}{c}{31.4} & 89.3 &
\multicolumn{1}{c}{\textbf{35.9}} & 91.3 &
\multicolumn{1}{c}{35.4} & 89.2 &
\multicolumn{1}{c}{8.9} & 35.5 &
\multicolumn{1}{c}{0.5} & 8.5 &
\multicolumn{1}{c}{0.1} & \textbf{3.1}\\
$\varepsilon$ = 1 &
\multicolumn{1}{c}{37.4} & 89.3 &
\multicolumn{1}{c}{42.5} & 91.3 &
\multicolumn{1}{c}{\textbf{42.9}} & 89.2 &
\multicolumn{1}{c}{16.0} & 35.5 &
\multicolumn{1}{c}{2.1} & 8.5 &
\multicolumn{1}{c}{0.3} & \textbf{3.1}\\
$\varepsilon$ = 1.5 &
\multicolumn{1}{c}{40.0} & 89.3 &
\multicolumn{1}{c}{46.3} & 91.3 &
\multicolumn{1}{c}{\textbf{46.5}} & 88.4 &
\multicolumn{1}{c}{17.2} & 35.5 &
\multicolumn{1}{c}{2.8} & 8.5 &
\multicolumn{1}{c}{0.6} & \textbf{3.1}\\
$\varepsilon$ = 2 &
\multicolumn{1}{c}{42.1} & 89.3 &
\multicolumn{1}{c}{49.8} & 91.3 &
\multicolumn{1}{c}{\textbf{50.5}} & 88.0 &
\multicolumn{1}{c}{18.7} & 35.5 &
\multicolumn{1}{c}{3.2} & 8.5 &
\multicolumn{1}{c}{0.8} & \textbf{3.1}
 \\

  \textcolor{gray}{DeepFool} \\
No $\varepsilon$ &
\multicolumn{1}{c}{38.1} & 89.3 &
\multicolumn{1}{c}{\textbf{41.3}} & 91.3 &
\multicolumn{1}{c}{39.7} & 89.2 &
\multicolumn{1}{c}{9.2} & 35.5 &
\multicolumn{1}{c}{0.8} & 8.5 &
\multicolumn{1}{c}{0.1} & \textbf{3.1}
\Bstrut \\ 
  
  \textcolor{gray}{CW2} \\ 
  $\varepsilon$ = 0.01 &
\multicolumn{1}{c}{37.9} & 89.3 &
\multicolumn{1}{c}{\textbf{41.0}} & 91.3 &
\multicolumn{1}{c}{39.5} & 89.2 &
\multicolumn{1}{c}{9.3} & 35.5 &
\multicolumn{1}{c}{0.8} & 8.5 &
\multicolumn{1}{c}{0.1} & \textbf{3.1}
\Bstrut \\ 
  
 \textcolor{gray}{HOP}   \\ 
 $\varepsilon$ = 0.1 &
\multicolumn{1}{c}{66.8} & 82.3 &
\multicolumn{1}{c}{\textbf{67.6}} & 84.2 &
\multicolumn{1}{c}{60.3} & 84.6 &
\multicolumn{1}{c}{16.4} & 35.5 &
\multicolumn{1}{c}{2.7} & 8.5 &
\multicolumn{1}{c}{0.7} & \textbf{3.1}
\Bstrut \\ \midrule
  

\textbf{Norm L$_\infty$} \\ \textcolor{gray}{PGDi, FGSM, BIM}\\
 $\varepsilon$ = 0.03125 &
\multicolumn{1}{c}{84.1} & 49.7 &
\multicolumn{1}{c}{\textbf{86.3}} & 46.9 &
\multicolumn{1}{c}{77.5} & 72.1 &
\multicolumn{1}{c}{22.2} & 33.2 &
\multicolumn{1}{c}{4.3} & 8.5 &
\multicolumn{1}{c}{1.2} & \textbf{3.1}\\
$\varepsilon$ = 0.0625 &
\multicolumn{1}{c}{87.4} & \textbf{0.2} &
\multicolumn{1}{c}{\textbf{88.9}} & 0.7 &
\multicolumn{1}{c}{87.5} & 0.6 &
\multicolumn{1}{c}{33.7} & 16.8 &
\multicolumn{1}{c}{7.4} & 6.8 &
\multicolumn{1}{c}{2.5} & 2.7\\
$\varepsilon$ = 0.25 &
\multicolumn{1}{c}{16.7} & 89.3 &
\multicolumn{1}{c}{51.6} & 88.9 &
\multicolumn{1}{c}{\textbf{52.0}} & 85.1 &
\multicolumn{1}{c}{35.4} & \textbf{0.1} &
\multicolumn{1}{c}{8.4} &\textbf{ 0.1} &
\multicolumn{1}{c}{3.0} & \textbf{0.1}\\
$\varepsilon$ = 0.5 &
\multicolumn{1}{c}{4.1} & 89.3 &
\multicolumn{1}{c}{\textbf{46.7}} & 86.7 &
\multicolumn{1}{c}{46.0} & 84.6 &
\multicolumn{1}{c}{35.4} & \textbf{0.1} &
\multicolumn{1}{c}{8.4} & \textbf{0.1} &
\multicolumn{1}{c}{3.0} & \textbf{0.1}
\Bstrut \\ 


\textcolor{gray}{PGDi, FGSM, BIM, SA} 
 \\ 
$\varepsilon$ = 0.125 &
\multicolumn{1}{c}{22.8} & 89.3 &
\multicolumn{1}{c}{32.9} & 91.3 &
\multicolumn{1}{c}{\textbf{43.6}} & 89.2 &
\multicolumn{1}{c}{30.3} & 32.7 &
\multicolumn{1}{c}{7.1} & 8.5 &
\multicolumn{1}{c}{2.5} & \textbf{3.1}
\Bstrut \\

  
 \textcolor{gray}{PGDi, FGSM, BIM, CWi} 
 \\ 
$\varepsilon$ = 0.3125 &
\multicolumn{1}{c}{4.7} & 89.3 &
\multicolumn{1}{c}{\textbf{41.3}} & 91.3 &
\multicolumn{1}{c}{40.8} & 89.2 &
\multicolumn{1}{c}{12.7} & 35.5 &
\multicolumn{1}{c}{1.7} & 8.5 &
\multicolumn{1}{c}{0.4} & \textbf{3.1}
\Bstrut \\ \midrule
  
\textbf{No norm} \\ \textcolor{gray}{STA} \\ 
No $\varepsilon$ & 
\multicolumn{1}{c}{89.3} & \textbf{0.0} &
\multicolumn{1}{c}{\textbf{91.2}} & 0.2 &
\multicolumn{1}{c}{85.9} & 23.4 &
\multicolumn{1}{c}{19.9} & 33.5 &
\multicolumn{1}{c}{4.2} & 8.3 &
\multicolumn{1}{c}{1.4} & 3.1
\Bstrut \\ 
  \bottomrule
\end{tabular}
}
\label{tab:svhn_nss}
\end{table*}


\begin{table*}[!htbp]
\centering
\caption{Simultaneous attacks detection: the proposed method on SVHN. We train NSS on natural and adversarial examples created with PGD algorithm and L$_\infty$ norm constraint. The perturbation magnitude $\varepsilon$ is shown in the columns. We indicate in \textbf{bold} the best result.}
\resizebox{2\columnwidth}{!}{%
\begin{tabular}{r|cc|cc|cc|cc|cc|cc}
\toprule
 &
  \multicolumn{12}{c}{\textbf{Ours}} \Bstrut \\ \cline{2-13} &
   
  \multicolumn{2}{c|}{0.03125} & \multicolumn{2}{c|}{0.0625}& \multicolumn{2}{c|}{0.125}& \multicolumn{2}{c|}{0.25} &
  \multicolumn{2}{c|}{0.3125} &   \multicolumn{2}{c}{0.5}
  \Tstrut\Bstrut \\ \cmidrule{2-13} 
  &
  \multicolumn{1}{c}{\auc} &
\fpr &
  \multicolumn{1}{c}{\auc} &
\fpr &
  \multicolumn{1}{c}{\auc} &
\fpr &
  \multicolumn{1}{c}{\auc} &
\fpr &
  \multicolumn{1}{c}{\auc} &
\fpr &
  \multicolumn{1}{c}{\auc} &
\fpr \\\cmidrule{2-13}
\textbf{Norm L$_1$} \\ \textcolor{gray}{PGD1}\\

  $\varepsilon$ = 5 &
\multicolumn{1}{c}{\textbf{79.3}} & \textbf{65.2} &
\multicolumn{1}{c}{77.4} & 73.4 &
\multicolumn{1}{c}{76.9} & 78.9 &
\multicolumn{1}{c}{76.9} & 79.0 &
\multicolumn{1}{c}{76.7} & 79.5 &
\multicolumn{1}{c}{74.0} & 84.4\\
$\varepsilon$ = 10 &
\multicolumn{1}{c}{\textbf{74.4}} & \textbf{65.1} &
\multicolumn{1}{c}{72.8} & 73.1 &
\multicolumn{1}{c}{71.9} & 81.6 &
\multicolumn{1}{c}{73.0} & 82.5 &
\multicolumn{1}{c}{71.9} & 84.2 &
\multicolumn{1}{c}{66.9} & 89.4\\
$\varepsilon$ = 15 &
\multicolumn{1}{c}{76.0} & \textbf{57.0} &
\multicolumn{1}{c}{75.7} & 64.6 &
\multicolumn{1}{c}{75.8} & 73.1 &
\multicolumn{1}{c}{\textbf{78.9}} & 72.5 &
\multicolumn{1}{c}{77.3} & 74.7 &
\multicolumn{1}{c}{71.9} & 84.9\\
$\varepsilon$ = 20 &
\multicolumn{1}{c}{77.3} & \textbf{48.1} &
\multicolumn{1}{c}{77.9} & 54.9 &
\multicolumn{1}{c}{79.2} & 61.9 &
\multicolumn{1}{c}{\textbf{83.6}} & 60.7 &
\multicolumn{1}{c}{82.2} & 64.3 &
\multicolumn{1}{c}{77.4} & 76.9\\
$\varepsilon$ = 25 &
\multicolumn{1}{c}{78.2} & \textbf{40.9} &
\multicolumn{1}{c}{79.4} & 44.4 &
\multicolumn{1}{c}{81.4} & 49.4 &
\multicolumn{1}{c}{\textbf{87.0}} & 48.6 &
\multicolumn{1}{c}{85.7} & 52.5 &
\multicolumn{1}{c}{81.4} & 66.7\\
$\varepsilon$ = 30 &
\multicolumn{1}{c}{78.8} & \textbf{34.4} &
\multicolumn{1}{c}{80.4} & 35.3 &
\multicolumn{1}{c}{83.0} & 36.6 &
\multicolumn{1}{c}{\textbf{89.3}} & 37.2 &
\multicolumn{1}{c}{88.1} & 41.6 &
\multicolumn{1}{c}{84.4} & 53.8\\
$\varepsilon$ = 40 &
\multicolumn{1}{c}{79.7} & 23.4 &
\multicolumn{1}{c}{81.6} & 22.4 &
\multicolumn{1}{c}{84.7} & 20.2 &
\multicolumn{1}{c}{\textbf{92.6}} & \textbf{20.0} &
\multicolumn{1}{c}{91.1} & 23.0 &
\multicolumn{1}{c}{87.8} & 30.5
 \\
  \midrule
  
  
\textbf{Norm L$_2$} \\ \textcolor{gray}{PGD2}\\
  $\varepsilon$ = 0.125 &
\multicolumn{1}{c}{\textbf{82.2}} & \textbf{61.7} &
\multicolumn{1}{c}{80.6} & 68.4 &
\multicolumn{1}{c}{80.3} & 72.4 &
\multicolumn{1}{c}{80.2} & 74.5 &
\multicolumn{1}{c}{80.1} & 73.5 &
\multicolumn{1}{c}{79.7} & 75.5\\
$\varepsilon$ = 0.25 &
\multicolumn{1}{c}{\textbf{75.7}} & \textbf{63.6} &
\multicolumn{1}{c}{74.0} & 71.7 &
\multicolumn{1}{c}{73.3} & 80.3 &
\multicolumn{1}{c}{74.0} & 81.7 &
\multicolumn{1}{c}{72.6} & 82.8 &
\multicolumn{1}{c}{67.8} & 89.0\\
$\varepsilon$ = 0.3125 &
\multicolumn{1}{c}{\textbf{75.5}} & \textbf{61.6} &
\multicolumn{1}{c}{74.3} & 70.1 &
\multicolumn{1}{c}{73.9} & 78.4 &
\multicolumn{1}{c}{75.2} & 79.4 &
\multicolumn{1}{c}{73.9} & 81.7 &
\multicolumn{1}{c}{70.6} & 86.7\\
$\varepsilon$ = 0.5 &
\multicolumn{1}{c}{77.2} & \textbf{50.6} &
\multicolumn{1}{c}{77.6} & 57.4 &
\multicolumn{1}{c}{78.6} & 64.1 &
\multicolumn{1}{c}{\textbf{82.5}} & 64.4 &
\multicolumn{1}{c}{81.2} & 67.1 &
\multicolumn{1}{c}{76.3} & 79.5\\
$\varepsilon$ = 1 &
\multicolumn{1}{c}{79.5} & 25.8 &
\multicolumn{1}{c}{81.3} & 24.8 &
\multicolumn{1}{c}{84.3} & \textbf{24.1} &
\multicolumn{1}{c}{\textbf{92.3}} & 24.7 &
\multicolumn{1}{c}{90.7} & 27.7 &
\multicolumn{1}{c}{87.1} & 36.4\\
$\varepsilon$ = 1.5 &
\multicolumn{1}{c}{80.2} & 19.5 &
\multicolumn{1}{c}{82.2} & 17.6 &
\multicolumn{1}{c}{85.6} & 14.3 &
\multicolumn{1}{c}{\textbf{94.1}} & \textbf{7.5} &
\multicolumn{1}{c}{92.9} & 8.6 &
\multicolumn{1}{c}{89.9} & 11.8\\
$\varepsilon$ = 2 &
\multicolumn{1}{c}{80.5} & 19.4 &
\multicolumn{1}{c}{82.5} & 17.5 &
\multicolumn{1}{c}{85.9} & 14.1 &
\multicolumn{1}{c}{\textbf{94.9}} & \textbf{5.3} &
\multicolumn{1}{c}{94.5} & 6.8 &
\multicolumn{1}{c}{90.7} & 9.5
 \\

  \textcolor{gray}{DeepFool} \\ 
No $\varepsilon$ & 
\multicolumn{1}{c}{\textbf{96.3}} & \textbf{8.6} &
\multicolumn{1}{c}{95.9} & 10.5 &
\multicolumn{1}{c}{95.0} & 12.9 &
\multicolumn{1}{c}{94.9} & 12.0 &
\multicolumn{1}{c}{95.3} & 12.1 &
\multicolumn{1}{c}{95.5} & 12.6
\Bstrut \\

  \textcolor{gray}{CW2} \\ 
$\varepsilon$ = 0.01 &
\multicolumn{1}{c}{\textbf{59.7}} & \textbf{76.3} &
\multicolumn{1}{c}{57.2} & 80.1 &
\multicolumn{1}{c}{53.4} & 89.9 &
\multicolumn{1}{c}{54.2} & 92.0 &
\multicolumn{1}{c}{51.1} & 93.5 &
\multicolumn{1}{c}{44.3} & 96.1
\Bstrut \\

  \textcolor{gray}{HOP}  \\
  $\varepsilon$ = 0.1 &
\multicolumn{1}{c}{\textbf{96.1}} & \textbf{7.9} &
\multicolumn{1}{c}{95.6} & 9.8 &
\multicolumn{1}{c}{95.9} & 11.7 &
\multicolumn{1}{c}{96.0} & 10.2 &
\multicolumn{1}{c}{95.9} & 9.9 &
\multicolumn{1}{c}{\textbf{96.1}} & 10.0
\Bstrut \\ \midrule
  

 \textbf{Norm L$_\infty$}\\\textcolor{gray}{PGDi, FGSM, BIM} \\
 $\varepsilon$ = 0.03125 &
\multicolumn{1}{c}{74.3} & \textbf{60.0} &
\multicolumn{1}{c}{75.8} & 60.3 &
\multicolumn{1}{c}{77.8} & 62.6 &
\multicolumn{1}{c}{\textbf{81.4}} & 64.9 &
\multicolumn{1}{c}{80.1} & 67.1 &
\multicolumn{1}{c}{76.7} & 75.5\\
$\varepsilon$ = 0.0625 &
\multicolumn{1}{c}{78.4} & 36.0 &
\multicolumn{1}{c}{80.3} & 34.1 &
\multicolumn{1}{c}{83.2} & 33.8 &
\multicolumn{1}{c}{89.1} & 3\textbf{3.3} &
\multicolumn{1}{c}{\textbf{87.9}} & 34.4 &
\multicolumn{1}{c}{85.7} & 37.4\\
$\varepsilon$ = 0.25 &
\multicolumn{1}{c}{80.1} & 19.4 &
\multicolumn{1}{c}{82.1} & 17.5 &
\multicolumn{1}{c}{85.2} & \textbf{15.8} &
\multicolumn{1}{c}{\textbf{92.3}} & 16.4 &
\multicolumn{1}{c}{92.1} & 16.8 &
\multicolumn{1}{c}{89.6} & 17.0\\
$\varepsilon$ = 0.5 &
\multicolumn{1}{c}{80.3} & 19.4 &
\multicolumn{1}{c}{82.3} & 17.5 &
\multicolumn{1}{c}{85.5} & \textbf{14.1} &
\multicolumn{1}{c}{\textbf{92.9}} & 14.4 &
\multicolumn{1}{c}{91.7} & 15.2 &
\multicolumn{1}{c}{90.1} & 14.8\\

  
 \textcolor{gray}{PGDi, FGSM, BIM, SA} \\ 
$\varepsilon$ = 0.125 &
\multicolumn{1}{c}{78.9} & 29.0 &
\multicolumn{1}{c}{80.8} & \textbf{28.1} &
\multicolumn{1}{c}{83.8} & 28.7 &
\multicolumn{1}{c}{\textbf{89.2}} & 29.1 &
\multicolumn{1}{c}{88.4} & 28.9 &
\multicolumn{1}{c}{86.8} & 28.4
\Bstrut \\ 

 \textcolor{gray}{PGDi, FGSM, BIM, CWi} \\
$\varepsilon$ = 0.3125 &
\multicolumn{1}{c}{78.7} & 33.4 &
\multicolumn{1}{c}{80.5} & 31.9 &
\multicolumn{1}{c}{83.1} & 34.0 &
\multicolumn{1}{c}{\textbf{88.2}} & 33.1 &
\multicolumn{1}{c}{88.1} & 31.7 &
\multicolumn{1}{c}{86.7} & \textbf{31.2}
\Bstrut \\\midrule

\textbf{No norm} \\ \textcolor{gray}{STA}\\
No $\varepsilon$ &
\multicolumn{1}{c}{\textbf{94.7}} & \textbf{14.5} &
\multicolumn{1}{c}{93.3} & 16.8 &
\multicolumn{1}{c}{89.9} & 23.1 &
\multicolumn{1}{c}{90.2} & 23.2 &
\multicolumn{1}{c}{91.0} & 22.4 &
\multicolumn{1}{c}{91.1} & 22.4
\Bstrut \\ 
  \bottomrule
\end{tabular}
}
\label{tab:svhn_salad}
\end{table*}

\subsection{The proposed aggregator against the adaptive-attacks in the {\mead} scenario}
\begin{figure}[t]
	\centering
		\begin{subfigure}[b]{0.3\textwidth}
		\centering
		\includegraphics[width=\textwidth]{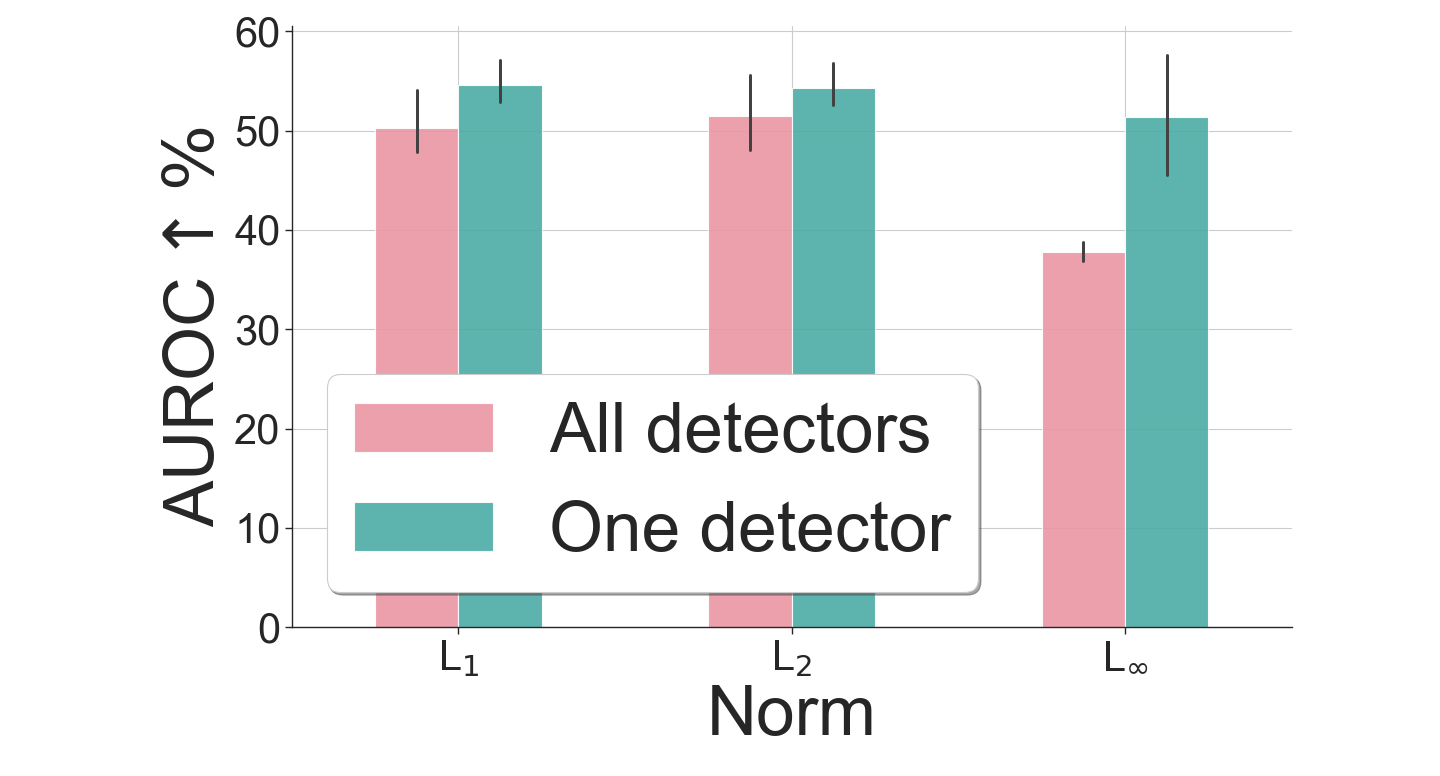}
		\vspace{-1.5\baselineskip}
		\caption{Analysis \auc}
		\label{fig:adaptive_auroc}
	\end{subfigure}
		\begin{subfigure}[b]{0.3\textwidth}
		\centering
		\includegraphics[width=\textwidth]{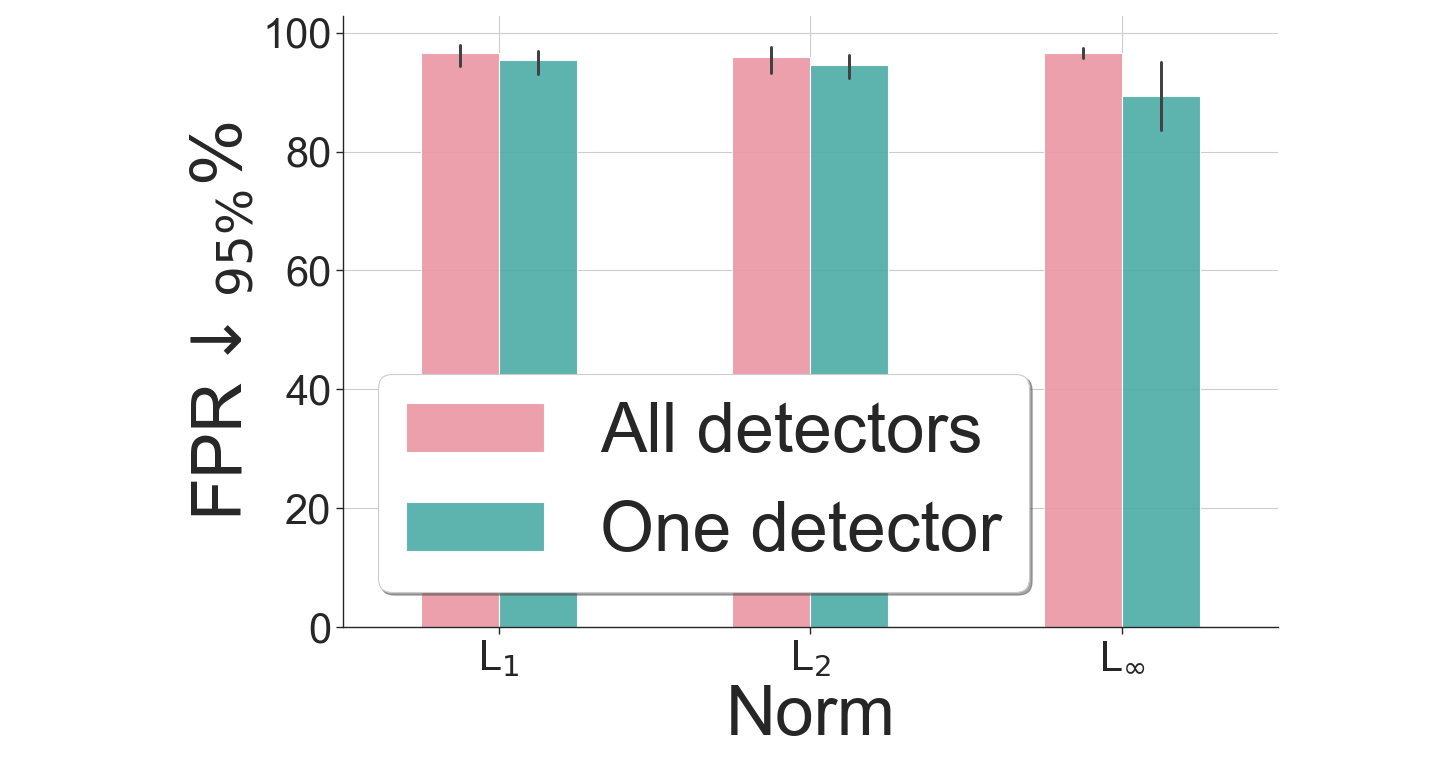}
		\vspace{-1.5\baselineskip}
		\caption{Analysis  \fpr}
		\label{fig:adaptive_fpr}
	\end{subfigure}
	\caption{Our method against the adaptive-attacks under {\mead}. We consider the worst case scenario in~\cref{tab:table_rebuttal_1,tab:table_rebuttal_1}, i.e., when $\alpha=0.1$.}
	\label{fig:adaptive}
\end{figure}
\begin{table*}[htbp!]
\centering
\caption{The proposed method against the adaptive-attacks under {\mead}. In the following setting, we attack each detector and the classifier once at a time. $\alpha$ is the parameter to control the losses.}
\ra{1.3}
\resizebox{2\columnwidth}{!}{%
\begin{tabular}{@{}r|bbcbbcbbcbbcbb@{}}\toprule
& \multicolumn{14}{c}{CIFAR10}  \\
\cmidrule{2-15}
& \multicolumn{2}{c}{$\alpha=0$} & \phantom{abc}& \multicolumn{2}{c}{$\alpha=.1$} &
  \phantom{abc} & \multicolumn{2}{c}{ $\alpha=1$} & \phantom{abc}& \multicolumn{2}{c}{$\alpha=5$} & \phantom{abc} &  \multicolumn{2}{c}{$\alpha=10$}\\ \cmidrule{2-3}
\cmidrule{5-6} \cmidrule{8-9} \cmidrule{11-12} \cmidrule{14-15}
  & \auc & \fpr  && \auc & \fpr && \auc & \fpr && \auc & \fpr && \auc & \fpr\\ 
 \midrule

\textbf{Norm L$_1$}\\

\underline{PGD1$^\star$}\\
$\varepsilon=5$ & 
\textbf{62.1} & \textbf{87.1} &&
61.3 & 88.6 && 
61.2 & 89.3 && 
63.1 & 89.2 && 
62.6 & 91.3
\\
$\varepsilon=10$ &
\textbf{56.8} & 90.6 &&
53.1 & 94.5 && 
54.4 & 93.9 && 
60.0 & 91.0 && 
60.6 & 91.9
\\
$\varepsilon=15$ & 
\textbf{69.3} & \textbf{84.4} &&
51.5 & 96.5 && 
54.7 & 94.6 && 
64.1 & 88.1 && 
65.7 & 87.7
\\
$\varepsilon=20$ & 
\textbf{78.7} & \textbf{73.1} &&
53.4 & 96.8 && 
55.9 & 94.9 && 
66.7 & 84.1 && 
69.4 & 82.7
\\
$\varepsilon=25$ & 
\textbf{87.1} & \textbf{50.8} &&
54.0 & 97.2 && 
56.7 & 94.6 && 
67.8 & 82.7 && 
71.1 & 79.0 
\\
$\varepsilon=30$ &
\textbf{90.3} & \textbf{35.4} &&
54.5 & 97.1 && 
56.6 & 94.4 && 
68.9 & 81.1 && 
71.9 & 78.4 
\\
$\varepsilon=40$ & 
\textbf{92.1} & \textbf{22.7} &&
54.4 & 97.0 && 
57.7 & 93.6 && 
69.4 & 79.7 && 
72.9 & 74.2 
\\
\midrule

\textbf{Norm L$_2$}\\

\underline{PGD2$^\star$}\\
$\varepsilon=0.125$ & 
\textbf{63.9} & \textbf{85.4} &&
61.4 & 88.0 && 
62.4 & 88.8 && 
63.7 & 88.5 && 
63.9 & 89.9
\\
$\varepsilon=0.25$ & 
\textbf{57.1} & \textbf{90.5} &&
52.9 & 94.2 && 
55.0 & 93.6 && 
60.6 & 89.7 && 
61.5 & 90.3 
\\
$\varepsilon=0.3125$ & 
\textbf{61.0} & \textbf{88.9} &&
51.6 & 95.7 && 
54.1 & 94.7 && 
62.2 & 87.8 && 
63.7 & 87.9 
\\
$\varepsilon=0.5$ & 
\textbf{79.4} & \textbf{73.2}&&
52.8 & 96.8 && 
55.3 & 94.3 && 
66.2 & 84.6 && 
68.8 & 81.5 
\\
$\varepsilon=1$ & 
\textbf{91.4} & \textbf{26.4}&&
52.7 & 96.8 && 
57.3 & 93.4 &&
69.0 & 78.3 && 
72.1 & 74.4 
\\
$\varepsilon=1.5$ & 
\textbf{91.9} & \textbf{24.2} &&
53.9 & 96.1 && 
57.9 & 91.4 && 
70.5 & 73.7 && 
74.1 & 68.1 
\\
$\varepsilon=2$ & 
\textbf{91.9} & \textbf{24.1} &&
54.6 & 94.6 && 
59.3 & 88.5 && 
72.3 & 67.8 && 
75.6 & 62.7 
\\



\midrule

\textbf{Norm L$_\infty$}\\
\underline{PGDi$^\star$, FGSM$^\star$, BIM$^\star$}\\\
$\varepsilon=0.03125$ & 
\textbf{82.3} & \textbf{59.7}&&
45.3 & 96.2 && 
46.0 & 96.4 && 
54.5 & 91.4 && 
57.4 & 89.3 
\\
$\varepsilon=0.0625$ & 
\textbf{92.0} & \textbf{29.6} &&
44.3 & 96.2 && 
49.8 & 93.8 && 
59.7 & 82.4 && 
64.3 & 76.4 
\\
$\varepsilon=0.5$ & 
\textbf{94.6} & \textbf{9.7} &&
62.1 & 81.3 && 
54.9 & 81.9 && 
66.1 & 60.8 && 
68.9 & 57.9 
\\

\underline{PGDi$^\star$, FGSM$^\star$, BIM$^\star$, SA}\\
$\varepsilon=0.125$ & 
\textbf{88.9} & \textbf{40.8} &&
48.6 & 90.7 &&
54.9 & 85.0 &&
61.9 & 73.1 &&
66.3 & 67.5 
\\

\underline{PGDi$^\star$, FGSM$^\star$, BIM$^\star$, CWi}\\
$\varepsilon=0.3125$ &
\textbf{80.0} & \textbf{61.1} &&
56.6 & 82.0 &&
56.3 & 79.6 &&
66.1 & 66.1 && 
69.2 & 64.4 
\\

\bottomrule
\end{tabular}
}
\label{tab:table_rebuttal_1}
\end{table*}
\begin{table*}[htbp!]
\centering
\caption{The proposed method against the adaptive-attacks under {\mead}. In the following setting, we attack all the detectors and the classifier together at the time. $\alpha$ is the parameter to control the losses.}
\ra{1.3}
\resizebox{2\columnwidth}{!}{%
\begin{tabular}{@{}r|bbcbbcbbcbbcbb@{}}\toprule
& \multicolumn{14}{c}{CIFAR10}  \\
\cmidrule{2-15}
& \multicolumn{2}{c}{$\alpha=0$} & \phantom{abc}& \multicolumn{2}{c}{$\alpha=.1$} &
  \phantom{abc} & \multicolumn{2}{c}{ $\alpha=1$} & \phantom{abc}& \multicolumn{2}{c}{$\alpha=5$} & \phantom{abc} &  \multicolumn{2}{c}{$\alpha=10$}\\ \cmidrule{2-3}
\cmidrule{5-6} \cmidrule{8-9} \cmidrule{11-12} \cmidrule{14-15}
  & \auc & \fpr  && \auc & \fpr && \auc & \fpr && \auc & \fpr && \auc & \fpr\\ 
 \midrule

\textbf{Norm L$_1$}\\

\underline{PGD1$^\star$}\\
$\varepsilon=5$ & 
\textbf{62.1} & \textbf{87.1} &&
61.2 & 90.4 && 
63.6 & 86.8 && 
65.8 & 83.9 && 
66.3 & 83.2
\\
$\varepsilon=10$ &
\textbf{56.8} & 90.6 &&
50.5 & 96.4 && 
55.9 & 91.6 && 
60.1 & 88.1 && 
61.1 & 87.2 
\\
$\varepsilon=15$ & 
\textbf{69.3} & \textbf{84.4} &&
47.3 & 97.6 && 
53.8 & 92.3 && 
62.0 & 84.9 && 
63.7 & 83.7
\\
$\varepsilon=20$ & 
\textbf{78.7} & \textbf{73.1} &&
47.1 & 97.9 && 
54.2 & 92.5 && 
64.2 & 82.8 && 
66.8 & 79.1
\\
$\varepsilon=25$ & 
\textbf{87.1} & \textbf{50.8} &&
47.8 & 98.0 && 
55.0 & 92.1 && 
66.5 & 79.5 && 
68.8 & 77.2
\\
$\varepsilon=30$ &
\textbf{90.3} & \textbf{35.4} &&
48.8 & 98.0 && 
55.8 & 91.3 && 
67.4 & 78.5 && 
70.4 & 75.0
\\
$\varepsilon=40$ & 
\textbf{92.1} & \textbf{22.7} &&
49.1 & 98.0 && 
56.8 & 90.5 && 
68.6 & 77.4 && 
72.5 & 71.6
\\
\midrule

\textbf{Norm L$_2$}\\

\underline{PGD2$^\star$}\\
$\varepsilon=0.125$ & 
\textbf{63.9} & \textbf{85.4} &&
62.4 & 88.5 && 
65.0 & 86.2 && 
66.9 & 82.9 && 
67.2 & 81.1
\\
$\varepsilon=0.25$ & 
\textbf{57.1} & \textbf{90.5} &&
51.2 & 96.0 && 
56.3 & 91.7 && 
60.6 & 87.2 && 
61.6 & 86.8
\\
$\varepsilon=0.3125$ & 
\textbf{61.0} & \textbf{88.9} &&
56.0 & 94.6 && 
57.9 & 93.6 && 
65.3 & 86.4 && 
66.7 & 86.6
\\
$\varepsilon=0.5$ & 
\textbf{79.4} & \textbf{73.2}&&
46.8 & 97.8 &&
54.6 & 91.3 && 
64.5 & 82.4 && 
66.8 & 79.5
\\
$\varepsilon=1$ & 
\textbf{91.4} & \textbf{26.4}&&
47.2 & 98.0 && 
57.8 & 89.4 && 
69.9 & 73.8 && 
73.1 & 71.7
\\
$\varepsilon=1.5$ & 
\textbf{91.9} & \textbf{24.2} &&
47.5 & 97.6 && 
59.9 & 86.9 && 
73.2 & 68.7 && 
76.5 & 63.1
\\
$\varepsilon=2$ & 
\textbf{91.9} & \textbf{24.1} &&
49.0 & 97.0 && 
62.8 & 83.3 && 
75.6 & 63.7 && 
79.5 & 56.6
\\



\midrule

\textbf{Norm L$_\infty$}\\
\underline{PGDi$^\star$, FGSM$^\star$, BIM$^\star$}\\\
$\varepsilon=0.03125$ & 
\textbf{82.3} & \textbf{59.7}&&
40.2 & 98.0 && 
47.6 & 95.5 && 
60.6 & 86.2 && 
65.0 & 81.8
\\
$\varepsilon=0.0625$ & 
\textbf{92.0} & \textbf{29.6} &&
37.9 & 98.0 && 
47.0 & 95.9 && 
61.9 & 82.1 && 
65.8 & 77.1
\\
$\varepsilon=0.25$ &
\textbf{95.9} & \textbf{8.8} &&
36.5 & 96.4 && 
47.4 & 97.7 && 
62.5 & 92.6 && 
65.4 & 90.8
\\
$\varepsilon=0.5$ & 
\textbf{94.6} & \textbf{9.7} &&
36.7 & 96.2 && 
46.0 & 97.7 && 
61.6 & 96.1 && 
66.0 & 94.8
\\

\underline{PGDi$^\star$, FGSM$^\star$, BIM$^\star$, SA}\\
$\varepsilon=0.125$ & 
\textbf{88.9} & \textbf{40.8} &&
38.5 & 95.9 &&
46.8 & 95.4 &&
60.1 & 85.0 &&
61.9 & 83.2\\

\underline{PGDi$^\star$, FGSM$^\star$, BIM$^\star$, CWi}\\
$\varepsilon=0.3125$ &
\textbf{80.0} & \textbf{61.1} &&
37.2 & 95.3 &&
46.7 & 97.4 &&
60.9 & 92.4 && 
64.1 & 90.1\\

\bottomrule
\end{tabular}
}
\label{tab:table_rebuttal_2}
\end{table*}

\label{subsec:adaptive_attacks}
We present a new experimental setting to address the case in which also the detectors are attacked at the same time as the target classifier, taking the cue from~\cite{Bryniarski2021,Carlini017,TramerCBM20,CarliniMagnet}. 
It is important to note that, in the spirit of the {\mead} framework, we are not simply considering a scenario in which a \textit{single} adaptive attack is perpetrated on the classifier and detectors, but rather \underline{multiple} adaptive attacks are concurrently occurring. This scenario has not yet been considered in~\cite{GranesePRMP2022ECMLPKDD}, so we are the first to deal with such a setting. We extend
the framework to include two main cases: \textit{(i)} for attacks on the classifier and the single detectors individually; \textit{(ii)} for attacks on the classifier and all the detectors simultaneously.

The tables with the complete results are \cref{tab:table_rebuttal_1,tab:table_rebuttal_2}, where $\alpha$ is the coefficient that controls the gradient's speed of the attack against the detectors. We try many different values $\alpha=\left\{.1, 1, 5, 10\right\}$. The case where $\alpha$ is equal to 0 is added for completeness, and it corresponds to the case where only the target classifier is attacked.
We report in~\cref{fig:adaptive} the comparison of the results between case \textit{(i)} and case \textit{(ii)} on CIFAR10 and $\alpha=0.1$, as this corresponds to the case with the worst performances. As can be seen, the performances of our aggregator improve when the detectors are attacked singularly. This is particularly interesting for the setting we are dealing with. Indeed, our method is not a new supervised adversarial detection method but a framework to aggregate detectors, in this case, applied to the adversarial detection problem. Hence, it does not propose solving the problem of finding a new robust method for adaptive attacks but rather creating a mixture of experts based on the proposed sound mathematical framework. Thus, an attacker to successfully fool our method needs to have the \textit{complete access to all} the underlying detectors and also \textit{an up-to-the-date knowledge of the detectors employed} as the defender can always include a new detection mechanism to the pool of the detectors.
\begin{table}[!htbp]
\vspace{5mm}
\centering
\caption{Comparison between the proposed method and the single detectors (\textit{stronger} version) against the adaptive-attacks. Norm L$_\infty$ and $\varepsilon=0.25$ (i.e., attacks PGDi$^\star$, FGSM$^\star$, BIM$^\star$).}
\ra{1.3}
\resizebox{.9\columnwidth}{!}{%
\begin{tabular}{@{}r|b|b|b|b|b@{}}\toprule
CIFAR10 & \multicolumn{1}{c}{Ours} &
\multicolumn{1}{c}{ACE} & \multicolumn{1}{c}{KL}  & \multicolumn{1}{c}{FR} & \multicolumn{1}{c}{{Gini}} \\ \midrule
\auc & \textbf{54.6} & 35.7 & 30.6 & 26.3 & 36.2 \\
\fpr & \textbf{73.0} & 96.5 & 97.0 & 97.4 & 99.6 \\
\bottomrule
\end{tabular}
}
\label{tab:adaptive_strong}
\end{table}
To give more insights on the proposed aggregator under this setting, we train a \textit{stronger} version of the four shallow detectors where the detectors at training time have seen the corresponding adaptive attacks generated through the PGD algorithm. We report the results in~\cref{tab:adaptive_strong} where we focus on the group of simultaneous attacks with L$_\infty$ norm and $\varepsilon=0.25$ as this represents the worst result of our method in~\cref{tab:table_rebuttal_2}.
If our method was only good as the best among the detectors, we should expect similar results in~\cref{tab:adaptive_strong}. In this case, the only solution would be to train a better detector. \textbf{However, the strength of the aggregator is not just mimicking the performance of its parts but rather creating a mixture of experts based on the proposed sound mathematical framework.} Therefore, we should expect better performances. Indeed, this consistently holds as the method performs much better than the best detector.
\subsection{AutoAttack}
\begin{table}[htbp!]
\centering
\caption{The proposed method on AutoAttack ({\mead} setting). The attacks are APGD-CE, APGD-DLR, FAB, SA.}
\ra{1.3}
\resizebox{0.5\columnwidth}{!}{%
\begin{tabular}{@{}r|bb@{}}\toprule
& \multicolumn{2}{c}{CIFAR10}  \\
\cmidrule{2-3}
& \multicolumn{2}{c}{Ours} \\ \cmidrule{2-3}
  & \auc & \fpr  \\ 
 \midrule

\textbf{Norm L$_1$}\\

\underline{}\\
$\varepsilon=5$ & 
57.1 & 88.4
\\
$\varepsilon=10$ &
67.1 & 75.7
\\
$\varepsilon=15$ & 
72.2 & 66.7
\\
$\varepsilon=20$ & 
72.7 & 65.2
\\
$\varepsilon=25$ & 
72.8 & 65.6
\\
$\varepsilon=30$ &
73.4 & 64.0
\\
$\varepsilon=40$ & 
73.6 & 64.0
\\
\midrule

\textbf{Norm L$_2$}\\

\underline{}\\
$\varepsilon=0.125$ & 
67.4 & 81.0
\\
$\varepsilon=0.25$ & 
58.0 & 89.0
\\
$\varepsilon=0.3125$ & 
58.1 & 88.8
\\
$\varepsilon=0.5$ & 
69.4 & 74.7
\\
$\varepsilon=1$ & 
75.1 & 61.6
\\
$\varepsilon=1.5$ & 
76.1 & 60.7
\\
$\varepsilon=2$ & 
76.1 & 60.5
\\

\midrule

\textbf{Norm L$_\infty$}\\
\underline{}\\\
$\varepsilon=0.03125$ & 
75.7 & 61.0
\\
$\varepsilon=0.0625$ & 
76.0 & 60.7
\\
$\varepsilon=0.125$ & 
76.8 & 60.3
\\
$\varepsilon=0.25$ &
76.8 & 60.0
\\
$\varepsilon=0.3125$ &
78.6 & 57.6
\\
$\varepsilon=0.5$ & 
76.1 & 60.3
\\
\bottomrule
\end{tabular}
}
\label{tab:autoattack}
\end{table}
We present an application of AutoAttack~\cite{Croce020a}, a state-of-the-art evaluation tool for robustness, redesigned for adversarial detection evaluation and adapted to our simultaneous attacks framework.
In its original version, AutoAttack evaluates the accuracy of robust classifiers. In so doing, ~\cite{Croce020a} proposes a multiple attacks framework to ensure that at least one attack succeeds in producing an adversarial example for each natural one. In their context, it does not matter which attack will succeed since any successful attack would undermine the accuracy of the target classifier in the same way. In our case, the number of different successful attacks for each natural sample will affect the detection quality since a detector is successful only if it can detect all of them. Because of the above mentioned differences, it is impossible to deploy it directly in our framework without any modifications. A modified version of AutoAttack, adapted to the evaluation of our proposed method, has been implemented, and the results are presented below. While AutoAttack suggests using different attack strategies, in our case, we combine different attack strategies matched with different losses to make the pool of attacks more strong and more diversified.
\label{subsec:non_simult}

\subsection{Additional plots}
\label{app:additional_plot}
\begin{figure}[!h]
	\centering
		\begin{subfigure}[b]{0.22\textwidth}
		\centering
		\includegraphics[width=\textwidth]{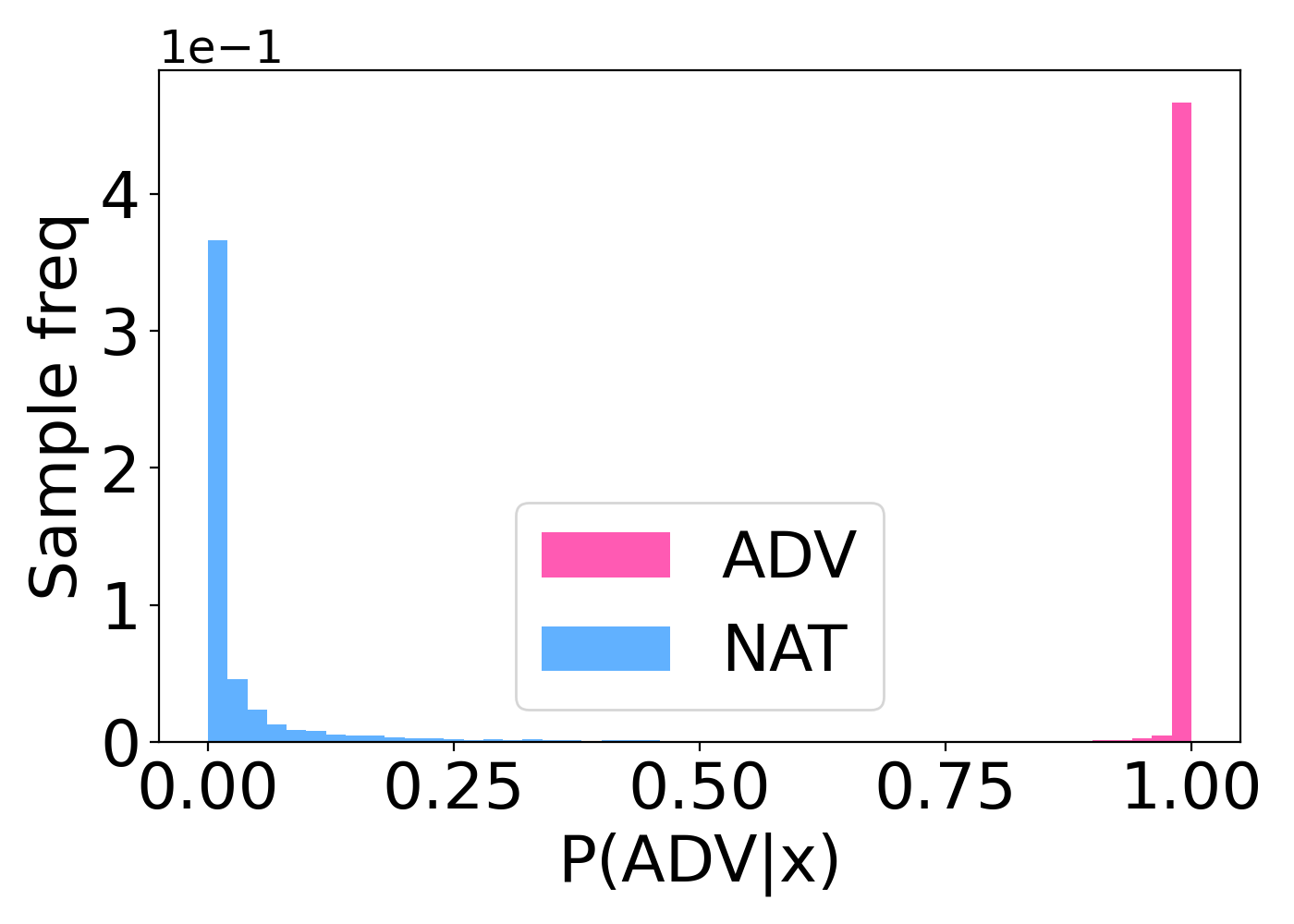}
		\vspace{-1.5\baselineskip}
		\caption{PGD-L$_1$-40-ACE}
		\label{fig:h_CE}
	\end{subfigure}
	    \hfill
		\begin{subfigure}[b]{0.22\textwidth}
		\centering
		\includegraphics[width=\textwidth]{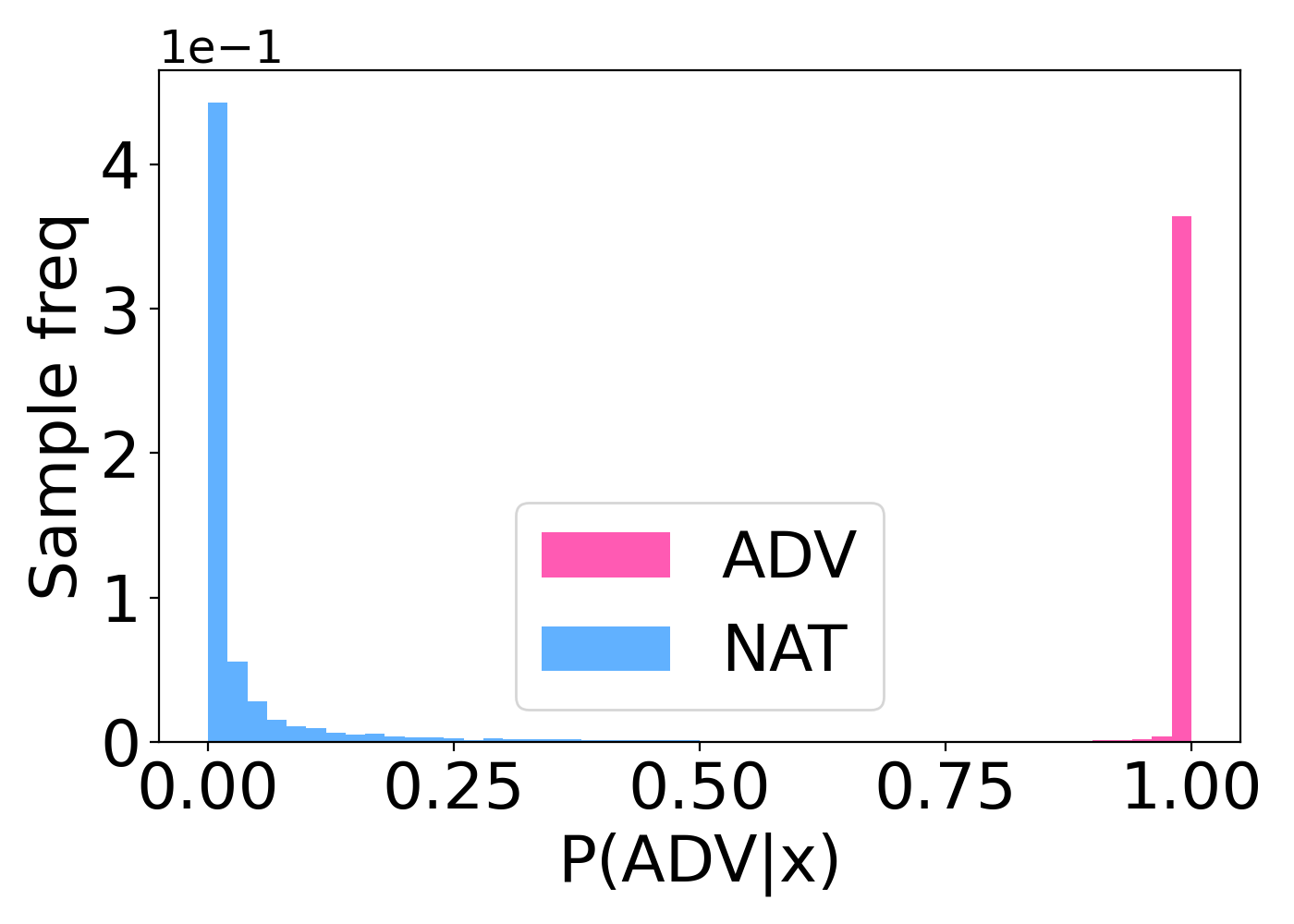}
		\vspace{-1.5\baselineskip}
		\caption{PGD-L$_1$-40-KL}
		\label{fig:h_KL}
	\end{subfigure}
	 \hfill
	\begin{subfigure}[b]{0.22\textwidth}
	    \centering
	    \includegraphics[width=\textwidth]{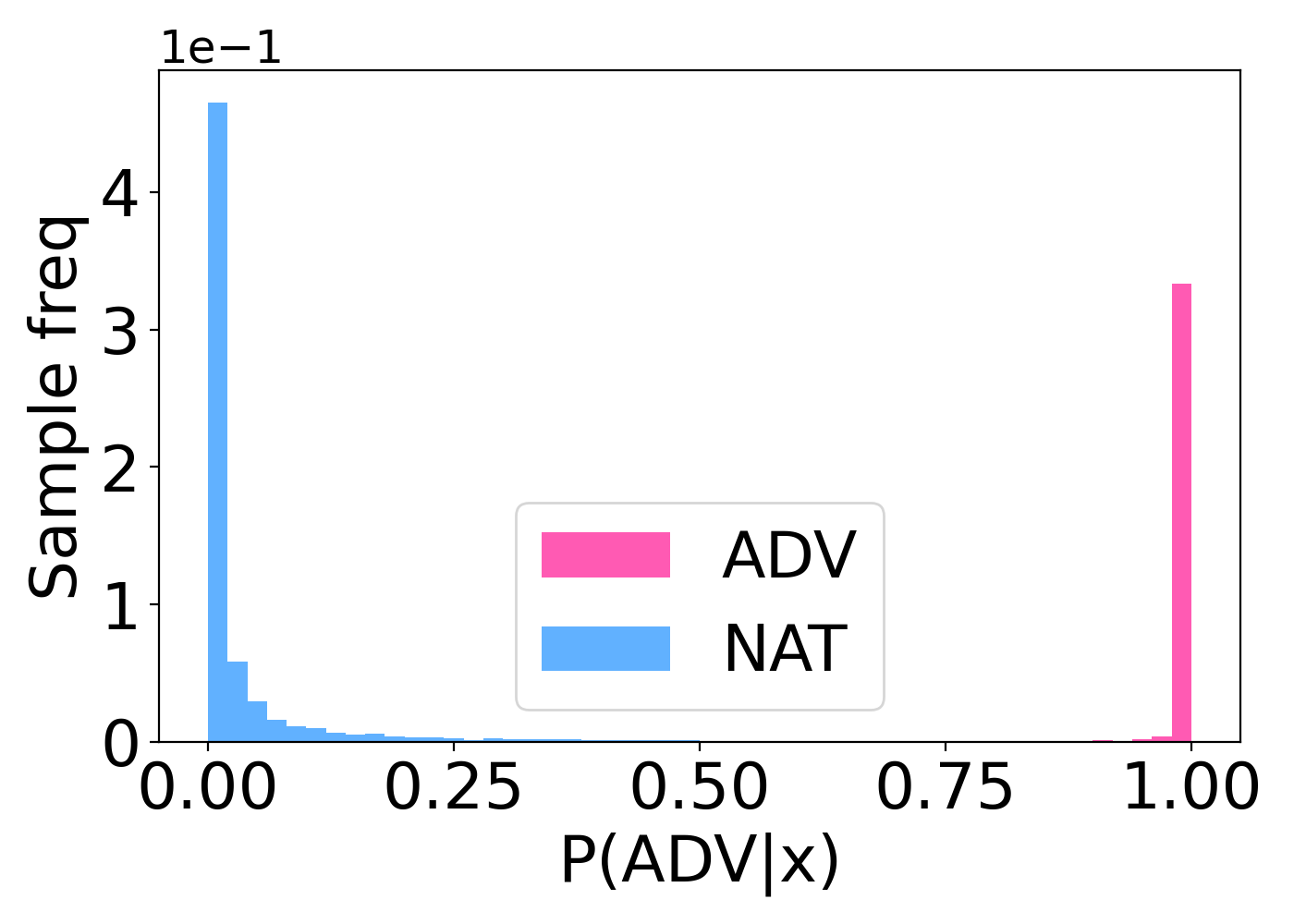}
	    \vspace{-1.5\baselineskip}
	    \caption{PGD-L$_1$-40-FR}
	    \label{fig:h_rao}
	\end{subfigure}
    \hfill
	\begin{subfigure}[b]{0.22\textwidth}
		\centering
		\includegraphics[width=\textwidth]{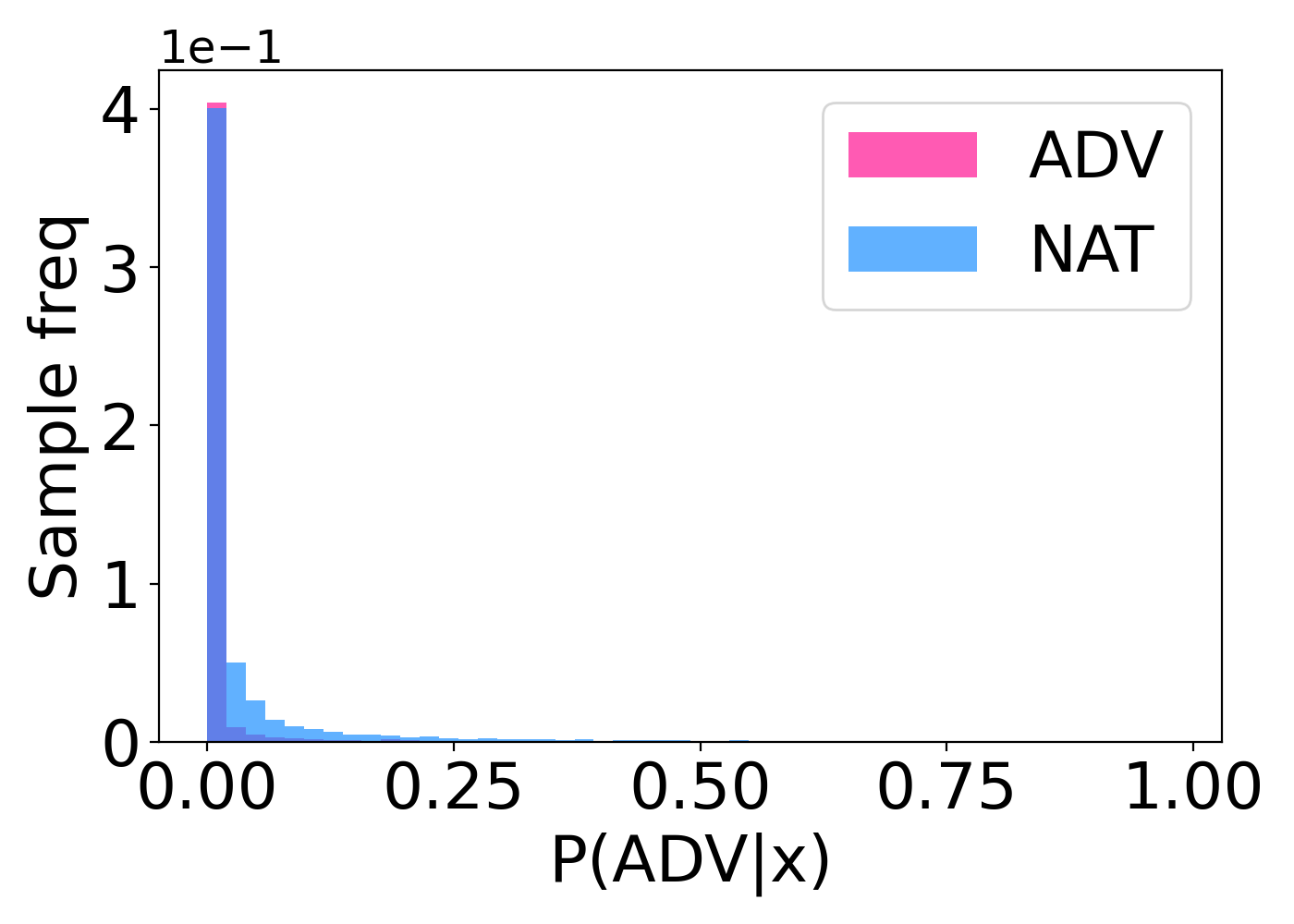}
		\vspace{-1.5\baselineskip}
		\caption{PGD-L$_1$-40-Gini}
		\label{fig:h_g}
	\end{subfigure}
	\caption{In pink the results for the adversarial examples and in blue the ones for the naturals. In this simulation, we consider a subset of the available detectors (ACE, KL, FR). Under each plot, we indicate the tested attack configuration parameters: algorithm-L$_p$-$\varepsilon$-loss.}
	\label{fig:evaluation_2}
\end{figure}
The specific shape in the histograms depends on the set of considered detectors. To shed light on this fact, we include the plots in~\cref{fig:evaluation_2} in which we consider a subset of the available detectors (ACE, KL, FR). These plots should be compared with the ones in~\cref{fig:evaluation}.

\end{document}